\documentclass[10pt, english, oneside, a4paper]{article}

\usepackage[utf8]{inputenc}
\usepackage{bm}
\usepackage[style=american]{csquotes}
\usepackage{rotating}
\usepackage[hidelinks]{hyperref}
\usepackage[left=1.25in,right=1.25in,top=1.25in,bottom=1.25in]{geometry} 
\usepackage{etoolbox}
\usepackage{tcolorbox}
\usepackage{enumitem}
\usepackage{scrextend}
\usepackage{centernot}
\usepackage{lipsum}
\usepackage{amsfonts}
\usepackage{amssymb}
\usepackage{amsmath}
\usepackage{amsthm}
\usepackage{amsopn}
\usepackage{graphicx}
\usepackage{epstopdf}
\usepackage{benstyle_arXiv}
\usepackage{tikz}
\usepackage{bbm}
\usepackage{mathtools}
\usepackage{array}
\usepackage{rotating}
\usepackage{booktabs}
\usepackage[normalem]{ulem}

\usepackage{graphicx}
\usepackage{varioref, babel, fancyvrb, listings, amsmath,amsthm,  mathtools,bbm}
\usepackage{amssymb}
\usepackage{dsfont}
\usepackage{enumitem}
\usepackage{array}
\usepackage{booktabs}
\let\oldcap\cap
\usepackage{mathabx}
\let\cap\oldcap
\usepackage{cite}
\usepackage{times}
\usepackage{listings}
\usepackage[normalem]{ulem}
\usepackage[style=american]{csquotes}
\usepackage{cleveref}

\usepackage{lipsum}
\usepackage{amsfonts}
\usepackage{graphicx}
\usepackage{epstopdf}

\usepackage{algorithm}
\usepackage{algpseudocode}

\usepackage[dvipsnames]{xcolor}
\usepackage{pgfplots}
\pgfplotsset{compat=1.18}
\usepackage{subcaption}
\usepackage{tikz}

\usepackage{bm}
\usepackage{dsfont}

\usepackage{caption}
\usetikzlibrary{calc}
\usetikzlibrary{decorations.pathreplacing,calc}

\ifpdf
  \DeclareGraphicsExtensions{.eps,.pdf,.png,.jpg}
\else
  \DeclareGraphicsExtensions{.eps}
\fi

\newcommand{\X}{\mathcal{X}}
\newcommand{\Y}{\mathcal{Y}}

\newcommand{\set}[1]{\{#1\}}
\newcommand{\R}{\mathbb{R}}
\newcommand{\C}{\mathbb{C}}
\newcommand{\V}{\mathcal{V}}
\newcommand{\drarrow}{\rightrightarrows}
\newcommand{\vertiii}[1]{{\left\vert\kern-0.25ex\left\vert\kern-0.25ex\left\vert #1 
    \right\vert\kern-0.25ex\right\vert\kern-0.25ex\right\vert}}
\newcommand{\Mo}{\mathcal{M}_1}
\newcommand{\Mt}{\mathcal{M}_2}

\DeclareMathOperator*{\diam}{diam}

\numberwithin{equation}{section}

\theoremstyle{plain}
\newtheorem{theorem}{Theorem}[section]

\newtheorem{proposition}[theorem]{Proposition}

\newtheorem{mainxx}[theorem]{Main result}

\theoremstyle{definition}
\newtheorem{definition}[theorem]{Definition}

\theoremstyle{remark}
\newtheorem{remark}[theorem]{Remark}

\usepackage{enumitem}
\usepackage[percent]{overpic}
\usepackage{pict2e}

\usepackage{amsopn}
\usepackage{cite}

\AtBeginEnvironment{figure}{\setlength{\tabcolsep}{2pt}}

\tikzstyle{stuff_fill}=[rectangle,draw,fill=pink,minimum size=0.5em]
\usetikzlibrary{calc}
\usetikzlibrary{decorations.pathreplacing,calc}
\newcounter{arrow}
\title{On Hallucinations in Inverse Problems: Fundamental Limits and Provable Assessment Methods \thanks{ The authors gratefully acknowledge the Gauss Centre for Supercomputing e.V. (www.gauss-centre.eu) for funding this project by providing computing time through the John von Neumann Institute for Computing (NIC) on the GCS Supercomputer. D.I was supported by a PhD fellowship from the Deutsches Zentrum für Luft und Raumfahrt (DLR). N.M.G. acknowledges research sponsored by the Laboratory Directed Research and Development Program of Oak Ridge National Laboratory, managed by UT-Battelle, LLC, for the U. S. Department of Energy.}}

\author{David Iagaru
  \thanks{CMAP, Ecole polytechnique, Institut Polytechnique de Paris, 91120 Palaiseau, France.}
  \thanks{German Aerospace Center (DLR), Remote Sensing Technology Institute, Wessling, Germany.}
  \and Nina M.~Gottschling
  \thanks{Computing and Computational Sciences, Oak Ridge National Laboratory, Oak Ridge, Tennessee.}
  \and Anders C.~Hansen
  \thanks{DAMTP, University of Cambridge, Cambridge, UK}
  \and Josselin Garnier
  \thanks{CMAP, CNRS, Ecole polytechnique, Institut Polytechnique de Paris, 91120 Palaiseau, France.}
}

\usepackage{amsopn}

\ifpdf
\hypersetup{
  pdftitle={Average Kernel Sizes},
  pdfauthor={David Iagaru, Nina M. Gottschling, Josselin Garnier, Anders C. Hansen}
}
\fi

\makeatletter
\renewenvironment*{displayquote}
  {\begingroup\setlength{\leftmargini}{0.6cm}\csq@getcargs{\csq@bdquote{}{}}}
  {\csq@edquote\endgroup}
\makeatother

\begin{document}

\maketitle

\begin{abstract}
Artificial intelligence (AI) has transformed imaging inverse problems, from medical diagnostics to Earth observation. Yet deep neural networks can produce \enquote{hallucinations}, realistic-looking but incorrect details, undermining their reliability, especially when ground truth data is unavailable.
We develop a theoretical framework showing that such hallucinations are not merely artifacts of particular models, but can arise from the ill-posed nature of the inverse problem itself. We derive necessary and sufficient conditions for hallucinations, together with computable bounds on their magnitude that depend only on the forward model.
Building on this theory, we introduce algorithms to: (1) estimate the minimum hallucination magnitude achievable by any reconstruction model for a given input; (2) assess the faithfulness of reconstructed details by a given reconstruction model. Experiments across three imaging tasks demonstrate that our approach applies broadly, including to modern generative models, and provides a principled way to quantify and evaluate AI hallucinations.
\end{abstract}


\section{Introduction}\label{sec:intro}
While deep learning has revolutionised inverse problems, its safe deployment is hindered by three primary reliability concerns: hallucinations, instabilities, and performance volatility \cite{renard_variability_2020}. Hallucinations manifest as high-fidelity features that are factually false; instabilities reflect heightened sensitivity to measurement noise; and performance volatility refers to significant fluctuations in reconstruction quality across the data, yielding high-fidelity results for some samples while failing on seemingly similar images. In many applications, the risk of generating realistic but unfaithful content can impede the safe deployment of AI methods for inverse problems. The choice of ``hallucinate'' as the Cambridge Dictionary's word of the year in 2023 illustrates this open problem \cite{hallucinate2023}. The problem of AI hallucinations persists, as the \textit{Financial Times} \cite{ft2026} highlighted that,
\begin{displayquote}
        \enquote{\textit{AI hallucinations haunt users more than job losses.}}
\end{displayquote}

\noindent A first step toward training AI methods that do not suffer from hallucinations is the assessment and identification of hallucinated outputs. Consider the inverse problem of recovering $x$ from noisy measurements
\begin{equation}
\label{eq:sampling1}
y = F(x,e),
\quad
x \in \mathcal{M}_1 \subset \mathcal{X},
\;
e \in \mathcal{E} \subset \mathcal{Y},
\end{equation}
\noindent where $x$ represents the signal of interest, $e$ represents the noise, and $F: \mathcal{M}_1\times\mathcal{E}\rightarrow \mathcal{M}_2$ is the forward model, with $\mathcal{M}_2=F(\mathcal{M}_1\times\mathcal{E})$. The sets $\cM_1$ and $\E$ represent the set of signals of interest and the set of potential noise vectors. The ambient spaces are $\Y = \mathbb{K}^{d_2}$ and $\X = \mathbb{K}^{d_1}$, with $\mathbb{K} \in \set{\R, \C}$ for $d_1,d_2 \in \mathbb{N}$. The set $\mathcal{M}_1$ corresponds to what is commonly referred to in the literature as the support of the data distribution \cite{Fang2024LearnabilityOOD}. In many cases, $\mathcal{M}_1$ is a finite set of input training and validation examples. For example, $\mathcal{M}_1$ can be a set of realistic input images, which follow a certain distribution that concentrates near a nonlinear submanifold of $\mathcal{X}$. Most, but not all, of our numerical examples work with a finite sample set, but we keep the assumptions very general to incorporate uncountable model sets. Recent theoretical work on inverse problems \cite{gottschling2025troublesome,gottschling2023existence} shows that, in the case of linear forward models, the theoretical minimum error for solutions and AI hallucinations can be derived from assumptions on the null space of the forward model in \eqref{eq:sampling1}. Despite theoretical and experimental efforts, AI hallucinations, instability, and unpredictable performance remain a problem in computing solutions to inverse problems.

Concerns about instabilities and artifacts in learning-based solutions to inverse problems have been demonstrated in \cite{PNAS_paper}. From a theoretical standpoint, hallucinations and other reconstruction artifacts are not merely empirical failures, but an intrinsic property of ill-posed inverse problems \cite{gottschling2025troublesome}. Hallucinations are a well-recognised failure mode of generative models, as noted in \cite{Huang2025HallucinationDetection}:
\begin{displayquote}
\enquote{\textit{One known pitfall of generative models is their susceptibility to hallucinate by producing information/images that are not based on factual data or reality. These hallucinations can range from subtle structural and/or colour inconsistencies to entirely fabricated content.}}
\end{displayquote}
In medical imaging, these effects may translate into anatomically incorrect content. For example, \cite{laves2020uncertaintyestimationmedicalimage} reports:
\begin{displayquote}
        \enquote{\textit{deep models trained on large data sets tend to hallucinate and create artifacts in the reconstructed output that are not anatomically present[.]}}
\end{displayquote}
\noindent In particular, in MRI, \cite{Hakim2026deep} found that hallucinations are not rare events:
\begin{displayquote}
\enquote{\textit{[Deep Resolve Boost (DRB) applied to accelerated acquisition (ACC)] DRB-ACC sequences demonstrated good or fair image quality in 94.5$\%$ of cases, with improved or unchanged quality
compared with standard acquisition in 95.7$\%$ of sequences. However, anatomic delineation was inferior in key regions such as the
hippocampus, brainstem, and cerebellum. Artifacts were more pronounced in 27.3$\%$ and newly introduced in 84.8$\%$ of the accelerated DRB sequences, commonly affecting the brainstem and deep gray matter.}}
\end{displayquote}

\begin{figure}[h!]
    \centering
    \includegraphics[width=0.6\textwidth]{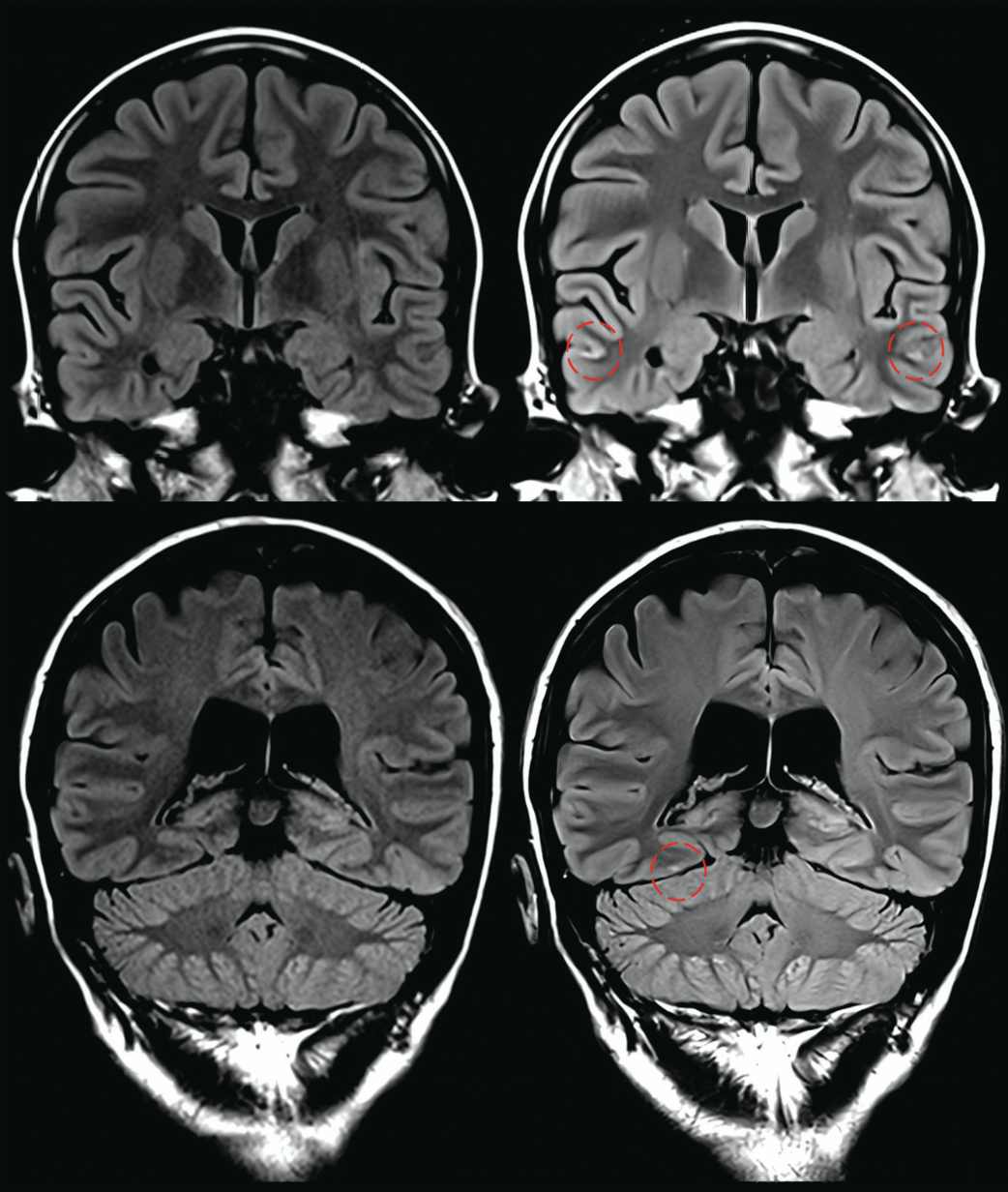}
    \caption{Figure 5 from \protect\cite{Hakim2026deep} shows \textit{"newly introduced artifacts in the cortex/juxtacortical regions on Deep Resolve Boost acquisition"} for Coronal FLAIR images from 2 patients. \textit{"In both cases, the DRB-ACC images (right) show slightly improved overall quality with reduced noise compared with standard images (left). However, new artifacts are visible in the temporal lobes (red circles), potentially mimicking pathologic signal abnormalities. Correlation with other sequences in these MRI examinations, including high-resolution 3D images, confirmed the absence of lesions in these regions."}\protect\cite{Hakim2026deep}}
    \label{fig:mriexamplestudy}
    \vspace{-15pt}
\end{figure}

Figure \ref{fig:mriexamplestudy} provides an example of newly introduced false artifacts from \cite{Hakim2026deep}. Similar commentary on such artifacts in inverse problems and various applications can also be found in 
\cite{hakim2025deep,laine2021avoiding, liu2022medical, morshuis2022adversarial, lustig_pnas22, mean_field_pmlr21, varoquaux2022machine, Akhaury_2024} and references therein. Figure 1 in \cite{gottschling2025troublesome} presents several examples of hallucinating AI-based decoders for different imaging tasks. In all cases, the recovered images exhibit realistic features that are absent in the corresponding ground-truth images.

Importantly, the notion of hallucinations in learned reconstruction remains inherently ambiguous. Rather than being a sharply defined artifact, the definition of AI hallucinations depends on how reconstruction fidelity is evaluated in a given application: a feature may be unacceptable for quantitative analysis because it is not strictly data-consistent, yet remain diagnostically useful, or appear reasonable under one reconstruction prior but incorrect under another \cite{Arr19}. In some applications, for example in microscopy and medical imaging, multiple plausible reconstructions can explain the same measurements. The boundary between a valid reconstruction and a hallucination is context-dependent and can only be determined by what is considered acceptable in a given application, such as in \cite{fastmri20, bel19}.

\noindent Additionally, trustworthy AI-driven decoders for inverse problems require verifiability \cite{kapoor_leakage_2023}. In particular, a decoder should not only return a reconstruction, but also provide reliable information about its confidence \cite{trustUQ} and whether the reconstruction remains within the prescribed set of signals of interest \cite{liu2020algorithmsverifyingdeepneural, sayyed2026encore, rifat2024dardadomainawarerealtimedynamic}. In inverse problems, this calls for theoretical tools in two directions: first, characterising the fundamental accuracy limits imposed by the forward model, and second, identifying inputs for which the inverse problem is intrinsically unstable and therefore prone to hallucinations. In particular, Theorems 4.1, 4.3, and 4.5 of \cite{gottschling2025troublesome} show that, for ill-posed and ill-conditioned inverse problems, over-accurate reconstructions on specific data points cause hallucinations with high probability. Another difficulty is that AI hallucinations are often small details that are not recognizable using popular norms and metrics, such as the root mean squared error or, for images, the structural similarity index \cite{wang2004image}. In this work, we develop a theoretical framework for detecting and quantifying AI hallucinations in solutions to inverse problems. We complement the sufficient conditions for AI hallucinations from Theorem 4.1 of \cite{gottschling2025troublesome} with necessary conditions, yielding a characterisation of when hallucinations occur. Based on this characterisation, we then develop algorithms to estimate unavoidable hallucination magnitudes and to assess hallucinations for given reconstructions.

\subsection{Related Work}
\noindent In the following, we provide a brief review of related work that proposes methods for detecting hallucinations in reconstructions for inverse problems. \cite{li2025chemestimatingunderstandinghallucinations} propose wavelet-based texture analysis to identify and quantify hallucinated artifacts and position this research area as an essential frontier for "trustworthy computer vision". Early work by \cite{bhadra2021hallucinations}, focused on tomographic image reconstruction, proposes decomposing the reconstruction error of a decoder into the measurement and null spaces of the forward model. While this work provides a rigorous framework for understanding how priors introduce non-physical features in tomography, its practical application can be hindered by the requirement of a ground-truth image to compute the null space components. Parallel to these analytical approaches, a growing category of methods leverages generative and diffusion models to produce an ensemble of reconstructions that are consistent with the measurements \cite{Bhadra2022MiningTM, OpenSRTest2025}. In these frameworks, the pixelwise variance of the generated images is interpreted as an uncertainty quantification (UQ) score, where increased variance is interpreted to indicate a higher likelihood of incorrect structural details and hallucinations. However, the reliability of these scores is contingent on the decoder’s ability to remain within the true data distribution is an open problem. Despite significant progress in image generation, ensuring that diffusion-based models do not produce realistic but entirely out-of-distribution (OOD) samples remains an active area of research \cite{hendrycks2019scaling,bitterwolf2020certifiably, sayyed2026encore,rifat2024dardadomainawarerealtimedynamic}. Without theoretical guarantees for filtering out OOD samples, the resulting uncertainty estimates may be miscalibrated, potentially masking hallucinations as high-confidence features. Recent information-theoretic perspectives further elucidate this vulnerability. \cite{blau2018perception} demonstrate a fundamental trade-off between perception and distortion, suggesting that the drive for high perceptual quality inherently increases the risk of hallucinations. Other Bayesian-inspired detection methods attempt to quantify this risk by comparing the posterior distribution to the prior distribution \cite{SpottingHallSampson2023} or to a reference distribution using the Hellinger distance \cite{HallIndexTivnan}.

\subsection{Contributions}

\noindent In this work, we address some of the challenges encountered in previous work on detecting hallucinations in solutions to inverse problems. We avoid the need of ground truth data or distributional assumptions that are not satisfied in some applications, such as in Bayesian inference, where incorrectly specified priors can bias posterior inference \cite{Barco_2025}. Inspired by descriptions of AI hallucinations in inverse problems in various applications, see $\S$ \ref{sec:intro} and \cite{fastmri20, PNAS_paper, bel19, laine2021avoiding, liu2022medical, morshuis2022adversarial, lustig_pnas22, mean_field_pmlr21, varoquaux2022machine, Akhaury_2024}, we provide a definition of hallucinations that formalises these descriptions. Our framework is based on assumptions that are satisfied in most inverse problems encountered in scientific computing and imaging, such as in MRI \cite{melba}, super-resolution of multispectral satellite data \cite{gottschling2025average}, microscopy \cite{yuan2016structured}, astronomy \cite{Richardson1972BayesianBasedIM,SpottingHallSampson2023}, geophysical imaging \cite{geophys_inv_pb}, and 
tomography \cite{natterer1986mathematics}. We now present a nontechnical summary of our main results and refer to $\S$ \ref{sec:main} for the formal statements.\newline

\noindent We show that, for possibly non-linear forward models with additive noise, detail transfer can occur only when the transferred detail is almost invisible in the measurement space, see Theorem \ref{prop:faithhall}. This extends the results from linear forward models with additive noise in \cite{gottschling2025troublesome} to non-linear forward models. In particular, we consider forward models with additive noise in \eqref{eq:sampling1}, where $F : \cM_1 \times \mathcal{E} \rightarrow \cM_2, (x,e) \mapsto F(x,e)=f(x)+e$ and $f:\mathcal{X} \to \mathcal{Y} $ is continuous and the image of the forward model $F$ is given by $\mathcal{M}_2 = F(\mathcal{M}_1\times \mathcal{E})$.  For set-valued decoders, their error is quantified with the Hausdorff distance $d$, which is introduced formally in $\S$ \ref{sec:problemoutl}. We say that a decoder $\phi$ hallucinates through \textit{"detail transfer"}, if it reconstructs $x + x_{det}$ from inputs $y = F(x,e)$, by incorrectly transferring false details $x_{det}$ to a ground truth $x \in \mathcal{M}_1$. Concise definitions are introduced in $\S$ \ref{sec:theory}. Theorem \ref{thm:iff_shift} provides necessary and sufficient conditions that enable computationally testing for hallucinations through the incorrect transfer of details and do not require knowledge of a ground truth. 

\begin{mainxx}[Necessary and Sufficient Conditions for Detail Transfer - Theorem \ref{thm:iff_shift}]
    Consider the inverse problem \eqref{eq:sampling1} and a decoder $\phi: \mathcal{M}_2 \drarrow  \mathcal{X}$ whose reconstructions are realistic, $\phi(y) \subset \mathcal{M}_1$, and consistent with their inputs, in the sense that
    $f(\phi(y))-\{y\} \subset \mathcal{E}$ for all $y \in \mathcal{M}_2$.
    The decoder $\phi$ reconstructs $x + x_{det}$ from measurements $y = F(x,e)$ by incorrectly transferring false details $x_{det}$ to a ground truth $x \in \mathcal{M}_1$ if and only if, for a suitable set of admissible noise vectors $\mathcal{V} \subset \mathcal{E}$, the following two conditions hold:
    \begin{itemize}
    \item[(i)] there exists $e \in \mathcal{V}$, such that the reconstruction of the ground truth with the false detail is perfectly accurate, $x+x_{det} \in \phi(f(x+x_{det})+e)$,
    \item[(ii)] for all $e \in \mathcal{V}$, the point-wise error, $d(\phi(f(x+x_{det})+e),x+x_{det})$, is too small.
    \end{itemize}
\end{mainxx}
\noindent Theorem \ref{thm:iff_shift} provides theoretical guarantees for the occurrence of hallucinations. To assess the conditions in Theorem \ref{thm:iff_shift}, we provide quantifiable bounds on the notion of a \textit{"too small error"}. These are computable lower bounds to the error of decoders with consistent reconstructions, as in Definition \ref{def:consistent}, and are based on the worst-case kernel size \cite{gottschling2025troublesome}. This provides a framework to quantify the accuracy-hallucination barrier and accuracy-stability tradeoff established in \cite{comp_stable_NN22,gottschling2025troublesome}. The worst-case kernel size provides a bound to guarantee that no larger hallucination can occur:
\begin{mainxx}[The Worst-Case Kernel Size Guarantees No Large Hallucinations - Proposition \ref{prop:nogo_Fy}]
Any decoder $\phi: \mathcal{M}_2 \drarrow  \mathcal{X}$ for the inverse problem \eqref{eq:sampling1} with reconstructions that are realistic $\phi(y) \subset \mathcal{M}_1$ and consistent with its inputs, in the sense that $f(\phi(y))-\{y\} \subset \mathcal{E}$, for all $y \in \mathcal{M}_2$, does not hallucinate by transferring details $x_{det}$ with a magnitude that is larger than the worst-case kernel size.
\end{mainxx}
\noindent In addition to a guarantee of reconstructions without hallucination of a certain magnitude, we provide a guarantee that any reasonably good decoder will hallucinate small details:
\begin{mainxx}[Transfer of Small Details is Inevitable - Theorem \ref{thm:suff_hall}]
Any decoder $\phi: \mathcal{M}_2 \drarrow  \mathcal{X}$ for the inverse problem \eqref{eq:sampling1}, that either 
\begin{itemize}
\item[(i)] is too stable, in that $\sup_{e \in \mathcal{E}}d(\phi(y+e),\phi(y))$ is too small,
\end{itemize}
or that 
\begin{itemize}
\item[(ii)] predicts samples that are consistent with the measurements, in the sense that $f(\phi(y))-\{y\} \subset \mathcal{E}$, and produces sufficiently concentrated reconstruction sets.
\end{itemize}
will hallucinate small details through detail transfer.
\end{mainxx}
\noindent In Theorem \ref{thm:suff_hall}, details are considered to be small relative to a point-wise restriction of the worst-case kernel size - the diameter of the feasible sets, see \eqref{eq:feasset} or \cite{gottschling2025average}. The feasible sets are the sets of possible \textit{solutions} $x\in\mathcal{M}_1$ corresponding to a measurement $y\in\mathcal{M}_2$. Additionally, we address how to compute these bounds and provide convergence guarantees. In Algorithm \ref{alg:feasapp} and Algorithm \ref{alg:diamFy}, we provide a method to compute approximate diameters of the feasible sets, and in Theorem \ref{thm:cvg_diam}, we prove convergence of the algorithms.
\begin{mainxx}[Convergence of the Approximate Diameter of Feasible Sets - Theorem \ref{thm:cvg_diam}]
Under suitable sampling assumptions, the approximate diameters of feasible sets converge to the analytical value.
\end{mainxx}
\noindent These results are fully stated and discussed in $\S$ \ref{sec:main}. In $\S$ \ref{sec:methods} we propose an algorithmic framework for detecting hallucinations for inverse problems. This framework allows to assess whether the reconstructions of a given decoder can be trusted. We validate the algorithms and theoretical guarantees for MRI acceleration and super-resolution of satellite images in $\S$ \ref{par:QualHall_S2SR}, \ref{par:QualHallMRI} and \ref{par:model_agn_S2SR}. We demonstrate that our framework helps to explain why hallucinations occur, using super-resolution of MNIST digits as an example in $\S$ \ref{sec:MNIST}.
We confront Definition \ref{def:detailtrans} and Theorem \ref{thm:iff_shift} with real situations in $\S$ \ref{par:verif_def_MNIST}, \ref{par:verif_def_MRI}, and \ref{par:verif_def_S2SR}. This follows the experimental setup described in $\S$ \ref{sec:exp_verif_setup}. We quantitatively validate the theoretical results from $\S$ \ref{par:bounds_dettrans} for MRI in $\S$ \ref{par:QuantHallMRI}. We also validate the theoretical results for super-resolution of multispectral satellite data in $\S$ \ref{par:quant_hall_S2SR}. Both validations follow an experimental setup described in $\S$ \ref{sec:exp_quant_setup}. We provide elements to reproduce the experiments in the following Github repository \url{https://github.com/davidiagraid/AccuracyBounds_private}.

\section{Preliminaries}\label{sec:theory}
In this section, we formulate and define the main problem studied in this work and, then, introduce a novel framework to mathematically capture the occurrence of AI hallucinations in solutions to inverse problems.  
\subsection{Problem Outline}\label{sec:problemoutl}
 \noindent We define an inverse problem as a triple $(F,\mathcal{M}_1, \mathcal{E})$ that describes a forward problem, equipped with the task from \eqref{eq:sampling1}. For any norm $\| \cdot \|$ on $\mathbb{K}^d$, we define closed balls around $z \in \mathbb{K}^{d}$ with radius $\delta>0$ by $B(z,\delta):=\{z'\in \mathbb{K}^d:\quad \|z'-z\| \leq \delta\}$ and open balls with the same center and radius $B^o(z,\delta):=\{z'\in \mathbb{K}^d:\quad \|z'-z\| < \delta\}$. The vector space $\mathcal{X}$ is equipped with a seminorm $\|\cdot\|$ and $\mathcal{Y}$ is equipped with a norm $\vertiii{\cdot}$. We equip $[0,\infty)$, $\mathcal{X}$ and $\mathcal{Y}$ with the standard Borel sigma algebras and all subsets with the corresponding induced sigma algebras. In this work, we assume that $\mathcal{M}_1 \subset \mathcal{X} = \mathbb{K}^{d_1}$ is bounded, $\mathcal{M}_2 \subset \mathcal{Y} = \mathbb{K}^{d_2}$ and $\mathcal{E} = B^o(0,\epsilon) \subset \mathcal{Y}$ with $\epsilon>0$, for $\mathbb{K} \in \{ \mathbb{R}, \mathbb{C} \}$ and $d_1, d_2 \in \mathbb{N}$. The forward model in \eqref{eq:sampling1} is defined as $F : (x,e) \in \cM_1 \times \mathcal{E} \rightarrow \cM_2, (x,e) \mapsto F(x,e)=f(x)+e$, where  $f:\mathcal{X} \to \mathcal{Y} $ is a continuous forward map and the image of the forward model is defined as $F(\mathcal{M}_1\times \mathcal{E})=\mathcal{M}_2$. If $f$ is linear, we write $F(x,e) = fx+e$ for every $x \in \X, e \in \E$. A map approximately solving \eqref{eq:sampling1} by providing for every 
 \textit{measurement} $y= f(x)+e \in \mathcal{M}_2$ an approximation $\hat{x} \in \mathcal{X}$ to a \textit{solution} $x$ is referred to as a decoder. In the following, we denote non-empty compact-set valued decoders by 
 \[
 \begin{aligned}
      \phi: \mathcal{M}_2 &\drarrow \mathcal{X} \nonumber \\
y &\mapsto \phi(y) \subseteq \X   .
 \end{aligned}
 \]
\noindent The image of a decoder $\phi$ on a subset $S$ of $\cM_2$ is defined as 
\[
\begin{aligned}
    \phi(S) = \bigcup_{y \in S} \phi(y)   .
\end{aligned}
\]
\noindent Considering multi-valued decoders allows the use of stochastic models, including diffusion-based generative models. Typically, when a finite number of reconstructions $(z_l)_{1 \leq l \leq L}$ are observed for  a given $y$, it can be modeled with the reconstruction set $\phi(y) = \set{z_l,\quad  1 \leq l \leq L}$. In Bayesian modeling \cite{dashti2017bayesian}, the output may instead be a posterior distribution $p(\cdot\mid y)$ on $\mathcal{X}$. In this case, $\phi(y)$ can be taken to be a credible set $B\subseteq\mathcal{X}$ satisfying $\int_B dp(x\mid y)\geq 1-\alpha$ for a prescribed level $1-\alpha$. Throughout this text, we denote by $d(A,B)$ the Hausdorff distance induced by $\|\cdot \|$ between two compact subsets $A$ and $B$ of $\X$. With a slight abuse of notation, if one of the sets is a singleton ($A = \{x\}$), we write $d(x, B)$. Note that if $\| \cdot \|$ is a seminorm, then $d$ is a semi-distance. For a multi-valued decoder with compact outputs, the point-wise error with respect to $x \in F_y$ is typically expressed as 
\[
 \begin{aligned}
 d(x, \phi(y)) = \sup_{u \in \phi(y)} \|u - x\|.
 \end{aligned}
 \]
\noindent This definition coincides with the standard error $\|\phi(y) - x\|$ when the decoder is single-valued. Using the Hausdorff distance between the reconstruction and a point ensures that we account for the worst-case decoder reconstruction, which is particularly relevant in applications where even a single hallucination in the reconstructions is problematic. The Hausdorff distance is thus appropriate in scenarios where a unique ground truth exists, despite the potential for multiple valid solutions given $y$ and $\cM_1$. The map denoted by $\pi_1\colon \mathcal{M}_1 \times \mathcal{E} \to \mathcal{M}_1, (x,e) \mapsto \pi_1(x,e)= x$ is the projection on the first component. Following \cite{gottschling2023existence}, the set of possible \textit{solutions} $x\in\mathcal{M}_1$ corresponding to a measurement $y\in\mathcal{M}_2$ is referred to as the \emph{feasible set} or projection onto the first component of the pre-image of $y \in \Mt$ and it is denoted by
\begin{equation}
        \begin{aligned}\label{eq:feasset}
    	F_y:=\pi_1(F^{-1}(y)) = \{x \in \Mo: \exists e \in \E \text{ s.t. } F(x,e)=y\}.
    \end{aligned}
\end{equation}
\noindent If $f$ is linear, its condition number is independent of the model set $\mathcal{M}_1$ used in the forward problem \eqref{eq:sampling1}. However, the ambiguity of the inverse problem over $\mathcal{M}_1$, as captured by the feasible sets, does depend on $\mathcal{M}_1$. For example, if the condition number of $f$ is infinite, then one can choose a model set $\mathcal{M}_1$ for which at least one feasible set contains more than one element.

\subsection{Definitions}
\noindent In addition to the standard inverse problem framework introduced in the previous subsection, we introduce key concepts to quantify the degree of ill-posedness of an inverse problem following \cite{gottschling2023existence,gottschling2025average}. Additionally, we provide a definition capturing the notion of AI hallucinations - described in \cite{bhadra2021hallucinations} as
\begin{displayquote}
    \enquote{\textit{false structures [... that] appear in the reconstructed
image that are absent in the object being imaged.}}
\end{displayquote}
\noindent Before presenting definitions to quantify the degree of ill-posedness of an inverse problem and AI hallucinations, we define what it means for a decoder to provide reasonable solutions.

\begin{definition}[Consistent Decoders]\label{def:consistent}
Let $(F,\mathcal{M}_1,\mathcal{E})$ be a forward problem. Let $\phi: \mathcal{M}_2 \drarrow \mathcal{X}$ be a decoder for the corresponding inverse problem \eqref{eq:sampling1}. The reconstruction is \textbf{consistent} for a fixed $y \in \cM_2$, if
\[
    f(\phi(y))- \set{y} = \set{f(z)-y:z \in \phi(y)} \subseteq \mathcal{E}.
\]
The decoder $\phi$ is \textbf{consistent}, if for all $y \in \mathcal{M}_2$ the reconstruction $\phi(y)$ is consistent.
\end{definition}
\begin{remark}
The underlying idea behind the consistency property is to check whether the forward model applied to the decoder's reconstruction is close to the measurement within the noise level. This idea appears in various applications. For instance, it is referred to as the 'consistency' property in remote sensing \cite{OpenSRTest2025}. It is also used in \cite{WaldRSensing1997} through Wald's protocol to ensure the applicability of pan-sharpening algorithms. Additionally, it is applied in medical image reconstruction \cite{Hall_DPS}. To enforce consistency, the difference between the measurement and the forward model applied to the reconstruction has also been used as a quantity to minimize. For example, when consistency is enforced as a loss term, it is called a data discrepancy functional. This functional is used alongside a regularization term in variational approaches for solving inverse problems \cite{ArridgeEtAlACTA}. It is also employed in deep-learning approaches and in methods to learn an approximate forward model from the data \cite{KamyabFaithful2021, Adler_2017}.
\end{remark}
\noindent As a measure of how ill-posed an inverse problem is in the worst-case, we consider the worst-case kernel size in inverse problems as defined in \cite{gottschling2023existence}. 
\begin{definition}[Worst-Case Kernel Size \cite{gottschling2023existence}]\label{def:kersizeoptmap}
    Let $(F,\mathcal{M}_1,\mathcal{E})$ be a forward problem. Then, for the corresponding inverse problem \eqref{eq:sampling1},
    the \textbf{worst-case kernel size} is defined as
    \[
    \begin{aligned}
     \operatorname{Kersize}(F,\mathcal{M}_1, \mathcal{E}) =    \sup_{\substack{(x,e),(x',e')\in \mathcal{M}_1\times\mathcal{E} \\F(x,e)=F(x',e')}} \|x-x'\|.
    \end{aligned} 
    \]     
\end{definition}
\noindent \cite[Theorem 3.4]{gottschling2023existence}
shows that half the worst-case kernel size of a specific inverse problem is a sharp lower bound on the worst-case error of any decoder for solving this inverse problem. 
\begin{remark}[Lower Worst-Case Accuracy Bound, Theorem 3.4 \cite{gottschling2023existence}]

    Let\newline $(F,\mathcal{M}_1,\mathcal{E})$ be a forward problem. Any decoder $\phi: \mathcal{M}_2 \drarrow \mathcal{X}$ for the corresponding inverse problem \eqref{eq:sampling1} satisfies
    \[
    \begin{aligned}
        \frac{1}{2}\operatorname{Kersize}(F,\mathcal{M}_1, \mathcal{E})\leq \sup_{(x,e) \in \mathcal{M}_1\times\mathcal{E}} d(x,\phi(F(x,e))).
    \end{aligned} 
    \]
\end{remark}
\noindent Additionally, the worst-case kernel size can be considered at a sample-wise level by relating it to the feasible sets in \eqref{eq:feasset}: The worst-case kernel size can be expressed in terms of the feasible sets $(F_y)_{y \in \cM_2}$ as 
\begin{equation}
    \begin{aligned}\label{eq:diamker}
        \operatorname{Kersize}(F,\mathcal{M}_1, \mathcal{E}) = \sup_{y \in \cM_2} \mathrm{diam}(F_y).
    \end{aligned}
\end{equation}
\noindent The last two remarks and definitions apply to any measurable forward model $F$. However, in this work we restrict to forward models with additive noise $F: \cM_1 \times \mathcal{E} \rightarrow \cM_2, (x,e) \mapsto F(x,e)=f(x)+e$. This ensures that the Definition \ref{def:consistent} of consistency is experimentally falsifiable.

\noindent \textit{Detail transfer: defining AI hallucinations in solutions to inverse problems.} Hallucinations and the incorrect transfer of details in solutions to inverse problems have been observed in various applications, see \cite{hoff21,bel19,bhadra2021hallucinations} or, for example in  \cite[Figure 1]{gottschling2025troublesome}. These observations of hallucinations are formalised in Definition \ref{def:detailtrans} and explained in the following.
\begin{definition}[Detail Transfer - Hallucinating Decoders]\label{def:detailtrans}
  Let $(F,\mathcal{M}_1,\mathcal{E})$ be a. A decoder  $\phi: \mathcal{M}_2 \drarrow \mathcal{X}$ for the corresponding inverse problem \eqref{eq:sampling1} \textbf{hallucinates} $x_{det} \in \mathcal{X}$ of size $\eta>0$ on $x \in \mathcal{M}_1$ for $\mathcal{V} \subset \mathcal{E}$, if the following conditions hold:
    \begin{enumerate}
        \item[(1)] $x+x_{det} \in \mathcal{M}_1$,
        \item[(2)] the detail is \textbf{incorrectly transferred} by the decoder: for all $e \in \mathcal{V}$,
\begin{equation}
    \begin{aligned}\label{eq:detailtransitionzzz0}
        d(\phi(f(x)+e), x+x_{det}) \leq \eta ,\\
        d(\phi(f(x)+e), x) \geq \eta,
    \end{aligned}   
\end{equation}
\noindent and there exists $e_0 \in \V$ such that $x+x_{det} \in \phi(f(x)+e_0)$.
\end{enumerate}
\end{definition}
\noindent Part $(1)$ of Definition \ref{def:detailtrans} states that hallucinations in reconstructions are realistic-looking. This means that the reconstructions with the incorrectly transferred detail are contained in the model set $\mathcal{M}_1$. The underlying assumption here is that there exists a subset of data $\mathcal{M}_1 \subset \mathcal{X}$, which describes possible reference data points or solutions. In practice for AI applications, this could include a union of subsets of the test, training, or validation data sets. In theoretical work, the set can also be equipped with additional structure to become a manifold.  Note that part $(1)$ alone is not sufficient to define a \textit{"incorrectly transferred detail of size $\eta$"}, e.g. without part $(2)$ for $\mathcal{M}_1 = \mathcal{X}$ all sufficiently large $x \in \mathcal{X}$ are details.
Part $(2)$ states that the decoder transfers the detail to reconstructions from measurements generated by $x$, although the detail $x_{det}$ is not present in $x$. Since $d$ is the Hausdorff distance to a singleton, this condition is strongest for set-valued decoders: it requires all reconstructions in $\phi(f(x)+e)$ to be close to the signal containing the false detail. This captures systematic hallucination of the detail across the reconstruction set.
According to \cite{gottschling2025troublesome}, the definition of a detail relies solely on its (semi-)norm exceeding a certain threshold. A key point for the definition to correspond to incorrect detail-transfer is the ability of the chosen (semi-)norm to distinguish semantic differences. Specifically, it must differentiate between two points in the model set $\cM_1$. For example, this distinction can be achieved by focusing on detail-sized areas (as introduced in $\S$ \ref{sec:metrics}). Alternatively, it can be done by applying the definition after wavelet transformation. However, a single threshold may not perfectly separate the two cases. By combining parts $(1)$ and $(2)$ of Definition \ref{def:detailtrans}, a detail $x_{det}$ can be understood as a potentially local feature. Its presence or absence must be perceptually salient, as described in part $(2)$. Additionally, it must be relevant to the task under consideration, as outlined in part $(1)$. Conversely, large norm differences need not correspond to hallucinated details if they do not represent task-relevant or semantically meaningful changes within the model set $\mathcal{M}_1$. The distinction between incorrect and correct details cannot be made perfectly with a single threshold separating the two cases, because hallucination is a fuzzy concept itself. However, experts can determine ranges of values generally corresponding to incorrect detail transfers. This can be done along with an appropriate choice of metric. In $\S$ \ref{sec:metrics}, we propose a family of such metrics focused on specified regions of interest for inverse problems in imaging. Although more sophisticated metrics may exist, the proposed ones are computationally simple and provide satisfactory performance in our experiments. Note that there may be multiple values $\eta$ for which Definition \ref{def:detailtrans} is satisfied, given a point $x$, a detail $x_{det}$ and set of noise vectors $\V$. The set of thresholds often is an interval and will be referred to as \textit{hallucination size interval}.

\section{Main Results}\label{sec:main}
In this section, we use the framework for inverse problems presented in $\S$ \ref{sec:problemoutl}. We also use the definitions from the preliminaries. Our goal is to provide quantifiable necessary and sufficient conditions for the occurrence of hallucinations in solutions to inverse problems. The section is divided into three subsections. Firstly, $\S$ \ref{sec:det_transfer} provides necessary and sufficient conditions for detail transfer. Secondly, $\S$ \ref{par:bounds_dettrans} introduces quantifiable bounds on detail transfer. Thirdly, $\S$ \ref{par:algs_kersize} establishes algorithms and convergence guarantees for quantifying these bounds on data sets. The proofs of the main results are referred to the Appendix.
\subsection{Necessary and Sufficient Conditions for Detail Transfer}\label{sec:det_transfer}
\cite[Theorem 4.1]{gottschling2025troublesome} show that a decoder will incorrectly transfer a detail to other reconstructions of measurements from $x$, if it is too accurate with respect to a reference $x+x_{det}$ and the measurement of the detail is almost invisible, in the sense that the norm is small relative to the noise level. In the following, we restrict our analysis to consistent decoders, as defined in Definition \ref{def:consistent}. We extend this result by providing quantifiable necessary and sufficient conditions in Definition \ref{def:detailtrans}. In Proposition \ref{prop:faithhall}, we show that decoders with consistent reconstructions only hallucinate details that are almost invisible in the measurements.
\begin{proposition}[Detail Transfer Always Entails Almost Invisible Details]\label{prop:faithhall}
 Let $\epsilon>0$, $\E = B^o(0, \epsilon)$ and $(F,\mathcal{M}_1,\mathcal{E})$ be a forward problem. Let $x \in \mathcal{M}_1$, $x_{det} \in \mathcal{X}$ and $\phi: \mathcal{M}_2 \drarrow \mathcal{X}$ be a \textbf{consistent} decoder for the corresponding inverse problem \eqref{eq:sampling1} and $\mathcal{V} \subset \mathcal{E}$. Define $e(x,x_{det}) = f(x+x_{det})-f(x)$. 
If the decoder hallucinates $x_{det}$ of size $\eta$ on $x$ for $\mathcal{V}$, then we have $\vertiii{e(x,x_{det})}\leq 2\epsilon$.
\end{proposition}
\noindent The Proposition \ref{prop:faithhall} implies that, if a consistent decoder will hallucinate a detail $x_{det}$ on a solution $x$, then $e(x,x_{det})$ is small relative to the noise level. This means that the difference between measurements of $x_{det}$ and $x$ is close to the noise level. Therefore, no decoder can distinguish between these in the measurement space. In the case where $f$ is linear, the condition reduces to $\vertiii{fx_{det}} \leq 2 \epsilon$. Recall that the noise set is assumed to be a ball around zero $\mathcal{E}=B^o(0, \epsilon)$ with radius $\epsilon>0$. This assumption can be omitted if the set of noise $\mathcal{E}$ is symmetric. In this case, the condition $\vertiii{e(x,x_{det})}\leq 2\epsilon$ can be rewritten as $e(x,x_{det}) \in \mathcal{E} + \mathcal{E}$.
\begin{theorem}[Necessary and Sufficient Conditions for Detail Transfer]\label{thm:iff_shift}
    Let $(F,\mathcal{M}_1,\mathcal{E})$ be a forward problem. Let $x \in \mathcal{M}_1$, $x_{det} \in \mathcal{X}$ and $\phi: \mathcal{M}_2 \drarrow \mathcal{X}$ be a \textbf{consistent} decoder for the corresponding inverse problem \eqref{eq:sampling1}, such that $\phi(f(x)+\mathcal{E}) \subset \cM_1$. Let $\mathcal{V} \subset \mathcal{E}$ and $\eta \in (0,\|x_{det}\|/2]$. Define $e(x,x_{det}) = f(x+x_{det})- f(x)$.
    Then, the decoder $\phi$ hallucinates $x_{det}$ of size $\eta$ on $x$ for $\mathcal{V}$, if and only if
        \begin{enumerate}
            \item[(i)] there exists $e \in (\{-e(x, x_{det})\}+\mathcal{V})\cap  \mathcal{E}$ such that $ x+x_{det} \in \phi(f(x+x_{det})+e)$,
            \item[(ii)] for all $e \in \{-e(x, x_{det})\}+\mathcal{V}$, we have  \[
            \begin{aligned}
                d\big( \phi(f(x+x_{det})+e), x+x_{det}\big) \leq \eta.
            \end{aligned}
            \]
            
        \end{enumerate}
\end{theorem}
\noindent Theorem \ref{thm:iff_shift} provides a useful computational tool, because in Definition \ref{def:detailtrans} detecting a hallucination in a reconstruction of a decoder would require the knowledge of the ground truth $x$ and the reconstruction $x+ x_{det}$. Now part (ii) of Theorem \ref{thm:iff_shift} shows that knowing the ground truth explicitly is no longer necessary to identify when incorrect transfer of details is occurring. Hallucinations occur if the decoder is too accurate on a given data point $x'$ - the error is smaller than $\eta$ - and if there exists a detail $x_{det} \in \mathcal{X}$ such that $x'-x_{det} \in \cM_1 $  and $2 \eta \leq \|x_{det}\|$. These conditions can easily be falsified or verified in computations. Part (i) of Theorem \ref{thm:iff_shift} can usually be computed on data sets containing pairs of points $(x,y) \in \cM_1 \times \cM_2$ such that $y \in f(x) +\mathcal{E}$.  Regarding limits to the size of hallucinations, Remark \ref{rem:sizebound} shows that $\eta \leq \|x_{det} \|/2  \leq \diam(F_y)/2$.  Hence, computing and assessing part (ii) of Theorem \ref{thm:iff_shift} requires having information about the diameter of the feasible set, $F_y$, where $y = f(x+x_{det})+ e_0$ with $e_0 \in \mathcal{V}$ from  Definition \ref{def:detailtrans}.
\subsection{Quantifiable Bounds on Detail Transfer}\label{par:bounds_dettrans}
In the following, we relate the size of transferred details to computable quantities. In particular, we consider the diameters of feasible sets, $\diam(F_y)$, for a given measurement $y \in \mathcal{M}_2$. Recall that in \eqref{eq:diamker} the diameters are a local measure of the worst-case kernel size in Definition \ref{def:kersizeoptmap}. In particular, they reflect the point-wise degree of ill-posedness of the inverse problem.
\begin{proposition}[Upper Bound to the Error for Consistent Decoders]\label{prop:sizenogo1}
  Let $(F,\mathcal{M}_1,\mathcal{E})$ be a forward problem. Let $\phi: \mathcal{M}_2 \drarrow \mathcal{X}$ be a decoder for the corresponding inverse problem \eqref{eq:sampling1}. If there exists $x \in \mathcal{M}_1$ and $e\in \mathcal{E}$ such that for $y = f(x)+e$, we have that
\[
\begin{aligned}
        \diam(F_y) < d(x,\phi(y)), \text{ and } \phi(y) \subset \mathcal{M}_1,
\end{aligned}
\]
     then the reconstruction $\phi(y)$ is \textbf{not consistent}.
\end{proposition}
\begin{remark}[Upper Bound on Hallucination and Detail Size]\label{rem:sizebound}
Proposition \ref{prop:sizenogo1} proves that there exists an upper bound on the hallucination size: for a consistent decoder $\phi$ that hallucinates a detail $x_{det}$ on $x \in \cM_1$, the size of the detail is bounded by
\[
\begin{aligned}
\|x_{det}\| = \|x+x_{det}-x\| \leq  d(\phi(y),x) \leq \diam(F_y),
\end{aligned}
\]
\noindent for $y = f(x)+e_0$
and $e_0 \in \mathcal{V}$ from part $(ii)$ of Definition \ref{def:detailtrans}. In most cases, the exact upper bound can only be computed knowing the complete model set $\cM_1$. In many data-driven applications, only an approximation from below to the diameter of the feasible set $\diam(F_y)$ can be computed. This approximation is obtained using finite samples of the model set $\mathcal{M}_1$. The better the available finite-sample data set represents the model set, the closer the approximation approaches an upper bound. In Theorem \ref{thm:cvg_diam} we provide conditions that guarantee this convergence. In practice, it is therefore often not possible to use $\diam(F_y)$ directly as a point-wise upper bound.
In particular, it cannot directly bound the magnitude of a hallucinated detail.
\end{remark}
\noindent Remark \ref{rem:sizebound} is a point-wise statement and Proposition \ref{prop:nogo_Fy} gives a data set-wide guarantee against hallucinations larger than the worst-case kernel size.
\begin{proposition}[The Worst-Case Kernel Size Guarantees No Larger Hallucinations]\label{prop:nogo_Fy}
 Let\newline $(F,\mathcal{M}_1,\mathcal{E})$ be a forward problem. Let $\phi: \mathcal{M}_2 \drarrow \mathcal{X}$ be a consistent decoder for the corresponding inverse problem \eqref{eq:sampling1}. Suppose that $\phi(\cM_2) \subseteq \cM_1$. Then, 
 \begin{itemize}
 \item[(i)] It satisfies : 
 \begin{align}\label{eq:worstcoptm}
 \sup_{(x,e)\in \mathcal{M}_1\times\mathcal{E}}d(x,\phi(F(x,e))) \leq \operatorname{Kersize}(F,\mathcal{M}_1, \mathcal{E}).
 \end{align}
 \item[(ii)] It \textbf{does not transfer details} of size $\eta >\operatorname{Kersize}(F,\mathcal{M}_1, \mathcal{E})$.
 \end{itemize}
\end{proposition}
\noindent Proposition \ref{prop:nogo_Fy} provides a guarantee that hallucinations larger than the worst-case kernel size do not occur in reconstructions of a decoder satisfying \eqref{eq:worstcoptm}. In contrast, Theorem \ref{thm:suff_hall} provides conditions that need to be falsified to avoid hallucinations. This means that if the conditions are verified, hallucinations of a certain size will occur with certainty. 
\begin{theorem}[Transfer of Small Details is Inevitable]\label{thm:suff_hall}
 Let $\epsilon>0$ and $(F,\mathcal{M}_1,\mathcal{E})$ be a forward problem. Let $\phi: \mathcal{M}_2 \drarrow \mathcal{X}$ be a decoder for the corresponding inverse problem \eqref{eq:sampling1}, such that $\phi(\cM_2) \subset \cM_1$. Let $y \in \cM_2$ with $\diam(F_y)>0$, and let $\eta \in (0, \diam(F_y)/2)$. \textbf{If either of the following is satisfied}
 \begin{itemize}
 \item[(i)]\textbf{(Too Stable Decoders Hallucinate)} the decoder is such that for one $y \in \mathcal{M}_2$ there exists $z \in \phi(y)$ verifying
 \begin{align}\label{eq:bdstable1}
     \sup_{e \in \mathcal{E} \cap (\cM_2- \set{y})} d(\phi(y+e),z)\leq \eta,
 \end{align}
\item[(ii)]\textbf{(Consistent Decoders Hallucinate)} the decoder is consistent, and $\diam(\phi(y)) \leq \eta$,
 \end{itemize} 
 then, there exist $x \in F_y$ and a non-empty set $\mathcal{V}\subset \mathcal{E}$, such that the decoder $\phi$ hallucinates some $x_{det} \in \mathcal{X}$ of size $\eta$ on $x$ for $\mathcal{V}$.  
\end{theorem}
\noindent Theorem \ref{thm:suff_hall} implies that hallucinations of size smaller than half the diameter of the feasible set of a measurement are inevitable for both too stable and consistent decoders. Hence, any decoder with these desirable properties will hallucinate details of size smaller than half the diameter of the feasible set. Incorrect transfer details may pose a risk to the application in question. To avoid them, the diameters of the feasible sets must be reduced and, consequently, also the worst-case kernel size. This requires changing the forward problem $(F,\mathcal{M}_1,\mathcal{E})$. In fact, $(ii)$ of Theorem \ref{thm:suff_hall} is independent of the reconstruction error of the decoder and therefore constitutes a method- and decoder-agnostic limit. It is determined solely by the available data and the forward problem $(F, \cM_1, \E)$. Consequently, it acts as a fundamental performance ceiling that no specific decoder can surpass, whether Bayesian, TV-regularized, or diffusion-based.
\begin{remark}[Training AI Decoders with Jittering Does Not Protect Against Hallucinations]
In particular, part $(i)$ of Theorem \ref{thm:suff_hall} implies that a decoder satisfying \eqref{eq:bdstable1} for some $y \in \cM_2$ will hallucinate details of size arbitrarily close to $\diam(F_y)/2$ from some $x \in F_y$ for some $\mathcal{V}\subseteq \mathcal{E}$. This means that jittering, as in \cite{jittering_theo, jittering_diff,jittering_1991}, during self-supervised and unsupervised training of DNNs does not protect against AI hallucinations. However, training with noise is recommended by \cite{genzel2020solving}, but it may reduce accuracy. A lower accuracy or, equivalently, a higher worst-case error may satisfy the conditions in Theorem \ref{thm:iff_shift} and, thus, avoid AI hallucinations. Specifically, to guarantee the absence of hallucinations, inequality \eqref{eq:bdstable1} and parts $(i)$ and $(ii)$ of Theorem \ref{thm:iff_shift} need to be falsified for a decoder.
\end{remark}
\subsection{Computing Bounds on Incorrect Detail Transfer}\label{par:algs_kersize}
To guarantee convergence to unique limits, throughout this section we assume that $\mathcal{X}$ is equipped with a norm rather than a semi-norm.

\noindent \textit{Preliminary step: approximating the diameter of feasible sets.} The results of $\S$ \ref{sec:det_transfer} establish bounds on the size of incorrectly transferred details. For a reference $x \in \mathcal{M}_1$ and corresponding measurement $y=f(x)+e \in \cM_2$ with $e \in \mathcal{E}$, the size is lower bounded by $\diam(F_y)/2$ and upper bounded by $\diam(F_y)$. Hence, computing approximations of the diameters of the feasible sets is crucial for predicting the occurrence and size of incorrectly transferred details in any decoder's reconstructions. In most data-driven approaches, one has access only to a finite sample data set of the model set $\mathcal{M}_1$.
\begin{align}\label{eq:dataset}
\mathcal{D}^{K, N} = \left((y_k)_{1 \leq k \leq K} \subset \cM_2, (x_n)_{1 \leq n \leq N} \subset \cM_1\right).
\end{align}
\noindent with $K,N \in \mathbb{N}$. We show that the diameter of the feasible set of a given measurement can be approximated using the data set $\mathcal{D}^{K, N}$. The approximation converges from below to the analytical value in the limit $N \to \infty$ under data sampling assumptions that are specified in Theorem \ref{thm:cvg_diam}. Algorithm \ref{alg:feasapp} applies to a finite sample data set and determines for every measurement, which points in the data set belong to the corresponding feasible set.  More specifically, the algorithm input are three sets $(y_k)_{1 \leq k \leq K } \subset \cM_2$, $(x_n)_{1 \leq n \leq N} \subset \cM_1$ and $(y'_n)_{1 \leq n \leq N}=   (f(x_n))_{1 \leq n \leq N} \subset \cM_2$. Algorithm \ref{alg:diamFy} approximates the diameter of the feasible sets with the output of Algorithm \ref{alg:feasapp}.
\begin{algorithm}[h!]
\caption{Computing the approximate feasible set information}\label{alg:feasapp}
\begin{algorithmic}
\Require $(y_k)_{1 \leq k \leq K}$, $(x_n)_{ 1 \leq n \leq N }$,  $(f(x_n))_{ 1 \leq n \leq N }$ 
\State $FA = 0^{N \times K}$
\State $D = 0^{N \times N}$
\For{$n \in \{1,...,N\}$}
    \For{$k \in \{1,...,K\}$}
        \If{$f(x_n) \in y_k + \mathcal{E}$}
            \State $FA_{n,k} = 1$
        \EndIf
    \EndFor
\EndFor
\State $Com = FA(FA)^T$
\For{$n \in \{1,...,N\}$}
    \State $J_n = \{m \in \{1, \dots, N\}: Com_{n,m}=1 \}$
    \For{$m \in J_n$}
        \State $D_{n,m} = \| x_n-x_m \|$
    \EndFor
\EndFor
\State
\Return $FA$, $D$
\end{algorithmic}
\end{algorithm}
\noindent Algorithm \ref{alg:feasapp} outputs the boolean matrix $FA$, which encodes the appartenance to feasible sets: $FA_{n,k} = 1$ if and only if $x_n \in F_{y_k}$. Within Algorithm \ref{alg:feasapp} the matrix $Com$ encodes the information of whether data points belong to a common feasible set: $Com_{n,m} = 1$ if and only if $x_n, x_m \in F_{y_k}$ for some $k \in \{1, \dots K\} $. This matrix allows to avoid the unnecessary computation of distances between points that do not belong to the same feasible set.
\begin{remark}[Application of Algorithm \ref{alg:feasapp} to Implicit Forward Maps]\label{rem:tuples_gen}
    Algorithm \ref{alg:feasapp} requires as input the sets of points $(y_k)_{1 \leq k \leq K}$, $(x_n)_{1 \leq n \leq N}$, and $(f(x_n))_{1 \leq n \leq N}$. If $f$ can be applied explicitly, the values $(f(x_n))_{1 \leq n \leq N}$ can be computed directly from $(x_n)_{1 \leq n \leq N}$. When $f$ is not explicitly defined but one has access to tuples of data-points $(x_n, y'_n) \in \cM_1 \times \cM_2$ such that for every $n \in \set{1 \dots N}$,  $y'_n$ is a measurement from $x_n$, the algorithm can be applied with $(y_k)_{1 \leq k \leq K}$, $(x_n)_{ 1 \leq n \leq N }$,  $(y'_n)_{ 1 \leq n \leq N }$ as inputs. This amounts to defining $f$ on a finite domain by setting $\forall n \in \set{1, \dots, N}, \quad f(x_n) = y_n'$. This definition is valid provided that the points $(x_n){1 \leq n \leq N}$ are distinct.
\end{remark}
\begin{algorithm}[h!]
\caption{Computing the approximate diameters of feasible sets}\label{alg:diamFy}
\begin{algorithmic}
\Require $FA \in \{ 0,1 \}^{N \times K}$, $D \in \mathbb{R}_+^{N \times N}$
\State $d = 0^{K}$
\For{$k \in \{1,...,K\}$}
    $I = \{n \in [1,N], FA_{n,k}=1 \}$
    \State $d_k = \max(D_{I\times I})$
\EndFor\\
\Return $d$
\end{algorithmic}
\end{algorithm}
\noindent The list $d$ outputted by Algorithm \ref{alg:diamFy} contains the diameters of each approximate feasible set. The $k$-th element of the output is the diameter $d_k = \diam(F_{y_k^N})$ and is an approximation from below to the diameter $\diam(F_{y_k})$. Theorem \ref{thm:cvg_diam} shows that for every $y \in \cM_2$, the approximate diameter $\diam(F_y^N)$ converges to $\diam(F_y)$ when the number of samples tends to infinity, under specific data sampling assumptions.
\begin{theorem}[Convergence of the Diameter of Approximate Feasible Sets]\label{thm:cvg_diam}
Let $\mu$ be a probability measure on $\cM_1$. Let $K, N \in \mathbb{N}$, $(X_n)_{n \in \mathbb{N}}$ be independent, identically distributed random variables with the distribution $\mu$, that is $(X_n)_{n\in\mathbb{N}} \sim \mu^{\otimes \mathbb{N}}$, and $(y_k)_{1 \leq k \leq K} \subset \cM_2$. Let $\diam(F_{y_k}^N)$ be the output of Algorithm \ref{alg:diamFy} combined with Algorithm \ref{alg:feasapp} using $(X_n)_{1\leq n \leq N}$, $(y_k)_{1 \leq k \leq K} \in (\cM_2)^K$ and $(f(X_n))_{1\leq n \leq N}$. Assume that for every non empty open set of $\cM_1$, $O$,  we have $\mu(O)>0$. Then, almost surely 
\[
\begin{aligned}
\max_{1 \leq k \leq K} |\diam(F_{y_k}^N)-\diam(F_{y_k})| \rightarrow 0,
\end{aligned}
\] 
as $N \to \infty$ and the convergence is from below: 
\[
\begin{aligned}
        \diam(F_{y_k}^N) \leq \diam(F_{y_k}),
\end{aligned}
\]      
for all $K \in \mathbb{N}$ and $N \in \mathbb{N}$.
\end{theorem}
\noindent Convergence rates for Theorem \ref{thm:cvg_diam} can be obtained under additional technical assumptions as described in the Appendix. Since the convergence is from below, the sufficient condition for detail transfer in part $(i)$ Theorem \ref{thm:suff_hall} can only be falsified approximately from below. The Theorem \ref{thm:suff_hall} implies that any decoder  satisfying the desirable assumptions of Theorem \ref{thm:suff_hall} cannot resolve details of size smaller than half the approximate diameter without hallucinating. As a consequence, half the approximate diameter of the feasible set, $\diam(F_y^N)/2$, is a decoder-agnostic lower bound on the error. The approximate diameter $\diam(F_y^N)$ should not be interpreted as an upper bound to the transferred detail size, but rather as order of magnitude of the how large the hallucination size and worst-case error can be. We provide empirical evidence of this relation in our experiments in $\S$ \ref{sec:experiments}. 
\begin{remark}[Discussing the Assumptions on the Distribution $\mu$] 
The convergence result in Theorem \ref{thm:cvg_diam} holds if, for every non-empty open set $O$ of $\cM_1$, we have $\mu(O)>0$. If the assumption is not verified, then there exist $x \in \cM_1$ and $r>0$ such that $\mu( {B}^o(x,r) \cap \cM_1) = 0$. In practice, this means that if points are sampled in $\cM_1$ according to $\mu$, then with probability $1$ there will be none in $ {B}^o(x,r)$. In other words, ${B}^o(x,r)$ can be excluded from $\cM_1$  considering $\cM_1' = \cM_1 \setminus  {B}^o(x,r) $. Repeating this process for every open ball that does not satisfy the property yields a model set for which the property holds.
\end{remark}
\section{Methods for Anticipating AI Hallucinations}\label{sec:methods}
\begin{figure}[h!]
    \centering
    \includegraphics[width=\textwidth]{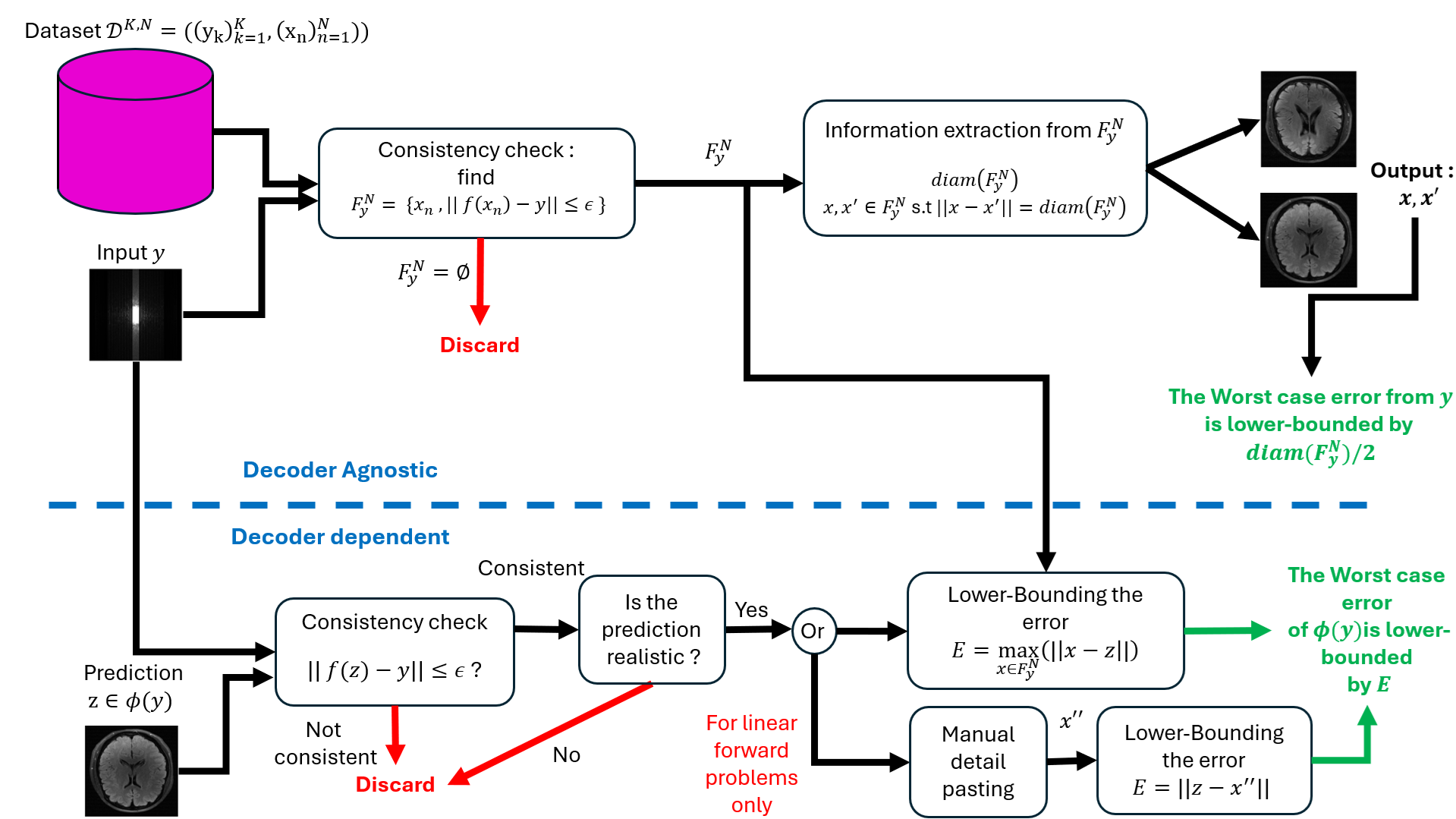}
    \caption{Methods to anticipate and detect hallucinations in this paper.}
    \label{fig:method}
\end{figure}

In this section, we present two methods to assess the trustworthiness of a decoder's reconstructions based on the results and the framework provided in $\S$ \ref{sec:theory}. Hallucinations are partly hard to detect because there is usually no reference image (or ground truth) available. Hence, we first provide a \textit{decoder-agnostic} method based on a data set. This method evaluates whether the reconstruction of any decoder from an input $y \in \cM_2$ can be trusted without requiring knowledge of the ground truth. The data set $\mathcal{D}^{K,N}$ is defined in \eqref{eq:dataset}. The decoder-agnostic method requires large data sets such that the feasible set of a new input is rarely empty. The method allows to anticipate the minimal decoder-independent worst-case error on a new reconstruction. The corresponding pipeline is illustrated in Figure \ref{fig:method}.\newline

\subsection{Decoder-Agnostic Method}
The upper half of Figure \ref{fig:method} provides a scheme of the \textbf{decoder-agnostic} method. This method part requires the data set, a new measurement $y \in \mathcal{Y}$ and a score to compare pairs of points in $\X$. This method first approximates the feasible set of $y$ with the empirical feasible set $F^N_y$ computed using Algorithm \ref{alg:feasapp}. If this set is empty, the method cannot conclude about the hallucinations in the input, and the algorithm terminates. Otherwise, with Algorithm \ref{alg:diamFy} we obtain the two data points $x,x' \in F_y^N$ that are the most distant to each other according to a chosen semi-norm. The results of $\S$ \ref{sec:theory} ensure that for any decoder $\phi$ and for any norm $\|\cdot \|$, there exists a valid reference point $x \in F_y$ such that $d( x,\phi(y)) \geq \frac{d(x,x')}{2} = \frac{\|x-x'\|}{2}$. A human in-the-loop should also inspect the points $x$ and $x'$, and the method enables evaluating whether there can exist a realistic reconstruction $\phi(y)$ matching both $x$ and $x'$ while being consistent according to the Definition \ref{def:consistent} of consistent decoders. 
\begin{remark}
    Note that the success of the decoder-agnostic method on a given inverse problem depends on the ability to find non-empty feasible sets $F_y^N$. On some given raw data sets, there may rarely exist measurements $y$ for which the method can find non-empty feasible sets, as is the case for the FastMRI data set described in $\S$ \ref{sec:MRI}. In other cases, the structure of the forward model can be leveraged to construct such elements even from relatively small raw data sets, as in the S2SR case of super resolution of Sentinel-2 data (S2SR) in $\S$ \ref{sec:metrics}.
\end{remark}

\subsection{Decoder-Dependent Method}
\noindent The lower half of Figure \ref{fig:method} provides a scheme of the \textbf{decoder-dependent} method. This method makes it possible to assess whether given details of a decoder's reconstruction can be trusted. This part of the method only applies to problems with a linear forward model. It first checks whether the reconstruction is consistent with the input. In a non-automated process, it is also possible to inspect whether the reconstruction is realistic (that is, in the model set $\cM_1$). After these checks, the method evaluates whether a given detail in the reconstruction $\phi(y)$ can be replaced with another chosen one. Crucially, the replacement detail is chosen by the user and made realistic. For example, in satellite imagery, the method answers questions such as: "Is that building in the reconstruction invented or existing?" or "Is it possible that the decoder missed a farm in the middle of this green area?". In MRI scan acceleration, possible questions are "Is this cyst artifact invented by the decoder or actually present?" or "Did the decoder miss a tumoral structure at the place of this unstructured artifact?". In super-resolution of MNIST digits, possible questions are "Did the decoder miss the detail turning the 3 into an 8?".

\begin{figure}[h!]
    \centering
    \includegraphics[width=\textwidth]{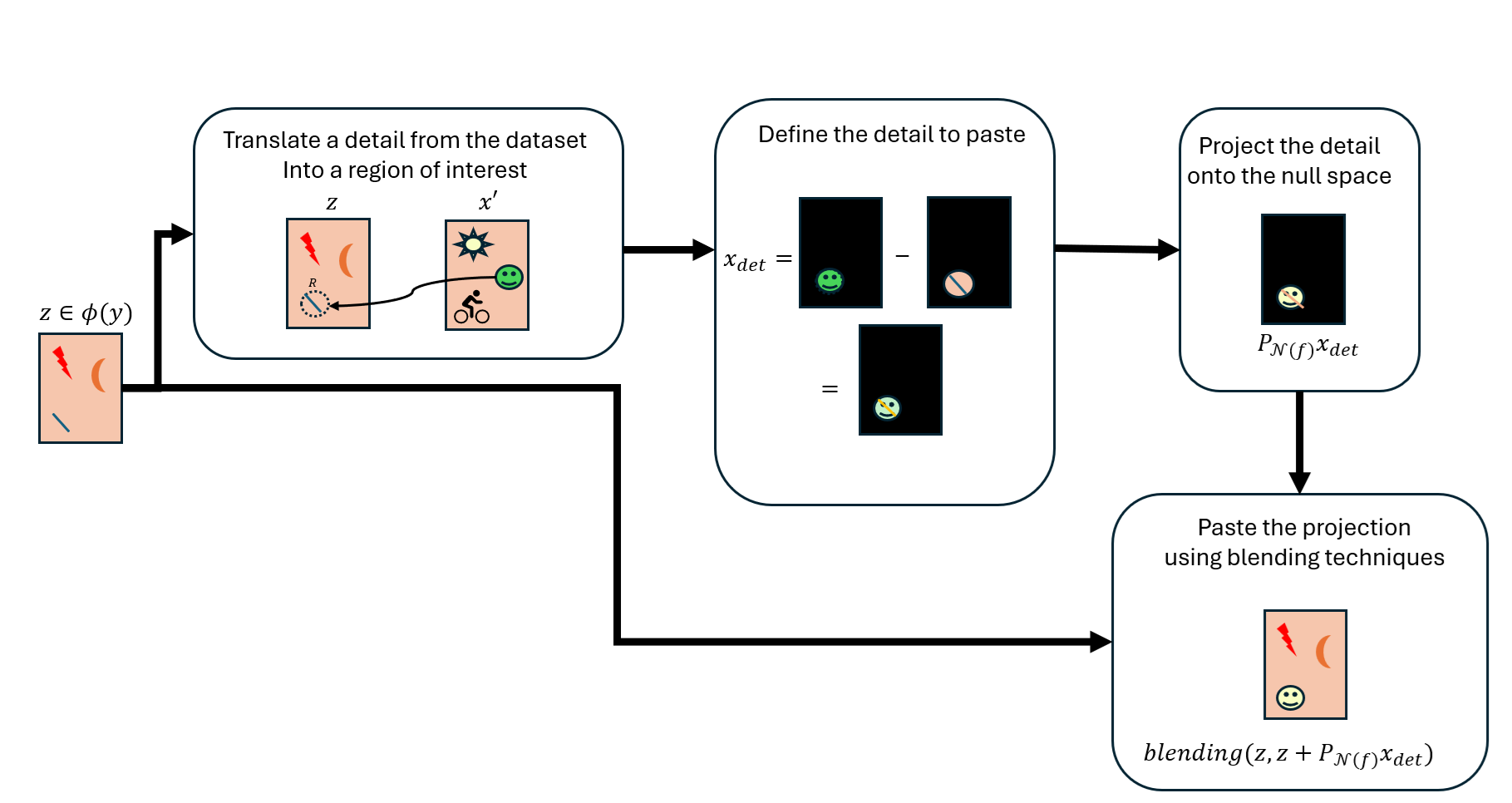}
    \caption{Manual detail pasting method for linear forward models with additive noise, illustrated for inverse problems in imaging.}
    \label{fig:kerdraw}
\end{figure}

\noindent A use case for this method is a scenario in which a radiologist (or an equivalent in satellite imagery) has access to measurements $y \in \cM_2$ and a reconstruction from a decoder $z \in \phi(y)$. The user asks whether no important detail (such as an anomaly) has been added (or equivalently, missed) in the reconstruction $z \in \phi(y)$. To that aim, we propose to use the available data and follow the following procedure to verify whether a detail found in the data set can be missed or added on a given region of interest $R$ of the reconstruction $z \in \phi(y)$:
\begin{itemize}
    \item First check whether the reconstruction is realistic (in the model set $\cM_1$) and consistent with $y$ according to the Definition \ref{def:consistent} of consistent decoders. If not, the reconstruction is inaccurate.
    \item Find a detail of interest in $x'$ of the data set and translate it to the region of interest $R$. After translation, define the detail $x_{det}$ as $x'-z$ on $R$ and $0$ outside of $R$.
    \item Stick the point $P_{N(f)}x_{det}+ z$ to $z$ using a blending method, such as $\alpha$ blending for images \cite{deVries2026Blending} to hide the borders of the detail with a smooth transition. 
\end{itemize}
\noindent The described procedure generates a point $x''$ (typically an image in imaging) that is consistent with $y$, given that the original reconstruction $z \in \phi(y)$ is consistent. This statement is valid because the detail $x_{det}$ is projected onto the null space before being pasted. Note that due to the use of blending techniques, the detail added does not belong strictly to the kernel $\mathcal{N}(f)$, in contrast to the projection of the detail onto the kernel $P_{N(f)}x_{det}$. In practice, residual artifacts in the measurement space resulting from the application of blending methods are generally comparable to noise when the techniques are well designed and used.
\noindent Then, there are the following possible situations relevant for the user, such as the radiologist for MRI or analyst of satellite imagery: 
\begin{itemize}
    \item[(1)] If $x''$ is realistic, it splits into the two following possibilities. Either $x''$ semantically differs from $z$ and every point in $\phi(y)$. Then, $x''$ is a plausible ground truth and the decoder hallucinates $z-x''$ on $x''$. Or $x''$ remains semantically close to $z$. In that second case, the detail is essentially outside of the null space and $z-x''$ cannot be hallucinated in $x''$ by the decoder; the predictions can be trusted regarding the considered detail.
    \item[(2)] If $x''$ is not realistic, then the kernel projection highly degraded the detail $z-x''$ and it cannot be hallucinated on $x''$ by the model.
\end{itemize}
\noindent Note that the decoder-dependent method allows to \textit{add} a detail on the point $z$ by selecting it on another point $x'$ and moving it to a flat area on $z$ (a region where the component values values vary very little across neighboring positions), but also to \textit{remove a detail} by selecting a flat area on $x'$ and moving it to a detail on $z$.
\noindent Finally, note that the manual detail pasting method described in Figure \ref{fig:kerdraw} only works when the forward model is known and linear.
\section{Experiments}\label{sec:experiments}
In $\S$ \ref{sec:exp_desc} we describe the different types of experiments carried out in this paper. In $\S$ \ref{sec:MNIST}, $\S$ \ref{sec:MRI} and $\S$ \ref{sec:S2SR}, we present the results of the experiments for three applications with common principles and some particularities for each application. In addition, we show applications of methods presented in $\S$ \ref{sec:methods} to detect or anticipate hallucinations.
\subsection{Experiments Description and Setup}\label{sec:exp_desc}
In this section, we tackle three inverse problems in imaging. We first present metrics for detail transfer detection specific to imaging.
\subsubsection{Metrics for Detail Transfer Detection}\label{sec:metrics}
In this section, we present a family of computationally easy Image Quality Assessment (IQA) semi-positive scores on $\mathcal{X}$ that capture hallucinated details when a reference image and a reconstruction are provided. The scores they must satisfy positivity, symmetry, and the triangle inequality.
One pitfall of the root mean squared error (RMSE), that is the $\ell^2$ distance on $\mathcal{X}$, is that it often fails to capture local details in a larger image \cite{fastmri20}. One identified reason is that the error in the detail averages out when summing the terms across the whole image. To support the point, \cite{morshuis2022adversarial}, obtain hallucinations when the adversarial attack, based on norm gradient ascent, is restricted to a specific region of interest. To isolate details, we propose to use $\ell^q$ norms on a specified region of interest $R$ to compare two points in $\mathcal{X}$ in the following metric : 
\[
\begin{aligned}
    \| x-x' \|_{p,q, \mathcal{R}} = \Big(\sum_{R \in \mathcal{R}} \| x-x' \|_{q, R}^p\Big)^{1/p} ,
\end{aligned}
\]
where $\mathcal{R}$ is finite, each of its elements $R$ is a subpart of the image, $\| .\|_{q,R}$ denotes the $\ell^q$ norm restricted to $R$, for $R \in \mathcal{R}, q \geq 1, p\geq 1$.
The choice of $\mathcal{R}, p, q$ can be adjusted based on the application. For example, taking $\mathcal{R}$ with only $1$ element allows to focus on one predetermined region of interest. Allowing $\mathcal{R}$ to have multiple elements distributes the focus across multiple parts of an image. Meanwhile, the parameter $p$ can be used to adjust the relative importance of the detail's amplitude and spatial size. In our experiments, we choose $\mathcal{R}$ with a single element.
These metrics apply to problems where the points of $\mathcal{X}$ are images, but they can be extended to more general settings, such as wavelet function representations. Detailing these generalisations is out of the scope of this study.
Note that if $\cup_{R\in \mathcal{R}}R$ does not cover the whole image, then $\|.\|_{p,q, \mathcal{R}}$ is a seminorm and not norm. In that case, the guarantees from $\S$ \ref{sec:det_transfer} and $\S$ \ref{par:bounds_dettrans} still hold, but the convergence theorems from $\S$ \ref{par:algs_kersize} do not apply due to the specific topology induced by a seminorm.  
\subsubsection{Verification of the Definition of Hallucinations}\label{sec:exp_verif_setup}
In this experiment, we are given an input $y \in \cM_2$, and we define two points in its feasible set : $x, x+ x_{det} \in F_y$. We sample a finite number $M$ of noise vectors $ \set{e_m,\quad 1 \leq m \leq M}$ and consider the resulting decoder's reconstructions $\set{\phi(y+ e_m), \quad 1 \leq m \leq M}$. We ensure that the detail $x_{det}$ is defined such that $x+x_{det} \in \phi(f(x)+e_0)$ for some $e_0 \in \E$. After verifying the second inclusion, we determine the values of $\eta$ for which the decoder $\phi$ hallucinates details $x_{det}$ of size $\eta$ on the image $x$. This is done for a subset of noise $\V \subseteq  \set{y+e_m-f(x),\quad 1 \leq m \leq M} \subset \E $. In this setup, these values form an interval $[\eta_{min}(\V), \eta_{max}(\V)]$ defined by the \textit{minimum hallucination size} $\eta_{min}(\V) = \max_{e \in \V} d(x + x_{det}, \phi(f(x)+ e))$ and the \textit{maximum hallucination size} $\eta_{max}(\V) = \min_{e \in \V} d(x, \phi(f(x)+e))$.  Note that this interval can be empty because $\eta_{min}(\V)$ can be larger than $\eta_{max}(\V)$. When the decoder $\phi$ is single valued we have $\eta_{max}(\V) \leq \| x_{det} \|$. When $\eta_{min}(\V) \leq  {\|x_{det}\|}/{2}$  Theorem \ref{thm:iff_shift} also applies with the same quantities. They are calculated and compared in cases where the decoder hallucinates and when it does not. In the following, we will most often omit the dependence of $\eta_{min, max}$ in $\V$ to lighten the notations.
\subsubsection{Quantitative Bounds Assessment}\label{sec:exp_quant_setup}
In this experiment, we consider a set of inputs $y \in \cM_2$ for a decoder $\phi$. For each of them, we consider the given elements in the feasible set of $y$: $\set{x_1, \dots, x_N}: = F_y^N$. The method used to find them varies between the applications. Each reconstruction $\phi(y)$ here is single-valued; in the case of stochastic decoders, we only consider the first one as reconstruction $\phi(y)$. After a visual check on a sample of reconstructions, we consider that the reconstructions are realistic (in $ \cM_1$). In this setting, the \textit{worst-case error} $\sup_{u \in F_y^N} \|u- \phi(y) \| $ is lower-bounded by $\diam(F_y^N)/2$ and upper-bounded by $\diam(F_y^N \cup \set{\phi(y)})$. If  $F_y^N$ approximates the feasible set $F_y$ well enough, the \textit{worst-case error} can be smaller than $\diam(F_y^N)$, but this is not guaranteed. Both Theorem \ref{thm:suff_hall} and Proposition \ref{prop:nogo_Fy} can be applied for $\V$ being a singleton, and we assess the sharpness of the bounds. In practice, we calculate the \textit{worst-case error} as $\max(\| \phi(y)-x \|, \| \phi(y)-x' \|)$ where $x,x' \in F_y^N$ are defined such that $\diam(F_y^N) = \| x-x' \|$ and Theorem \ref{thm:suff_hall} should still hold.
\subsection{Super Resolution of MNIST Digits}\label{sec:MNIST}
The MNIST data set \cite{MNIST} is a standard benchmark for image classification and pattern recognition, particularly in the context of handwritten digit recognition. It consists of $70,000$ grayscale images of handwritten digits representing the classes $0$ through $9$. Each image has a resolution of $28 \times 28$ pixels and is stored in a normalized, centered format, with mean $0.1307$ and variance $0.3081$, in order to reduce variability due to translation and scale. The data set is split into $60,000$ training images and $10,000$ test images, each accompanied by a label indicating the corresponding digit class.

\noindent In this section, we address the problem of super-resolution of downsampled MNIST images. The associated forward operator $f$ performs downsampling and is linear. It is composed of a Gaussian pooling with standard deviation $3$ followed by a mean pooling with a scale factor of $3$. The forward model is with additive noise : $F(x,e) = fx+e$. Taking into account the intermediate padding operations, the resulting downsampled images have a spatial resolution of $12 \times 12$. Note that, without incorporating $\mathcal{M}_1$ and $\mathcal{E}$, the inverse problem can be ill-posed, since the forward model is linear with additive noise and has a nontrivial null space.
Optionally, noise can be added to the downsampled images. We model this noise using a Poisson sampling scheme: each pixel intensity follows the distribution $\alpha \mathcal{P}(i/\alpha)$, where $i$ denotes the intensity of the noiseless image. The parameter $\alpha$ is chosen so that the expected $\ell^2$ norm of the noise vector matches a prescribed noise level in $\set{0.3, 0.6}$, which is negligible with respect to the $\ell^2$ norm of usual images. This noise level can be compared to the typical distance between nearby low-resolution images in the MNIST data set. Using Algorithm \ref{alg:feasapp}, we observed that the $\ell^2$ distance between the closest low-resolution images lies between $0.2$ and $0.4$.
Finally, throughout this section, all considered decoders are assumed to be single-valued. Under this assumption, for any input $y \in \mathcal{Y}$, the decoder output satisfies $\phi(y) \in \mathcal{X}$.
\subsubsection{Ill-Posedness of a Problem Causes Hallucinations}\label{par:exp_VDSR_MNIST}
Following the results of $\S$ \ref{sec:main}, the size of hallucinations is constrained by the diameter of a feasible set $\diam(F_y)$ of a given input $y$ and by the \textit{worst-case kernel size} $\operatorname{Kersize}(F, \cM_1, B^o(0, \epsilon))$ for $\epsilon>0$ being the noise level.  When the noise level increases, the decoder-independent bounds on the hallucination magnitude increase, leading to larger and more frequent hallucinations, because the quantities $\diam(F_y)$ and $\operatorname{Kersize}(F, \cM_1, B^o(0, \epsilon))$ are non-decreasing with respect to $\epsilon$. 

\noindent To validate these results,  we train the same VDSR decoder introduced in \cite{VDSR2016}  with an added noise of level $\epsilon \in \set{0,0.3, 0.6}$\footnote{$\epsilon = 0 $ is theoretically not possible because we assume $\E$ to be open. But we consider instead that $\epsilon = \epsilon_M/2$ where $\epsilon_M$ is the machine precision}. 
The architecture of the VDSR decoder consists of 64 feature channels and 20 residual blocks, each comprising two 2D convolutions with an intervening ReLU activation.

\noindent Training is performed on the $60,000$ images from the initial training set using a 0.7/0.3 train-validation split. The digits $1,3$ and $4$ are removed from this set. We consider that these withdrawn digits remain in distribution (in the model set $\cM_1$). This reflects real-world applications in which the full distribution (or model set) may not be fully represented during training, but can be inferred from the training set. Predicting the missing digits in the training set is then part of a decoder's generalisation ability. The train-validation split ratio is  0.7/0.3. We employ the Adam optimiser \cite{ADAMOpti2017} with a learning rate of $2 \times 10^ {-4}$ for $300$ epochs to train the decoders. 

\begin{figure}[h!]
\centering
\setlength{\tabcolsep}{1pt}        
\renewcommand{\arraystretch}{0.9}  
\begin{tabular}{|cc|ccc|}
\hline
& & & SR $\phi_{\epsilon}(y)$ & \\
LR image $y$ & HR image $x$ & $\epsilon = 0$ &  $\epsilon = 0.3$ &  $\epsilon = 0.6$ \\
\includegraphics[width=0.16\textwidth]{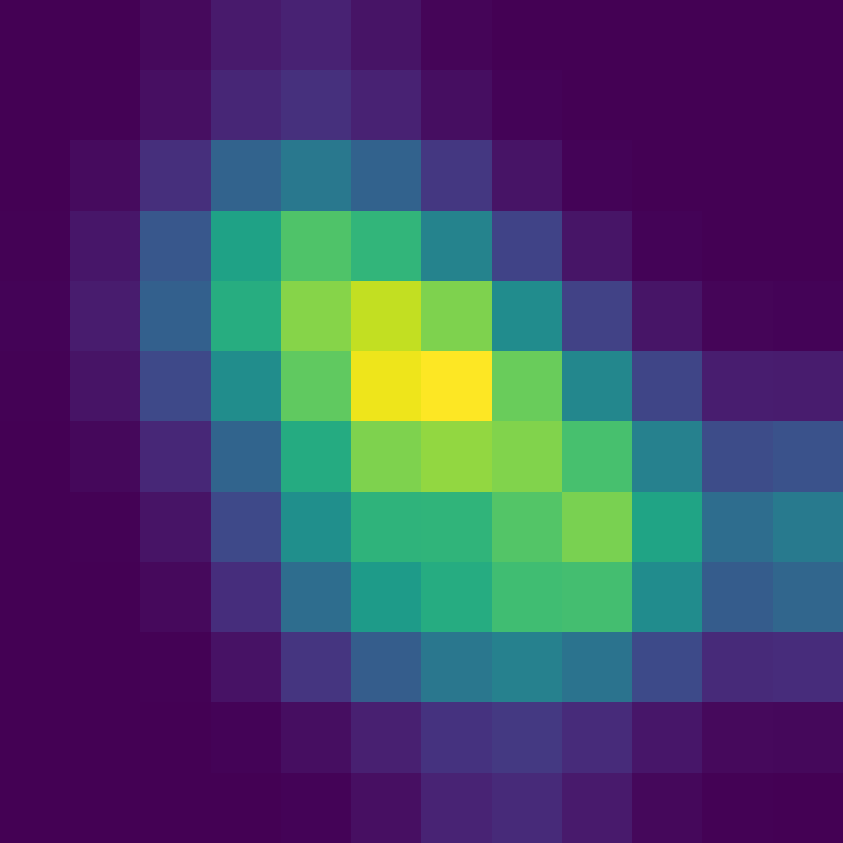} &
\includegraphics[width=0.16\textwidth]{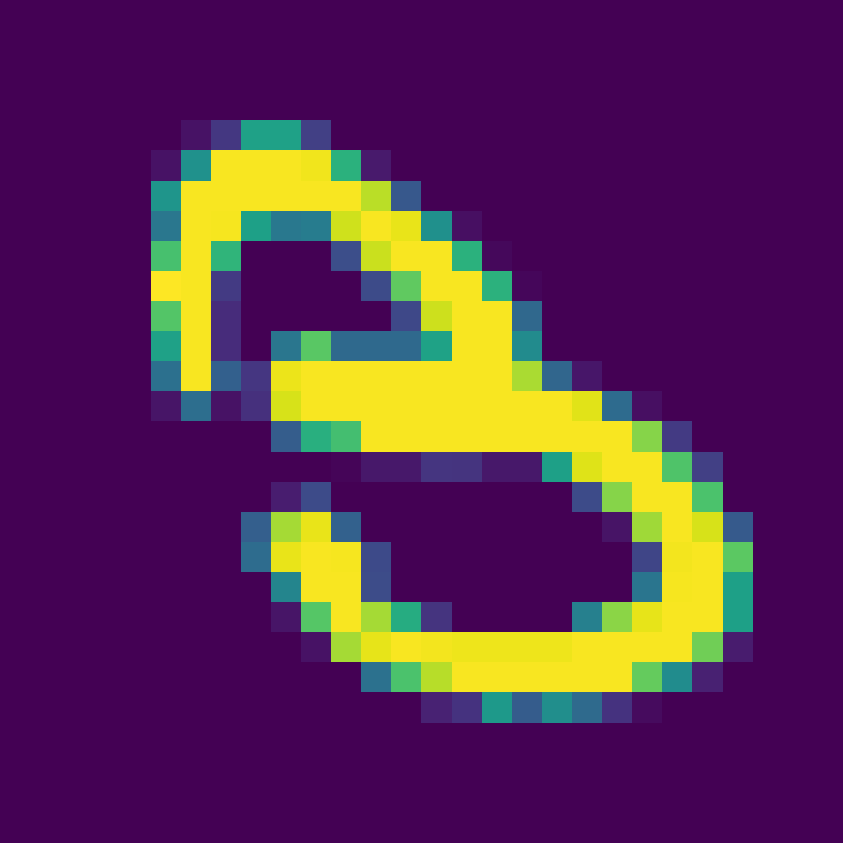} &
\includegraphics[width=0.16\textwidth]{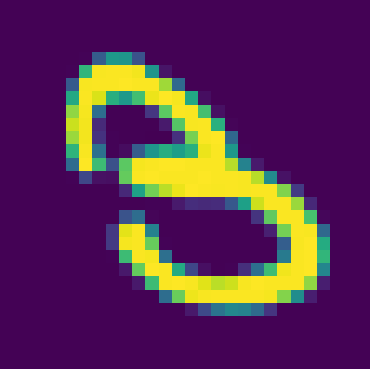} &
\includegraphics[width=0.16\textwidth]{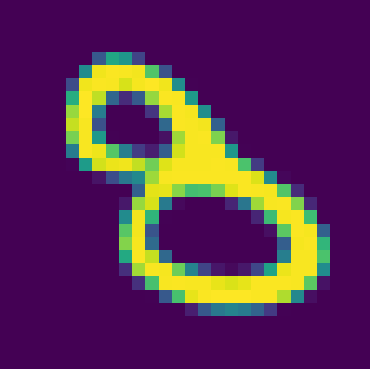} &
\includegraphics[width=0.16\textwidth]{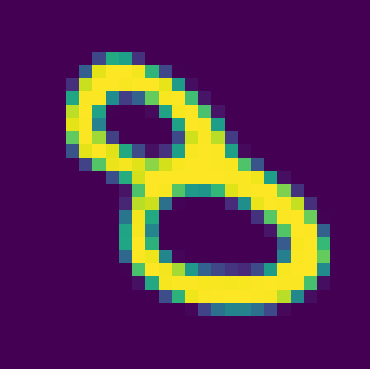}\\[-2pt]
\includegraphics[width=0.16\textwidth]{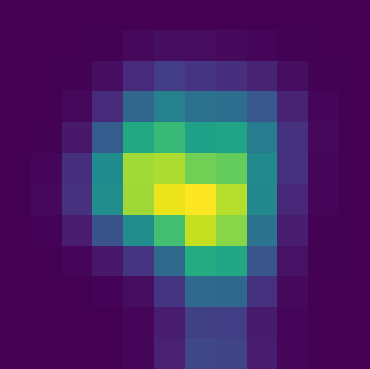} &
\includegraphics[width=0.16\textwidth]{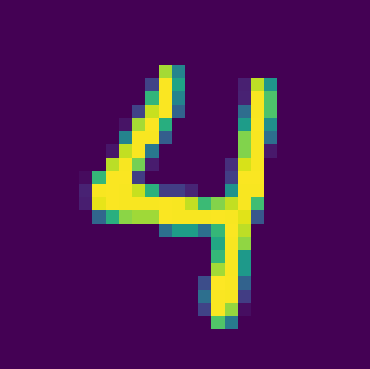} &
\includegraphics[width=0.16\textwidth]{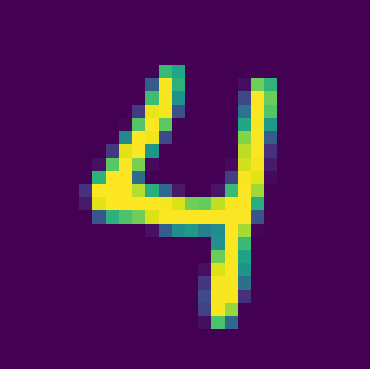} &
\includegraphics[width=0.16\textwidth]{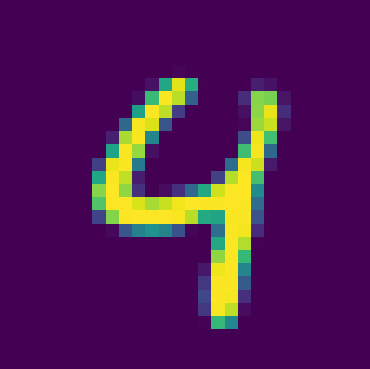} &
\includegraphics[width=0.16\textwidth]{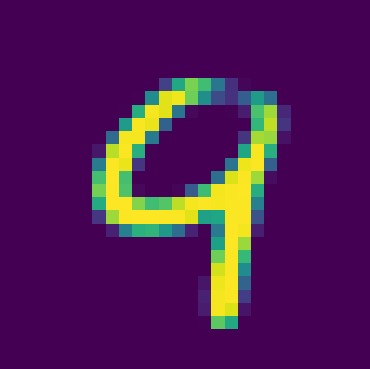}  \\[-2pt]
\includegraphics[width=0.16\textwidth]{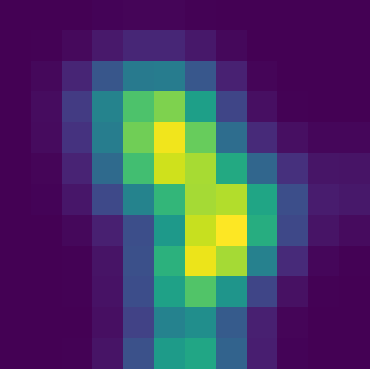} &
\includegraphics[width=0.16\textwidth]{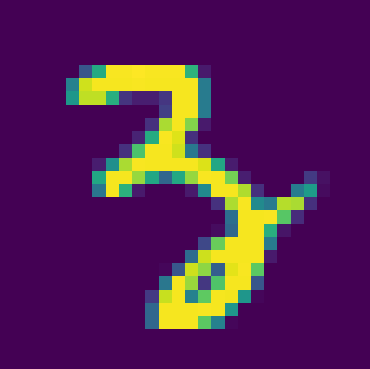} &
\includegraphics[width=0.16\textwidth]{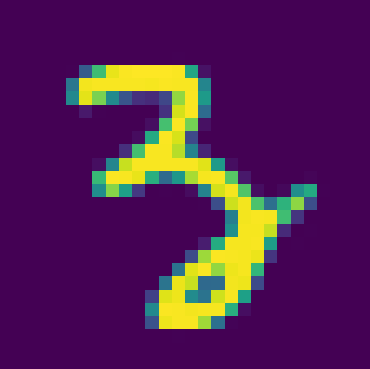} &
\includegraphics[width=0.16\textwidth]{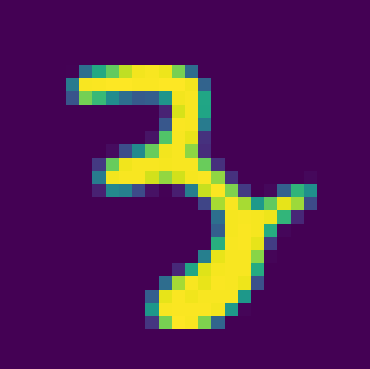} &
\includegraphics[width=0.16\textwidth]{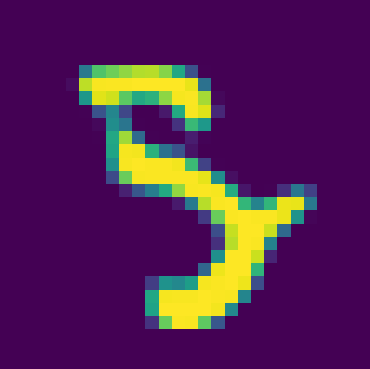}  \\[-2pt]
\multicolumn{2}{|c|}{$\operatorname{Kersize}(F, \cM_1, \E)$} &
0.00 & 0.27 & 0.45 \\
\hline
\multicolumn{2}{|c|}{
  {\centering  Hallucinated/Ambiguous/Correct}
} & $0/0/100$ & $6/11/83$ & $13/15/72$ \\
\hline
\end{tabular}
\caption{The number of hallucinations increases with the worst-case kernel size: Super-resolution results for MNIST digits using VDSR decoders $\phi_{\epsilon}$ trained with three different noise levels $\epsilon \in \{0, 0.3,0.6\}$. Each row displays a unique digit from the data set. Low resolution (LR) denotes the input measurement, high resolution (HR) represents ground truth, and super-resolved (SR) indicates the decoder's reconstructions. All reconstructions generated by decoders trained with additive noise are consistent according to the Definition \ref{def:consistent} of consistent decoders. The worst-case kernel size is computed with $\cM_1$ consisting of the MNIST training data set including all digits and $\|\cdot \|$ is the $\ell^1$ norm on the whole image. The last row indicates the number of hallucinated reconstructions among the first $100$ images of the MNIST test data set for each trained decoder. These values were obtained through individual visual inspection. The two last reconstructions of the second row are examples of "ambiguous" hallucinations.}
\label{fig:MNIST_examples}
\end{figure}

\noindent To assess the three decoder's behaviour qualitatively, we look into their reconstructions on the $100$ first images of the test set. The results are reported in Figure \ref{fig:MNIST_examples}, showing $3$ examples and the number of seen hallucinations on the $100$ first reconstructions, compared to the worst-case kernel size calculated using the full training set, including the digits $1,3,4$.
\begin{itemize}
    \item The decoder trained without noise produces almost perfect reconstructions without hallucination on the $100$ first images of the test set. This can be explained by the fact that the associated worst-case kernel size is null. Indeed, high resolution (HR) images $x,x'$ of the $MNIST$ data set are in a common feasible set if and only if $x-x' \in \mathcal{N}(f)$, or equivalently, $fx = fx'$. There is no pair of points with strictly equal low-resolution (LR) versions in the MNIST data set. Consequently, the learned problem is well-posed if $\cM_1$ is restricted to the training set. Despite the difference between some LR images being visually small or imperceptible, they cannot be interpreted as noise under the noiseless forward model. In this case, $\operatorname{Kersize}(F, \cM_1, B^o(0, \epsilon)) = 0$ and therefore, the results of $\S$ \ref{sec:main} tell that the reconstruction accuracy is only limited by the expressivity of the VDSR decoder. Moreover, the accuracy of the reconstructions on digits $1,3,4$ (not included in the training procedure) confirms that they can be considered in $\cM_1$.
    \item The decoder trained with a $0.3$ noise level adds noise hallucination to some images, but less than and not as badly as the one trained with a $0.6$ noise level, as expected from $\S$ \ref{sec:main}.
\end{itemize}
\subsubsection{Verification of the Definition of Hallucinations}\label{par:verif_def_MNIST}
\begin{figure}[h!]
\centering
\renewcommand{\arraystretch}{1.1}
\begin{table}[H]
    \centering
    \setlength{\extrarowheight}{2pt}
    \begin{tabular}{|m{2.6cm}|*5{>{\centering\arraybackslash}m{2cm}|}}
        \hline
        & \textbf{HR image} ($x$) & \textbf{SR image} ($x+x_{det}$) & \textbf{Average reconstruction} & \textbf{3 standard deviations} & \textbf{ROI} \\
        \hline
        \small{$|\V| = 51$} \newline \small{$\eta_{min}(|\V|) = 0.36$} \newline \small{$\eta_{max}(|\V|) = 0.85$} \newline \small{$\| x_{det} \| = 0.85$} &
        \includegraphics[width=2cm]{img18_HR.png} &
        \includegraphics[width=2cm]{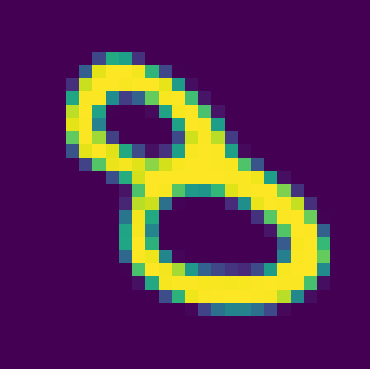} &
        \includegraphics[width=2cm]{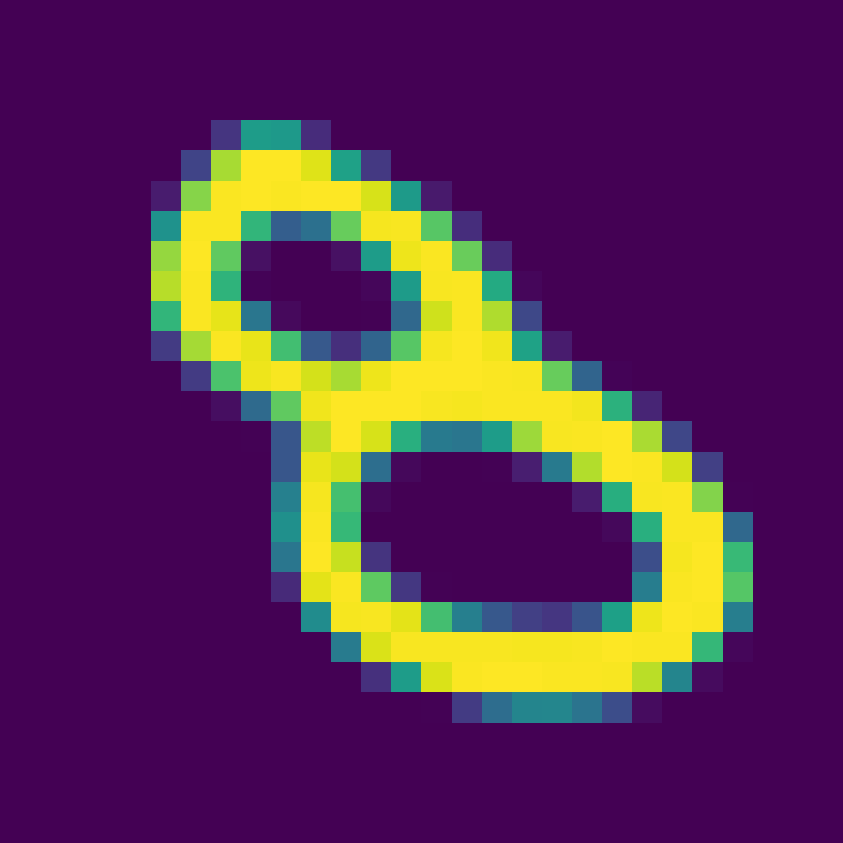} &
        \includegraphics[width=2cm]{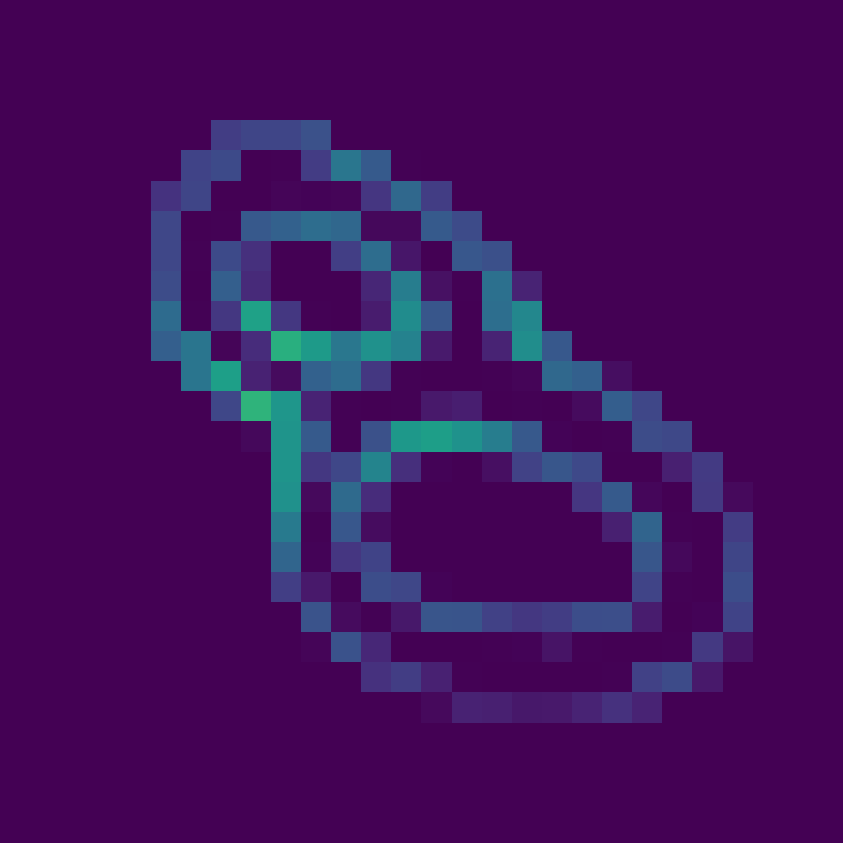} &
        \includegraphics[width=2cm]{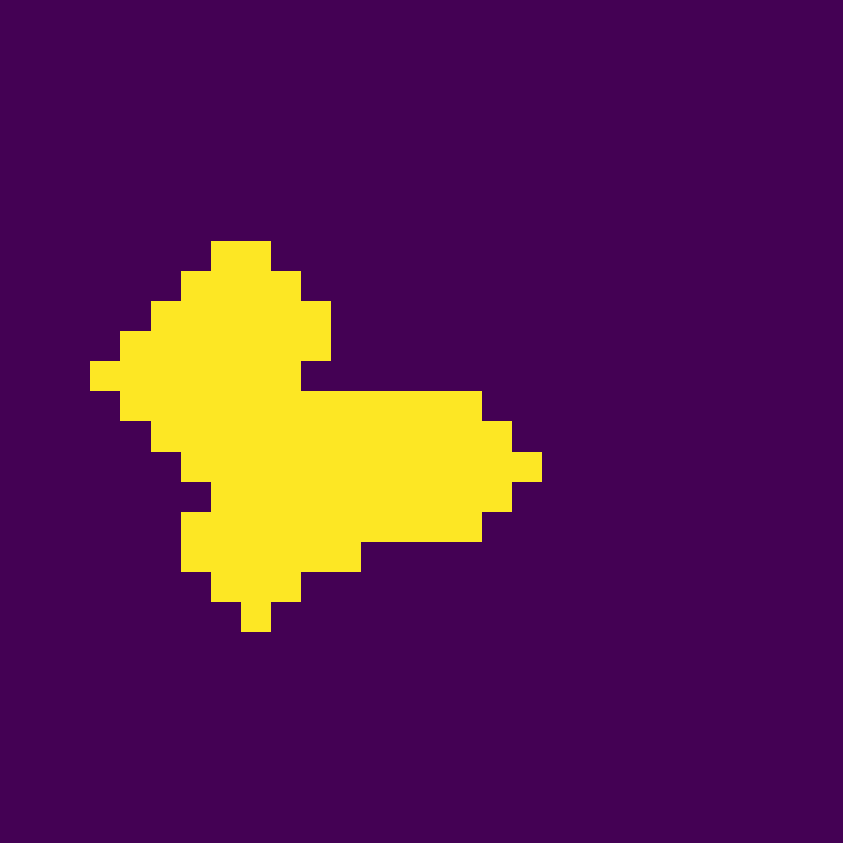} \\
        \hline
        \small{$|\V| = 51$} \newline \small{$\eta_{min}(|\V|) = 0.70$} \newline \small{$\eta_{max}(|\V|) = 0.90$} \newline \small{$\| x_{det} \| = 0.90$} &
        \includegraphics[width=2cm]{img27_HR.png} &
        \includegraphics[width=2cm]{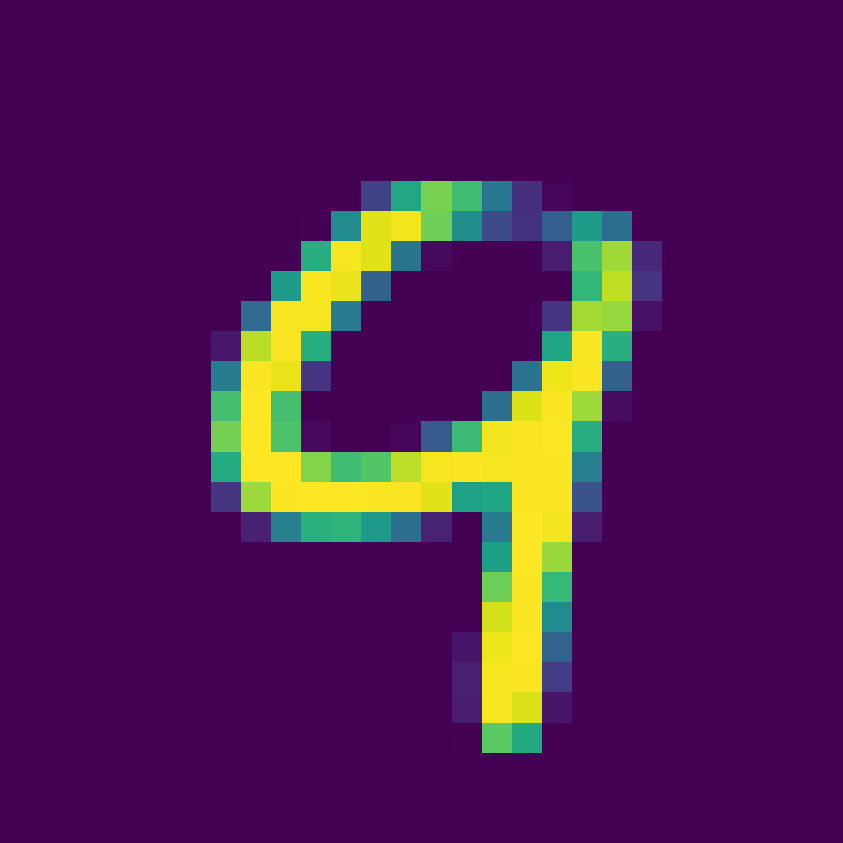} &
        \includegraphics[width=2cm]{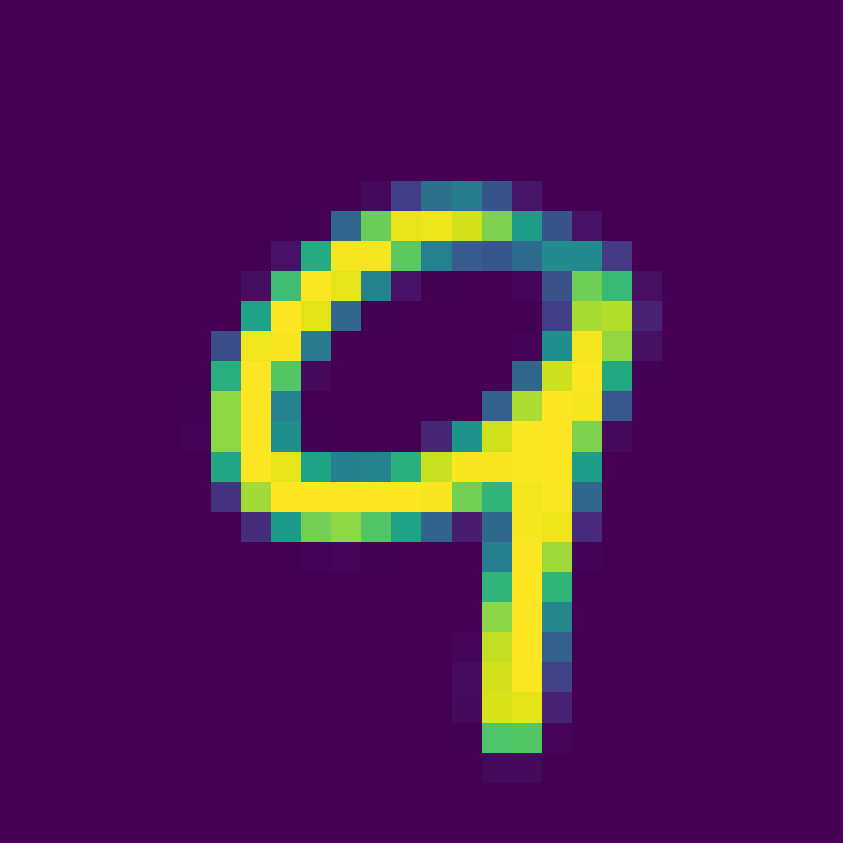} &
        \includegraphics[width=2cm]{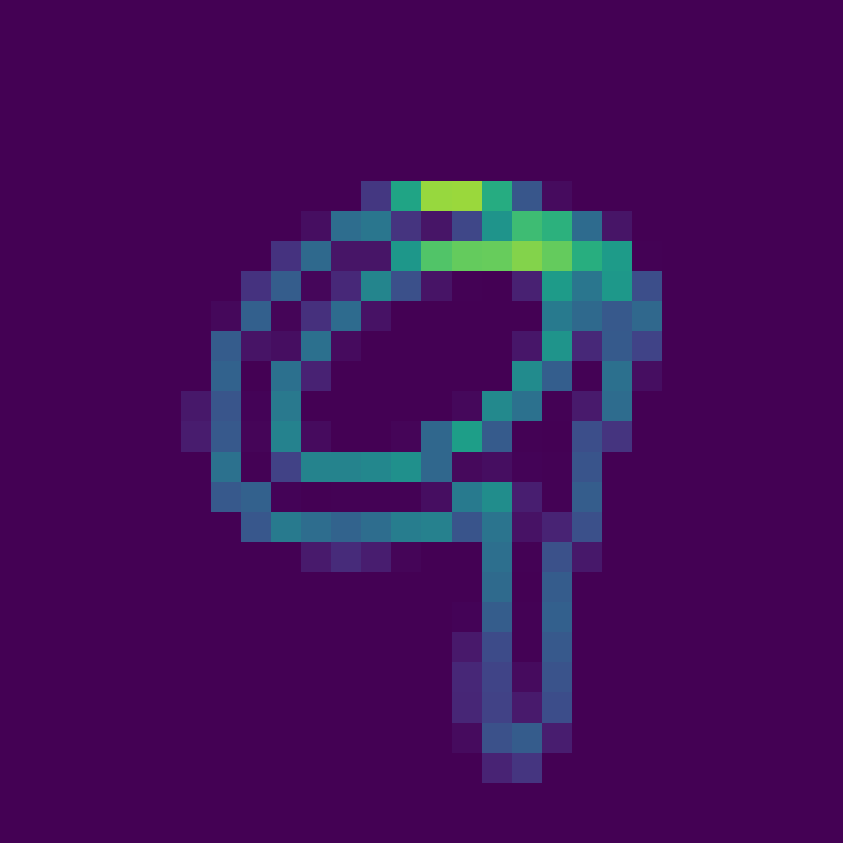} &
        \includegraphics[width=2cm]{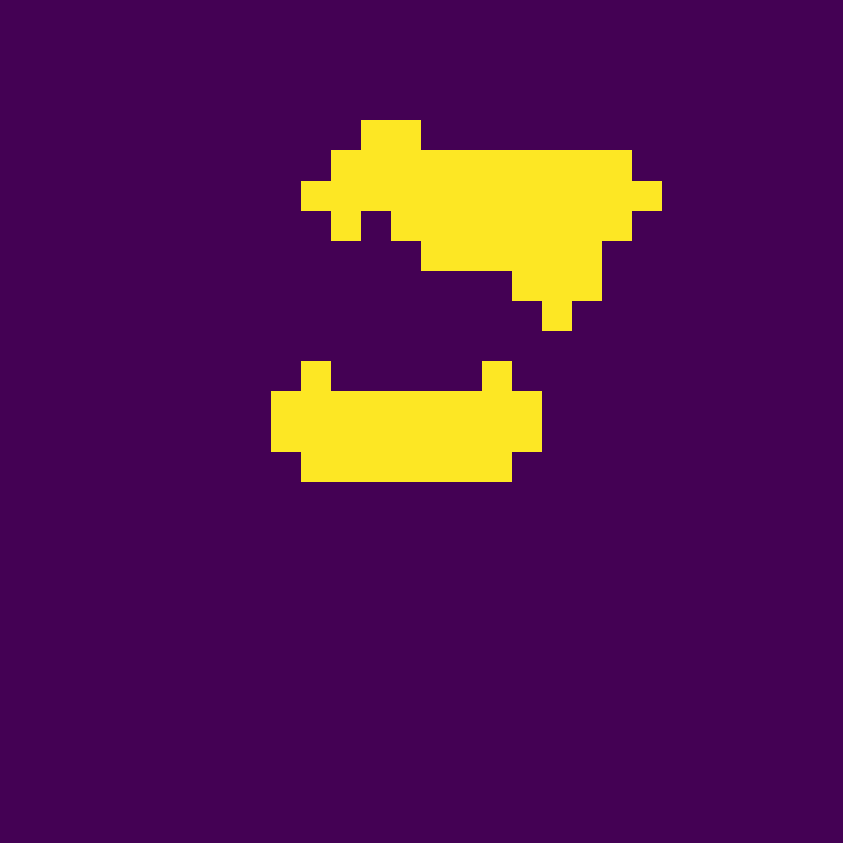} \\
        \hline
        \small{$|\V| = 43$} \newline \small{$\eta_{min}(|\V|) = 0.17$} \newline \small{$\eta_{max}(|\V|) = 0.37$} \newline \small{$\| x_{det} \| = 0.37$} &
        \includegraphics[width=2cm]{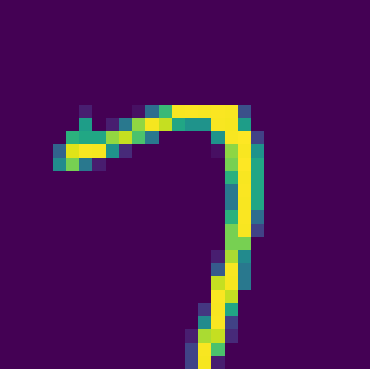} &
        \includegraphics[width=2cm]{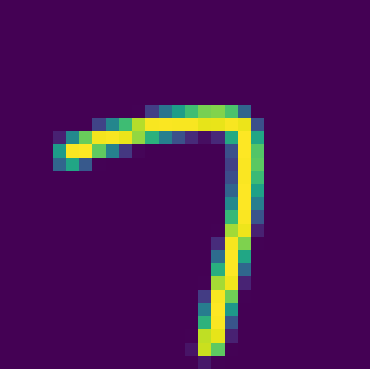} &
        \includegraphics[width=2cm]{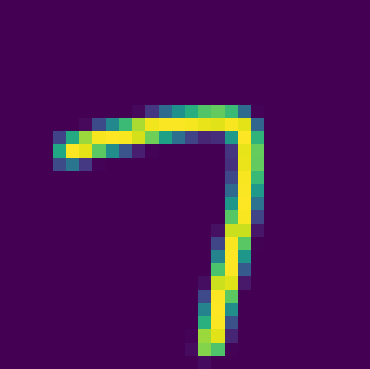} &
        \includegraphics[width=2cm]{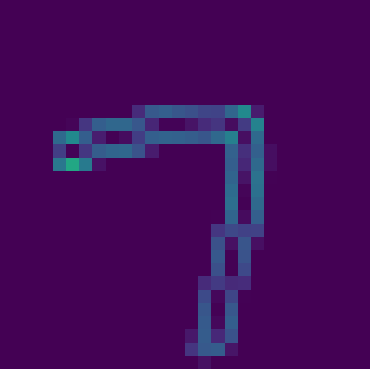} &
        \includegraphics[width=2cm]{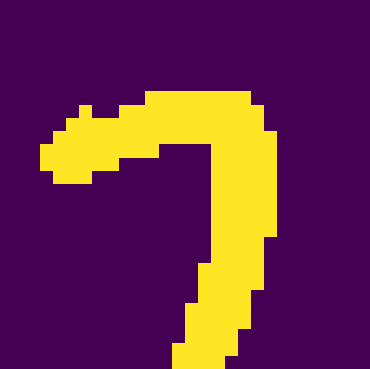} \\
        \hline
    \end{tabular}
\end{table}
\caption{Illustration of Definition \ref{def:detailtrans} and Theorem \ref{thm:iff_shift}.
The reconstructions are generated by a decoder trained with a noise level of $0.6$ and excluding digits $1,3,4$.
The variables $x$, $x_{det}$, $\eta_{min}$, $\eta_{max}$, and $\V$ are defined in $\S$ \ref{sec:exp_verif_setup}.
The first two rows present cases where the reconstruction is a hallucination, while the third row shows an accurate decoder reconstruction.
For the pixelwise average reconstruction and standard deviations, only those closer to $x + x_{det}$ than to $x$ under the semi-norm $\|.\|$ are included in $\V$; others are excluded, as they cannot contribute to the subset of noise for any $\eta$ in Definition \ref{def:detailtrans}.
The color scale for standard deviations matches that of the other images.
All reconstructions are consistent, and distances are normalized by the number of pixels in the region of interest.
In the last row, only reconstructions with an error less than $\|x_{det}\|/2$ are retained.}
\label{fig:def_iff_illustr_MNIST}
\end{figure}

In the first experiment, the reconstructions generated by the decoder trained without the digits 1, 3, and 4 are clear hallucinations.  Using these examples, we illustrate how hallucinations relate to the Definition \ref{def:detailtrans} and Theorem \ref{thm:iff_shift}.
The setup is described in $\S$ \ref{sec:exp_verif_setup}.
More specifically, we consider the low-resolution (LR) image of the data set as the input $y$, and the high-resolution (HR) image as $x \in \cM_1$ such that $fx = y$. The point $x+x_{det}$ is defined by the first reconstruction from the LR image $\phi(y) = x + x_{det}$ (with $e_0 = 0$). After verifying that the reconstruction $\phi(y)$ is consistent with the input $y$, we sample $n = 50$ noisy LR images $(f\phi(y) + e_m)_{1 \leq m \leq M}$, where $e_m$ is a noise vector for every $m \in \set{1, \dots M}$ : $e_m \in \mathcal{E} = B^o(0, 0.6)$.  We first confirm that the reconstruction $\phi(y)$ is consistent, namely $f\phi(y)-y\in\mathcal{E}$. Then, for each noisy LR image, we apply the decoder to obtain the reconstruction $\phi(f\phi(y) + e_m)$ and verify its consistency as well. We then use the consistent reconstructions to compute the \textit{minimum and maximum hallucination sizes} defined in $\S$ \ref{sec:exp_verif_setup}.
Here, the semi-norm $\|.\| = \|.\|_{1, 1, \{R\}}/N_R$ is defined with $R$ obtained by segmenting the region where the points $x$ and $x + x_{det}$ clearly differ and is displayed in Figure \ref{fig:def_iff_illustr_MNIST} for each example. $N_R$ denotes the number of pixels in $R$.
\noindent In Figure \ref{fig:def_iff_illustr_MNIST}, we observe that when the decoder $\phi$ hallucinates, the \textit{maximum hallucination size} $\eta_{max}$ is significantly larger than in cases without such hallucinations. Thus, the \textit{hallucination size} $\eta$ can quantify the severity of a hallucination when the metric $\|.\|$ is appropriately chosen. All reconstructions $(\phi(f\phi(y) + e_m))_{1 \leq m \leq M}$ obtained from noisy measurements of the synthetic image $x + x_{det}$ remain semantically accurate.

\noindent Note that the semi-norm focuses on a region of interest (ROI) $R$, selected based on the difference between synthetic image $x + x_{det}$ and the image $x$. However, region of interest $R$ cannot be directly identified using only the decoder-agnostic part of the method described in $\S$ \ref{sec:methods}.
To address this, the region of interest $R$ can be defined as a rectangular patch in the grid in the image. 
\subsection{Sub-Sampled MRI Scans Reconstruction}\label{sec:MRI}
The fastMRI challenge provides raw multi-coil knee and brain MRI data acquired on clinical scanners using standard pulse sequences. Participants receive undersampled k-space measurements that mimic accelerated acquisitions commonly used in clinical practice to reduce scan time. The objective is to reconstruct high-quality images from these incomplete measurements, achieving radiologically usable image quality. Ground-truth reference images are obtained from fully sampled acquisitions, enabling supervised training and quantitative evaluation. This experimental setup reflects realistic clinical constraints on scan duration and coil configurations. In the 2020 fastMRI challenge, hallucination artifacts were reported, even among the highest-ranked submissions \cite{fastmri20}.

\noindent To evaluate our method, we focus on brain images and consider a single-coil setting for simplicity. The original fastMRI data set \cite{fastmriData} contains only multi-coil acquisitions. For each slice, a number of $r$ coils are available, each consisting in a $2$ dimensional complex-valued image. To synthetically construct single-coil data, we first crop, align, and resize each coil image of this slice: $\forall c \in \set{1,\dots, r}, x_{c} \in \mathbb{C}^{320 \times 320}$. The single-coil image corresponding to the slice is then generated using the element-wise root-sum-of-squares operation:
\[
\begin{aligned}
    \forall (i,j) \in \set{1, \dots, 320}^2,\quad & x^{(i,j)} = \sqrt{\sum_{c=1}^{r} | x_{c}^{(i,j)} |^2}  ,
\end{aligned}
\]
where $| \cdot |$ denotes the modulus.
The corresponding measurement $y \in \mathbb{C}^{320 \times 320}$ is obtained by applying the forward model
\[
\begin{aligned}
    y = M\odot\mathcal{F}(x) ,
\end{aligned}
\]
where $\odot$ denotes element-wise multiplication (the Hadamard product), $\mathcal{F}$ is the two-dimensional fast Fourier transform, and $M$ is a binary sampling mask. The mask $M$ corresponds to an $8\times$ acceleration using uniform Cartesian sub-sampling with $22$ fully sampled central k-space lines. The forward model is linear and will therefore be denoted by $M \odot \mathcal{F}(x) = fx$.
The data set consists of more than $16,500$ inferior axial brain slices. The metric on the image space $\mathcal{X}$ is chosen as $\| \cdot \|_{1,{R}}$, where $R$ denotes a predefined region of interest, while the metric on the measurement space $\mathcal{Y}$ is the standard complex $\ell^2$ norm. For the decoder under evaluation, we use the U-Net baseline decoder trained on multi-coil brain data from \cite{fastmriData}. During training, the coil images for each slice were combined using the root-sum-of-squares operation before being fed into the network. As a result, the U-Net operates effectively on single-coil inputs and can be applied directly in the current setting. Since the decoder is single-valued, we denote a reconstruction of a decoder by $\phi(y) \in \mathcal{X}$ and refer to the reconstruction error using the norm $\| \cdot \|$ throughout this paragraph.\\
In accordance with the methods described in $\S$ \ref{sec:methods}, we first assess whether an automatic search in $F_y$ is feasible using the FastMRI data set. This assessment can be framed as the following question: given an input $y \in \cM_2$, does there exist a point $x$ in the data set such that $x \in F_y$ for a reasonable noise level $\epsilon$ ?

\noindent Since a relevant noise level $\epsilon$ for the MRI acceleration inverse problem is unknown, we propose an alternative to the method in $\S$ \ref{sec:methods}: rather than fixing $\epsilon$, we define $\epsilon_0$ as the minimal value $\epsilon_0$ such that, for a given sub-sampled k-space $y \in \cM_2$, $F_y^N$ contains two distinct data set images $x, x' \in \cM_1$.

\noindent To determine whether $f(x) - f(x')$ can be considered noise, we consider two approaches. First, we visually inspect the images corresponding to the measurements $f(x)$ and $f(x')$. Second, we examine the reconstructions from a pretrained decoder, $\phi(f(x))$ and $\phi(f(x'))$. Assuming that the decoder $\phi$ captures the noise model, we conclude that the differences between measurements $f(x) - f(x')$ is not a noise vector if the reconstruction $\phi(f(x))$ contains details of the image $x$ absent in the image $x'$, and vice versa.

\noindent We evaluated this condition on the closest pairs of measurements of images $f(x), f(x')$ in the single-coil FastMRI data set and found that it is not satisfied. Consequently, automated searching in the feasible set $F_y$ within the given data set cannot be used to anticipate or detect hallucinations.
\subsubsection{Hallucinations from a Qualitative Perspective}\label{par:QualHallMRI}
\noindent Under the assumption of a linear forward model, verifying the consistency of a reconstruction $\phi(y)$ according to Definition \ref{def:consistent} enables the detection of detail transfers that have a distinguishable component in the measurement space. Details without such a component are invisible in the measurements, and their partial transfer cannot be prevented. In this paragraph, we evaluate whether the latter case encompasses significant details in the context of MRI imaging acceleration.

\noindent To identify relevant details, we implemented two approaches. First, we selected details from anomalies in the FastMRI data set and segmented them based on clinical pathology annotations provided in \cite{zhao_fastmri_2022}. Second, we selected arbitrary details from the brain MRI scan data set.

\begin{figure*}[!htbp]
\centering
\setlength{\tabcolsep}{2pt}
\renewcommand{\arraystretch}{1}
\begin{tabular}{c*{5}{c}}
  & $x$ (base image) & $x+x_{det}$ & $x'$ (detail source) & $\phi(fx)$ & $\phi(f(x+x_{det}))$ \\
\textbf{A} &
\includegraphics[width=0.18\textwidth]{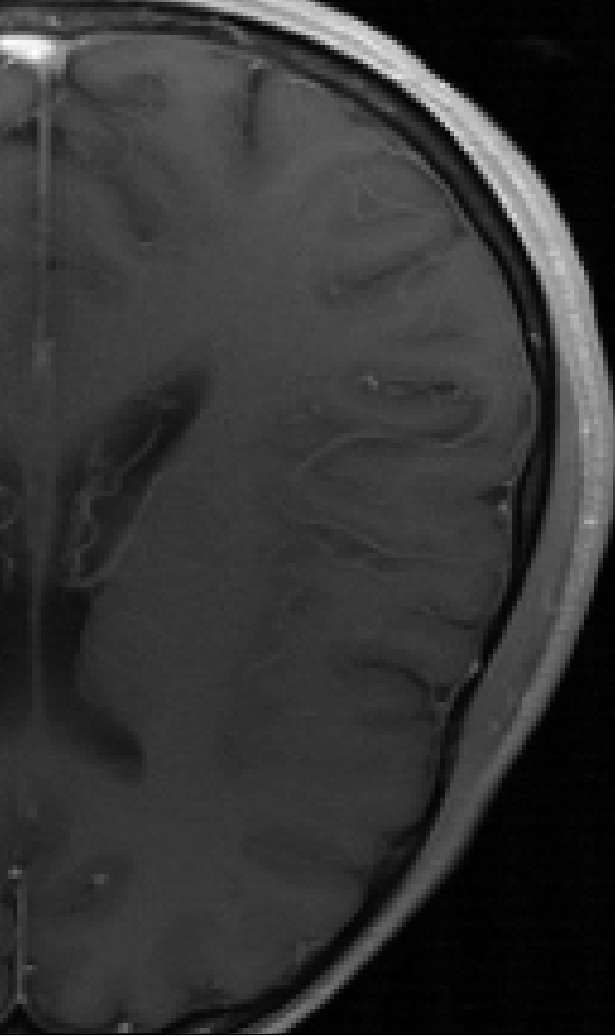} &
\includegraphics[width=0.18\textwidth]{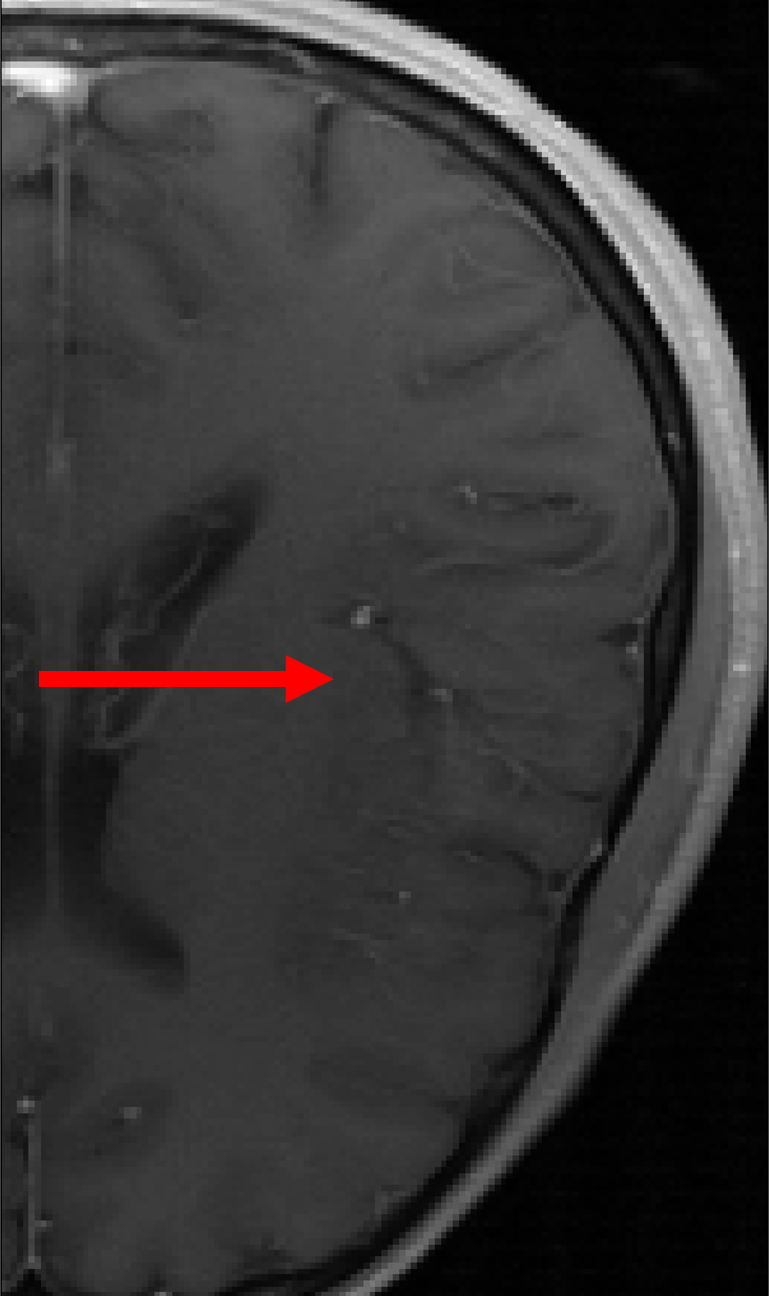} &
\includegraphics[width=0.18\textwidth]{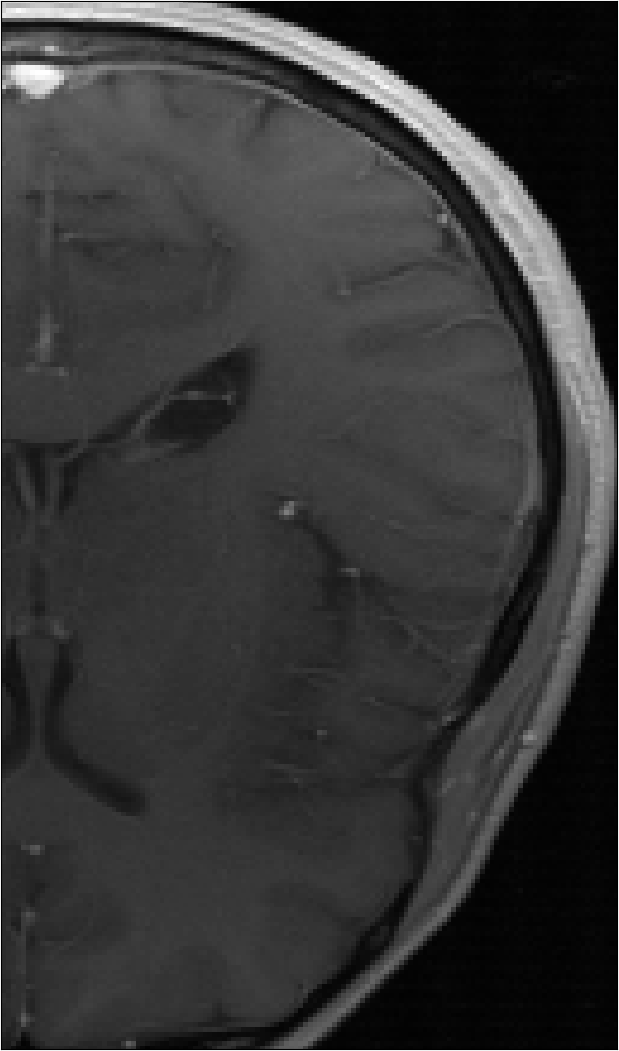} &
\includegraphics[width=0.18\textwidth]{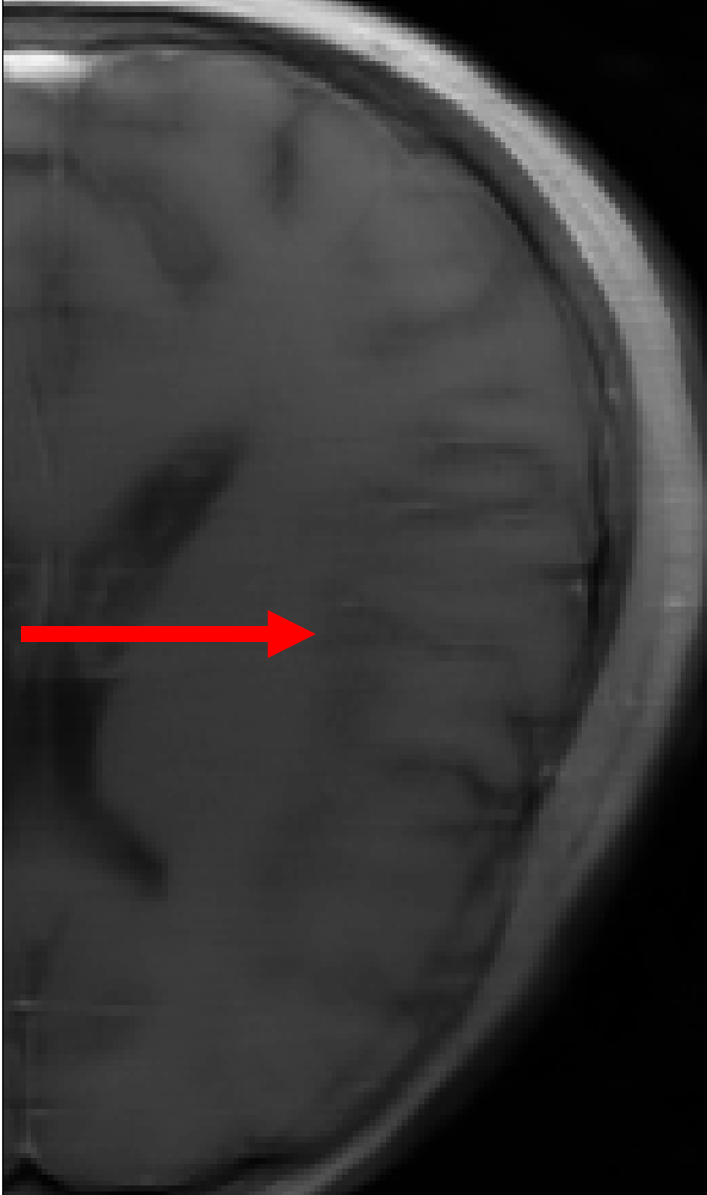} &
\includegraphics[width=0.18\textwidth]{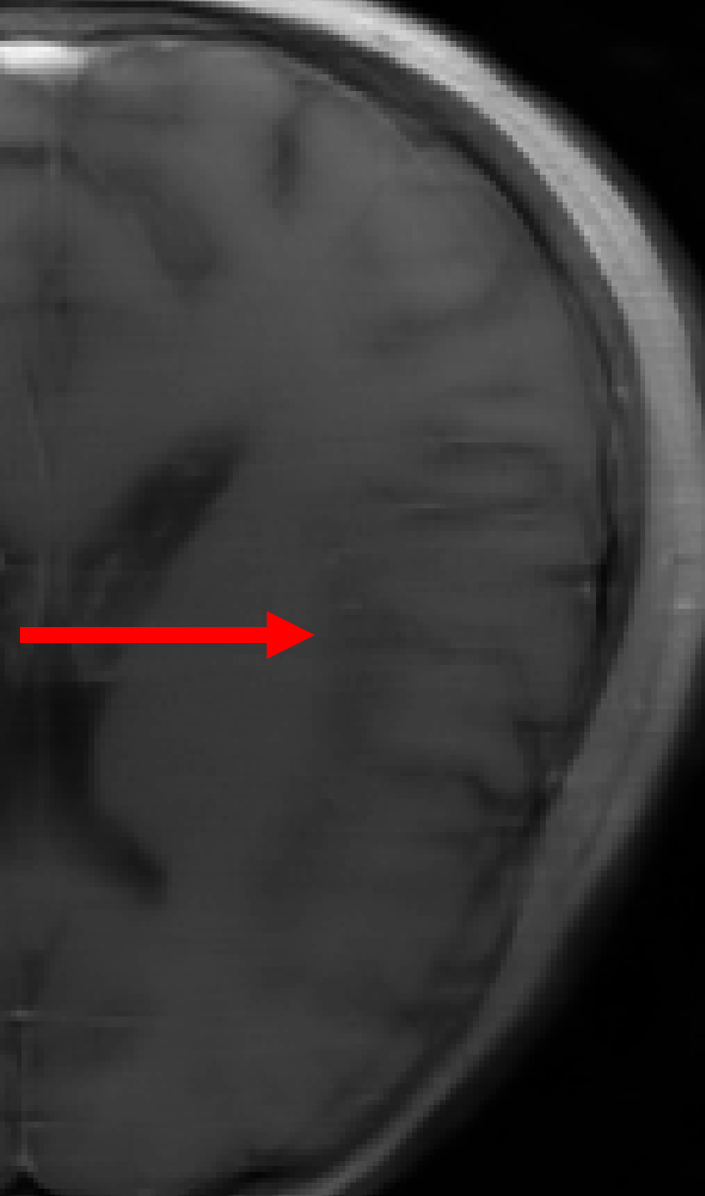} \\
\textbf{B} &
\includegraphics[width=0.18\textwidth]{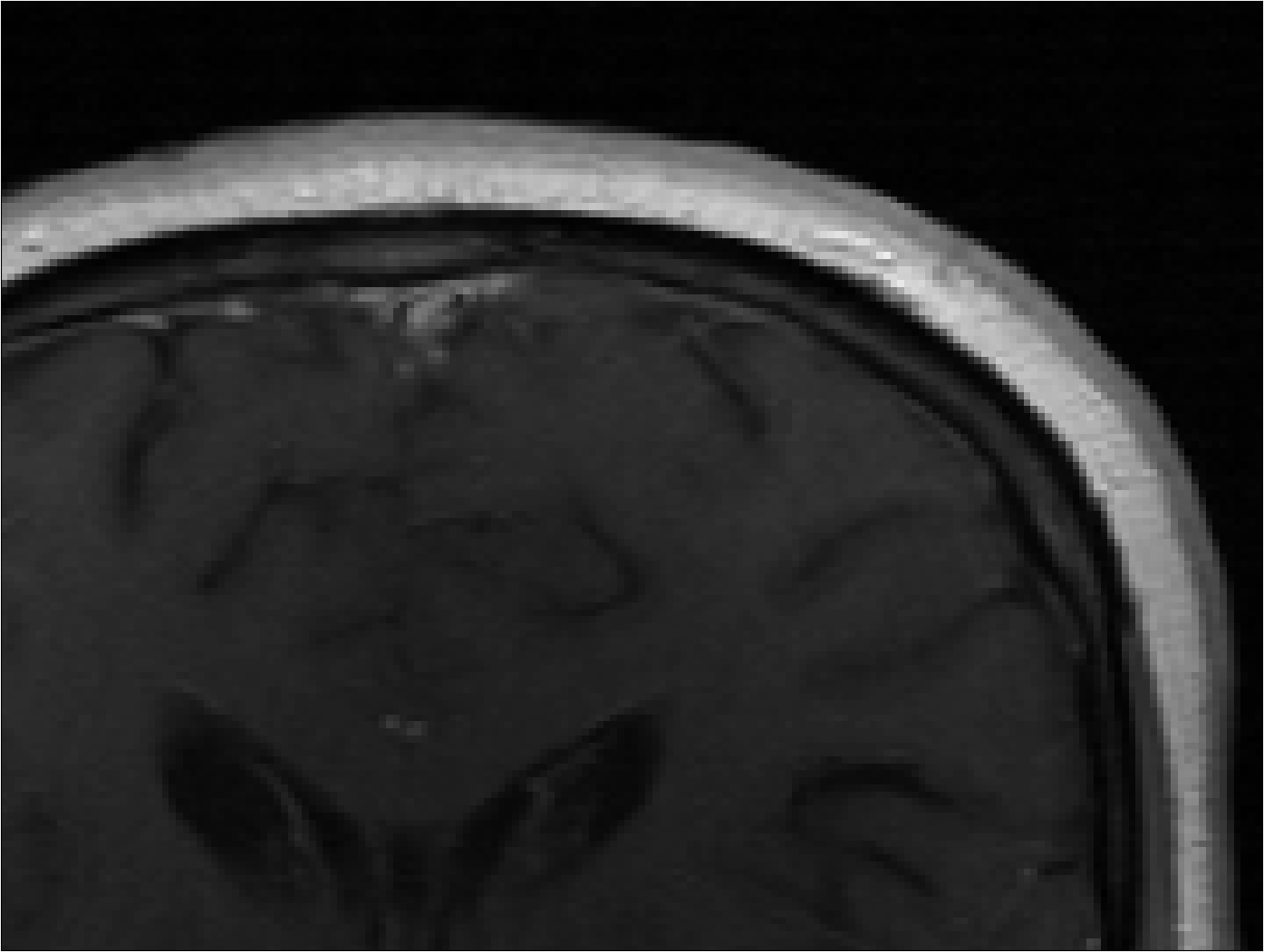} &
\includegraphics[width=0.18\textwidth]{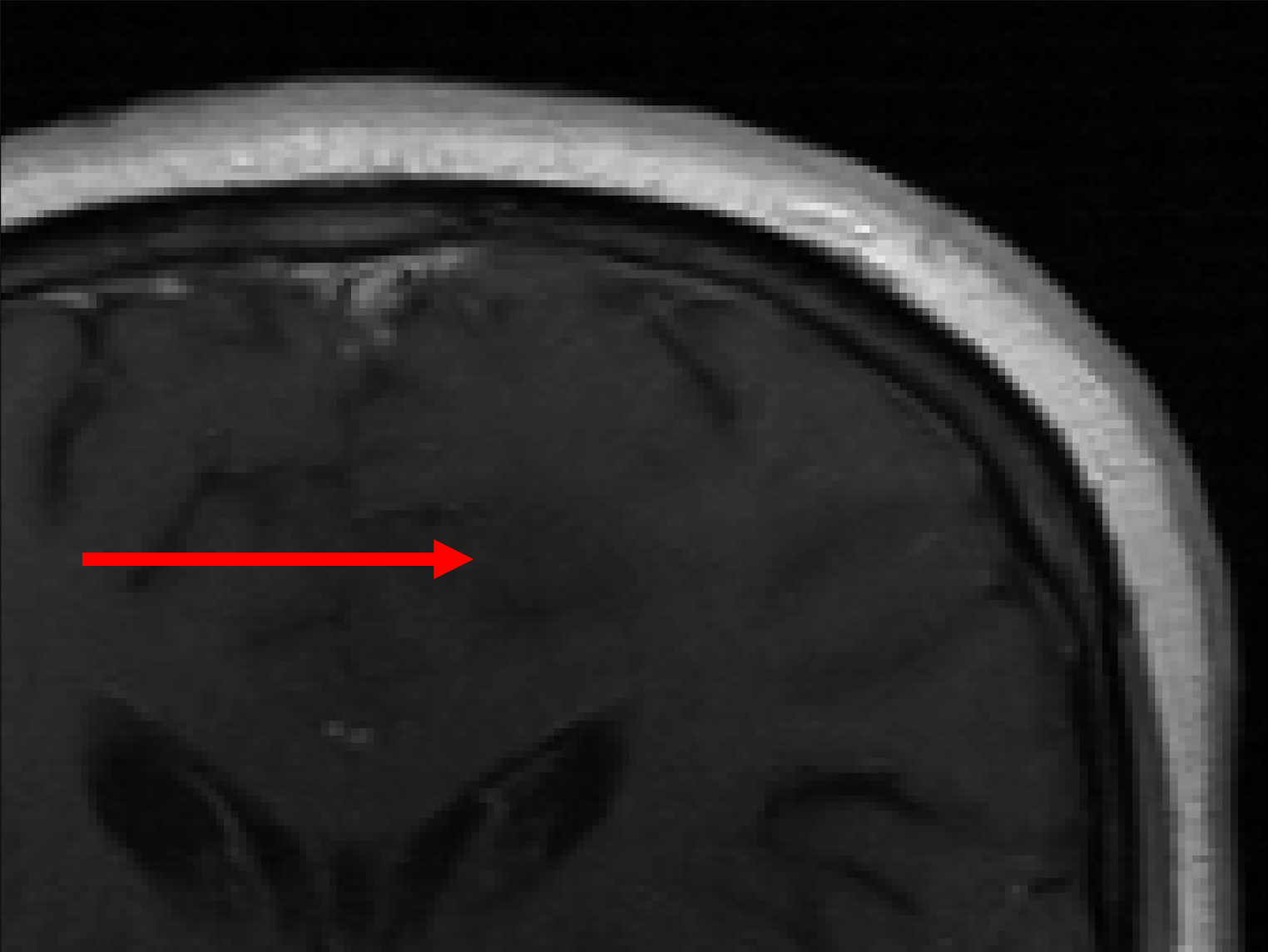} &
\includegraphics[width=0.18\textwidth]{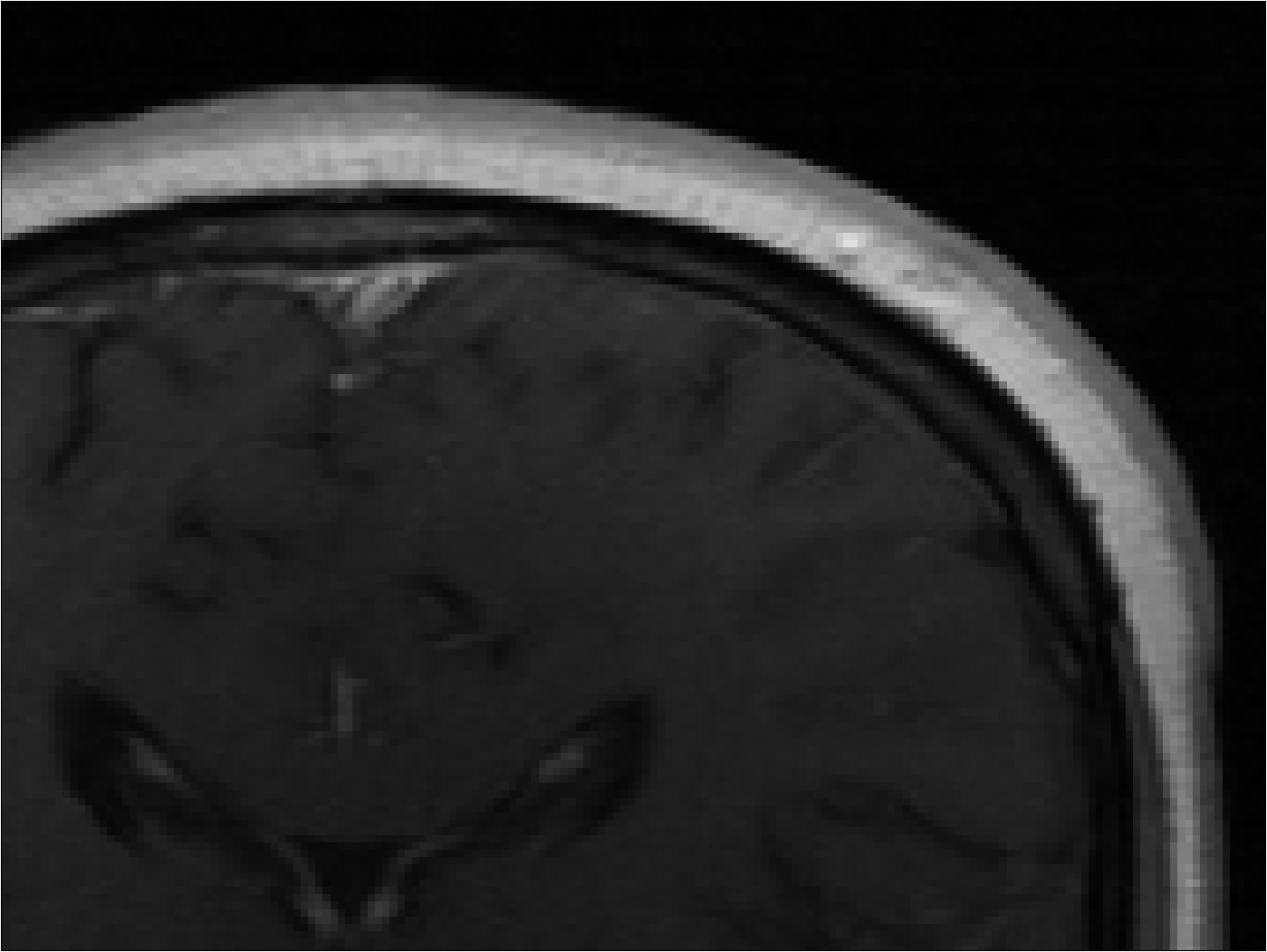} &
\includegraphics[width=0.18\textwidth]{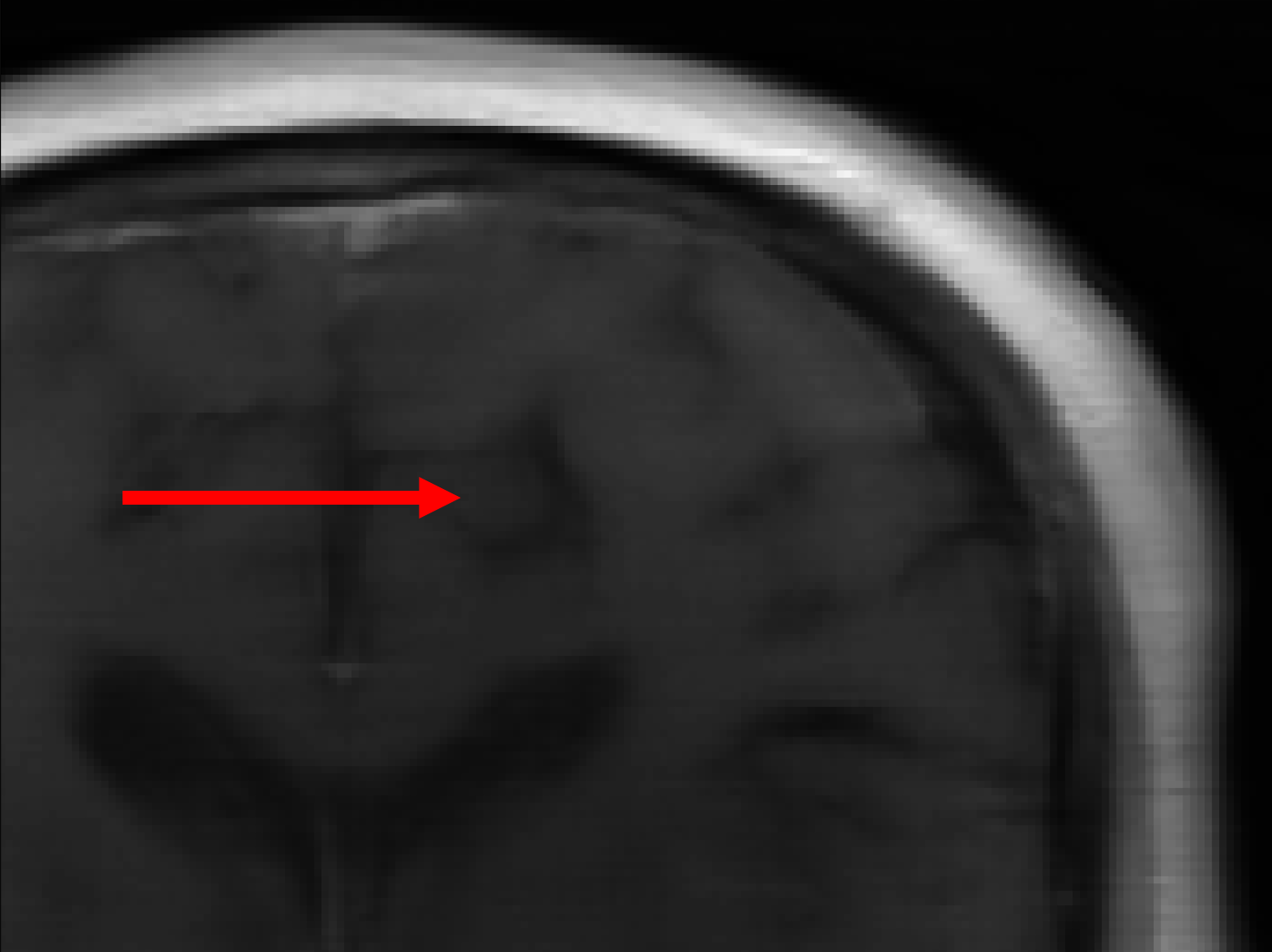} &
\includegraphics[width=0.18\textwidth]{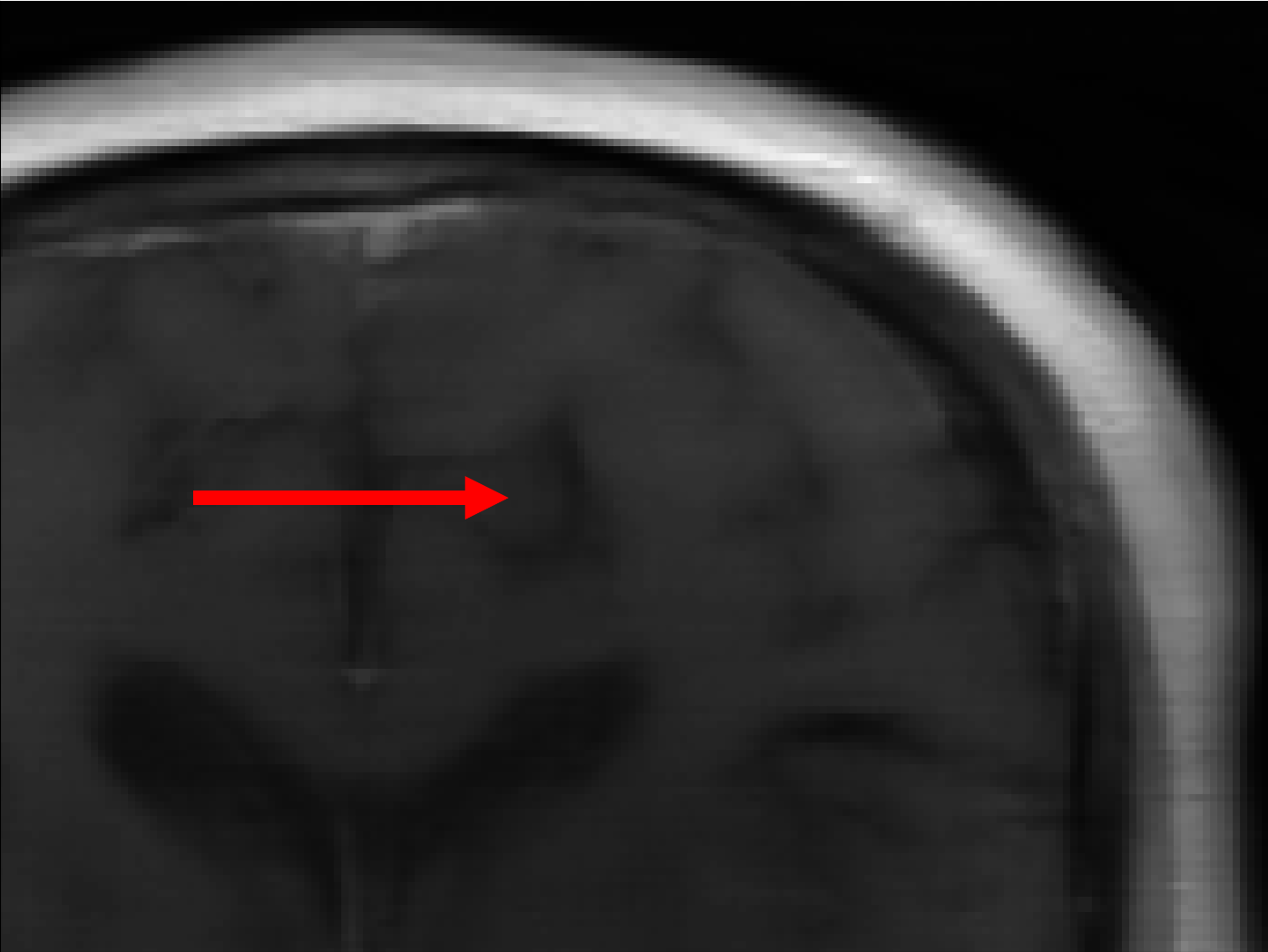} \\
\textbf{C} &
\includegraphics[width=0.18\textwidth]{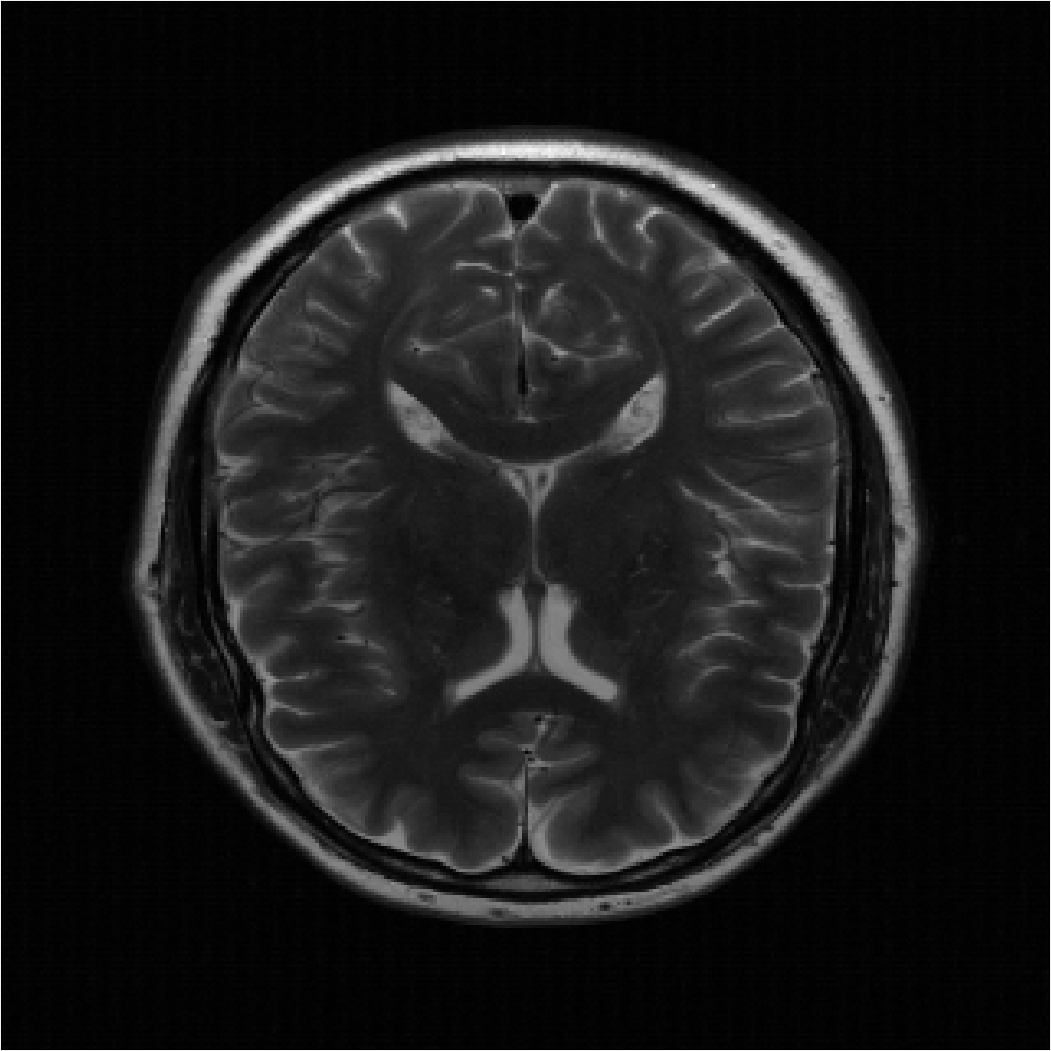} &
\includegraphics[width=0.18\textwidth]{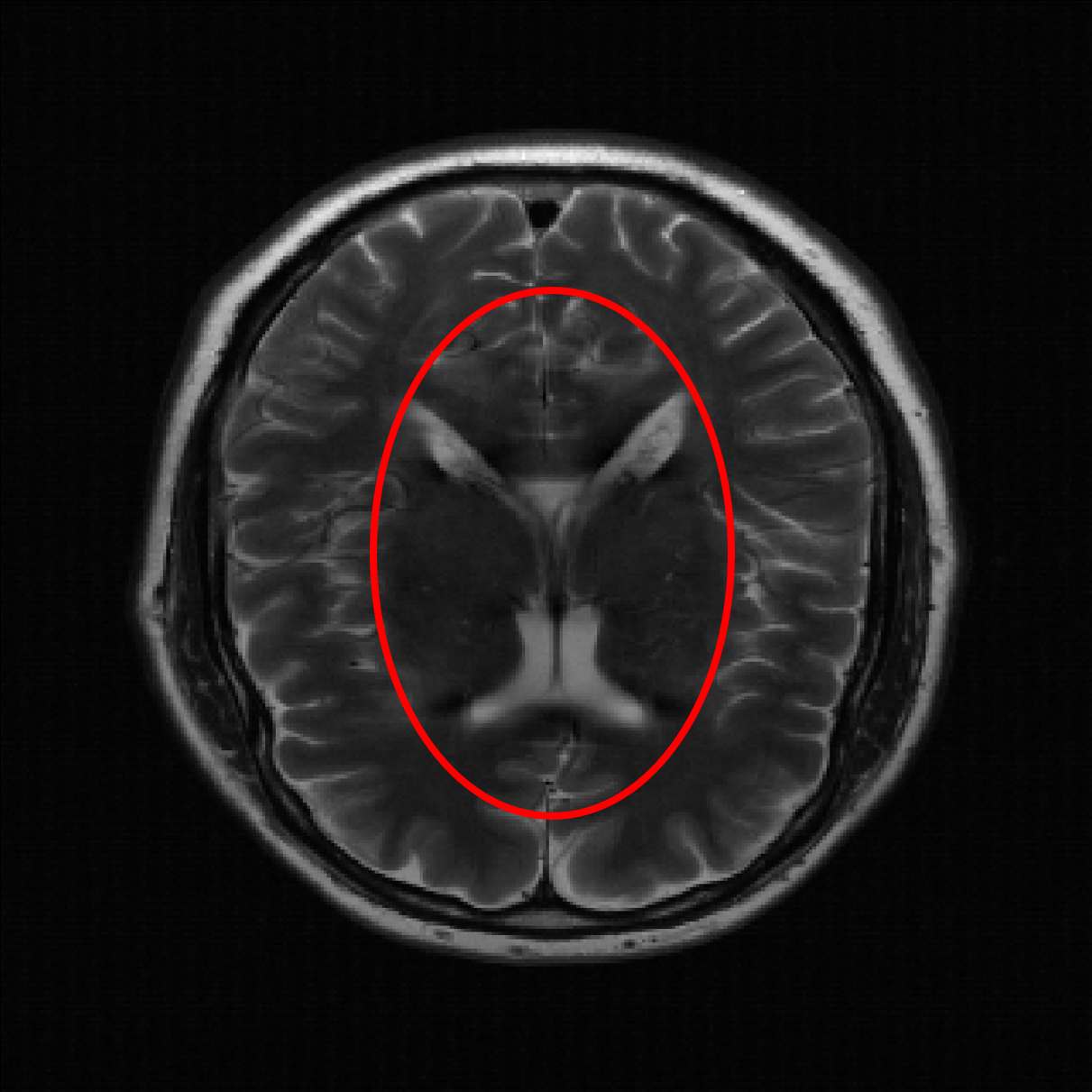} &
\includegraphics[width=0.18\textwidth]{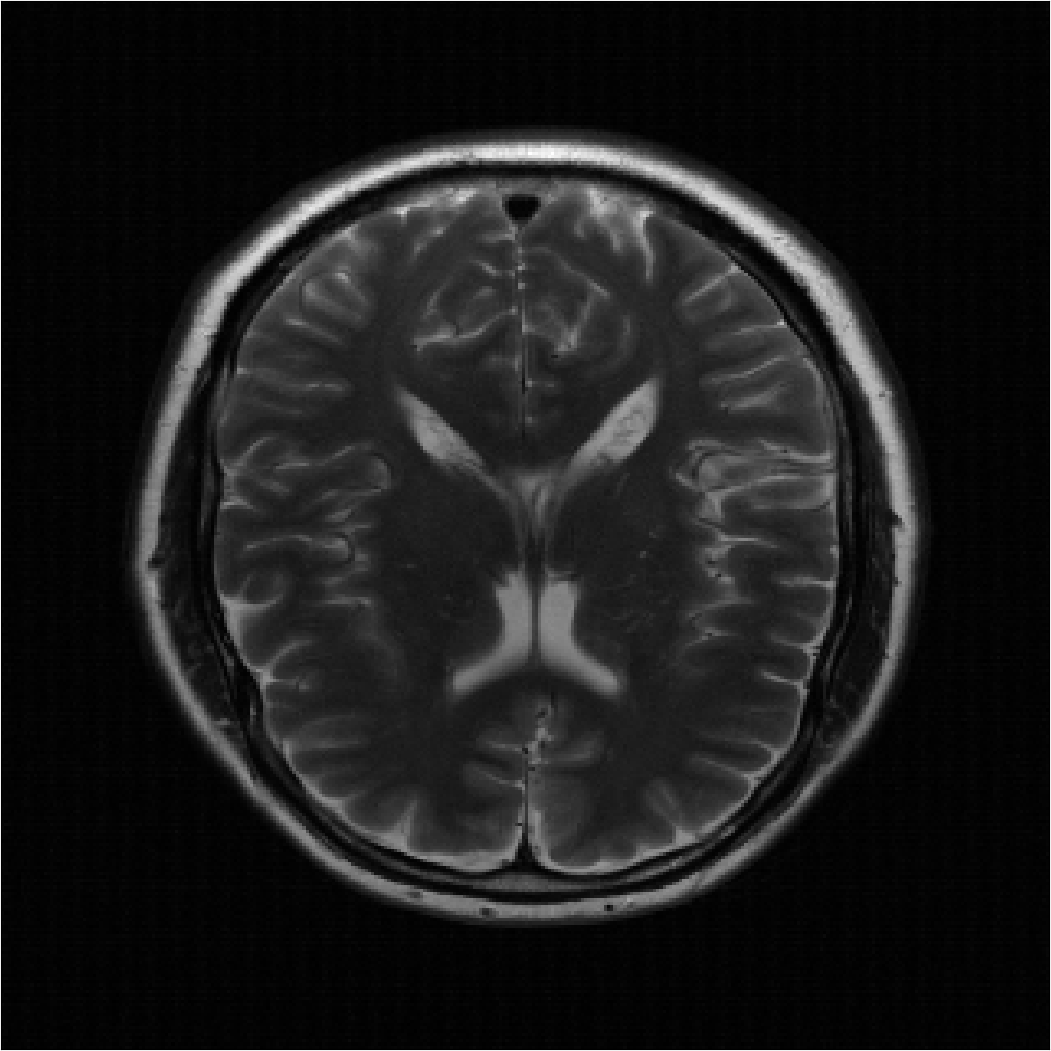} &
\includegraphics[width=0.18\textwidth]{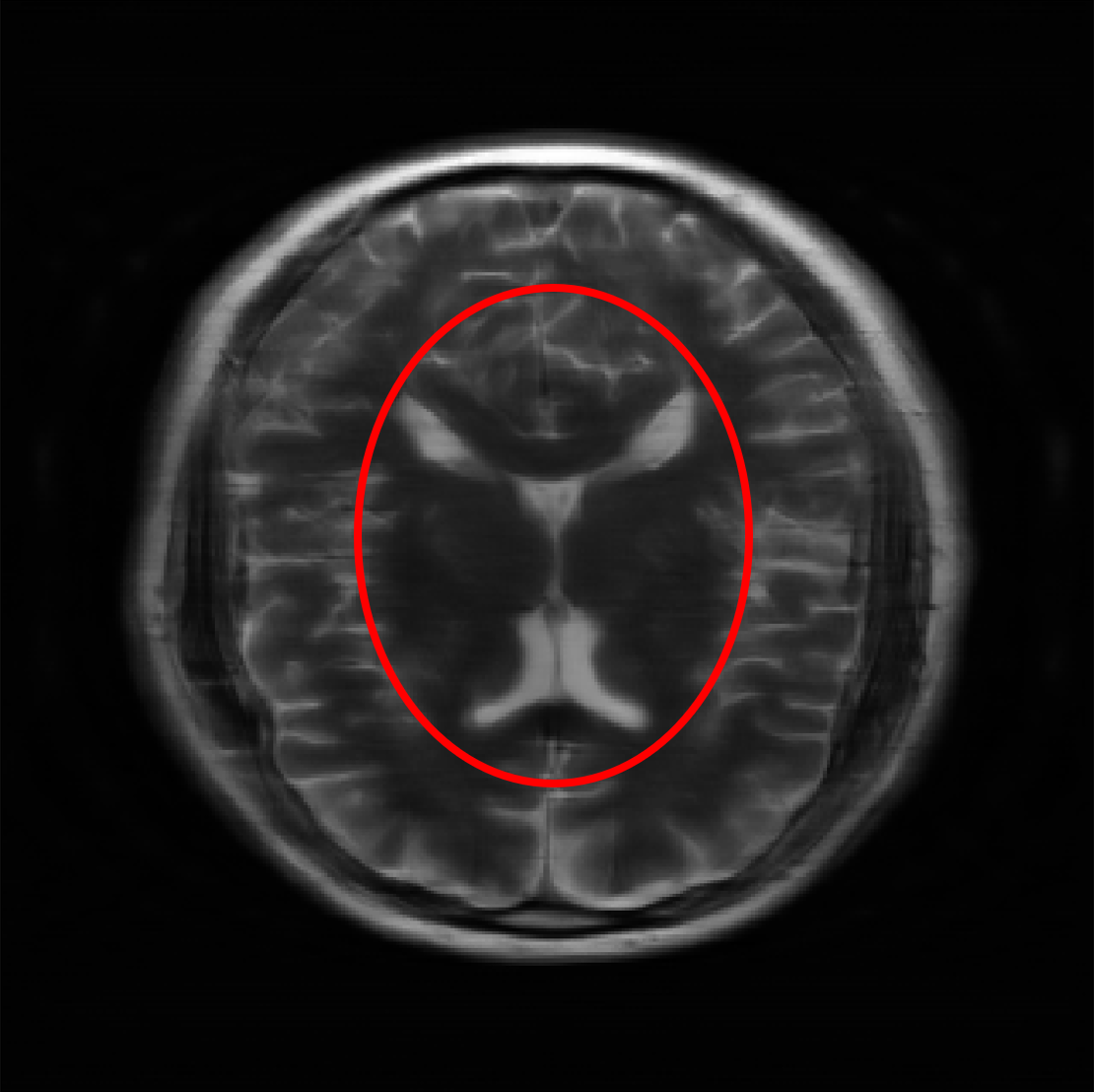} &
\includegraphics[width=0.18\textwidth]{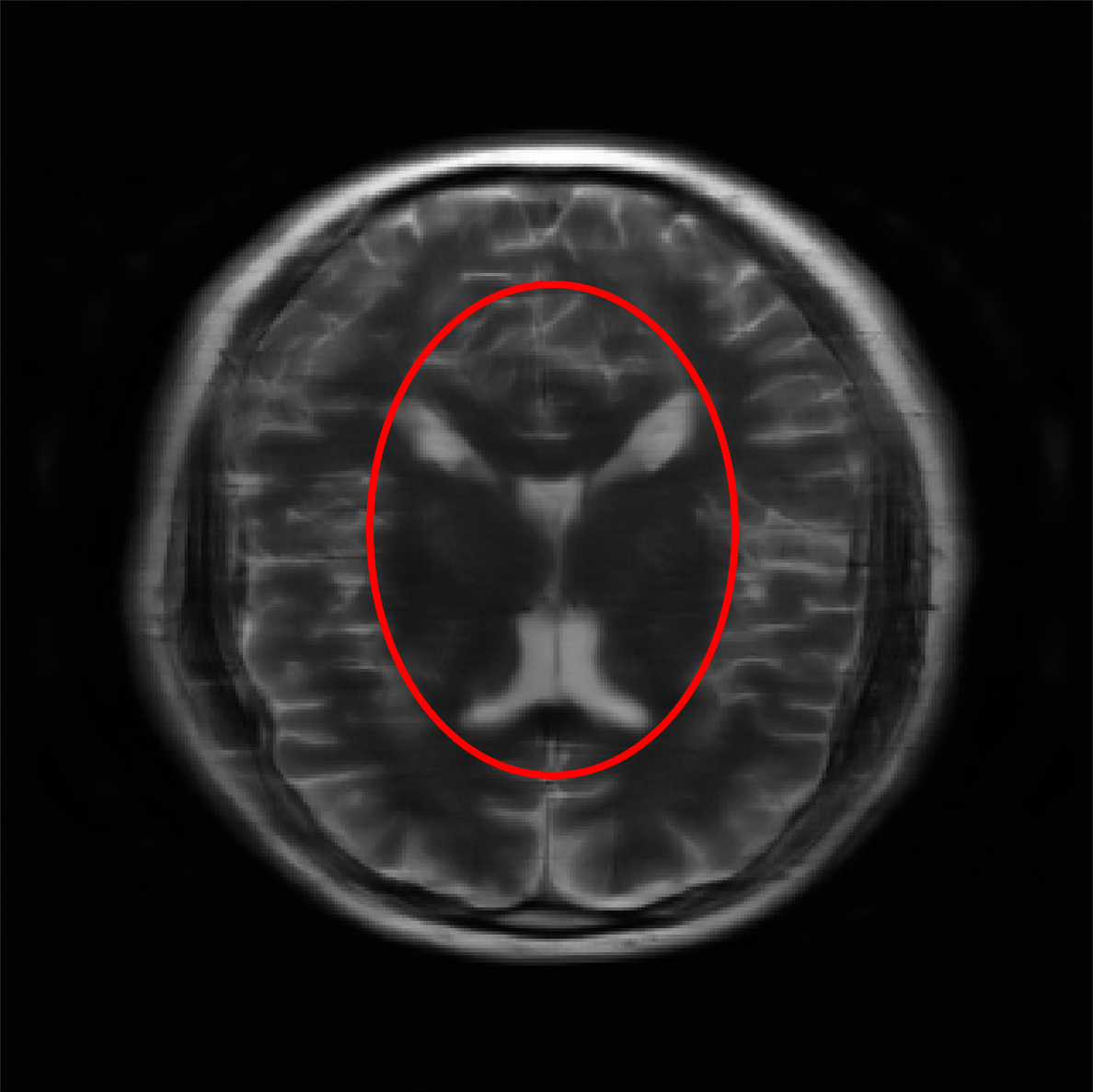} \\
\textbf{D} &
\includegraphics[width=0.18\textwidth]{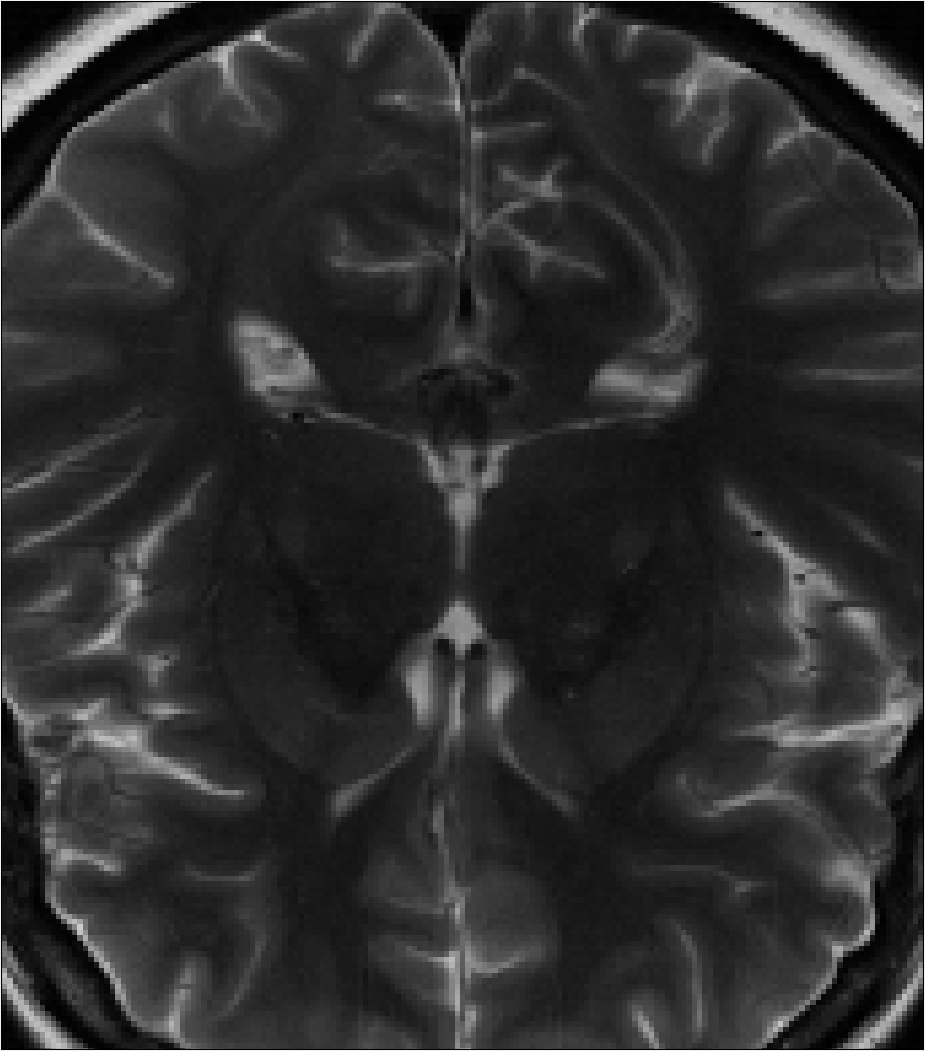} &
\includegraphics[width=0.18\textwidth]{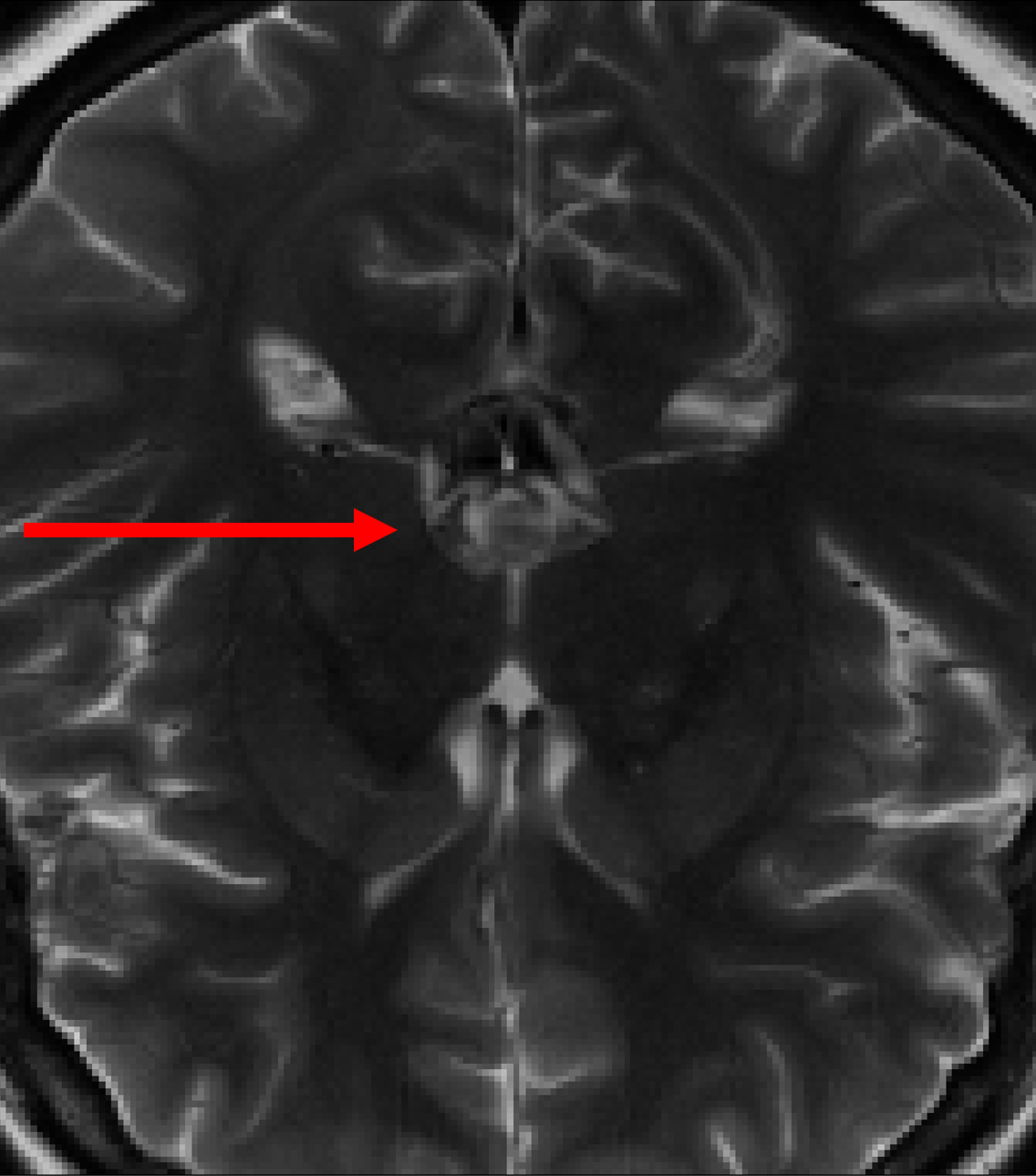} &
\includegraphics[width=0.18\textwidth]{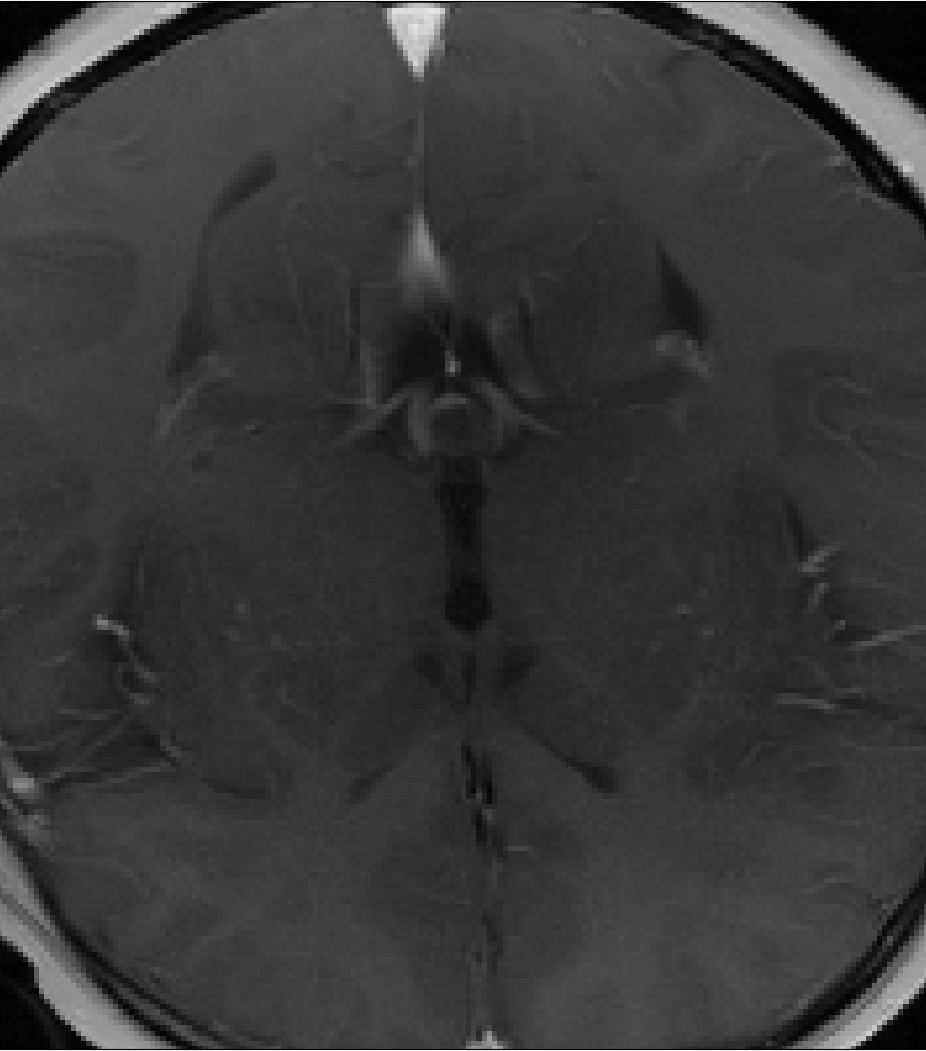} &
\includegraphics[width=0.18\textwidth]{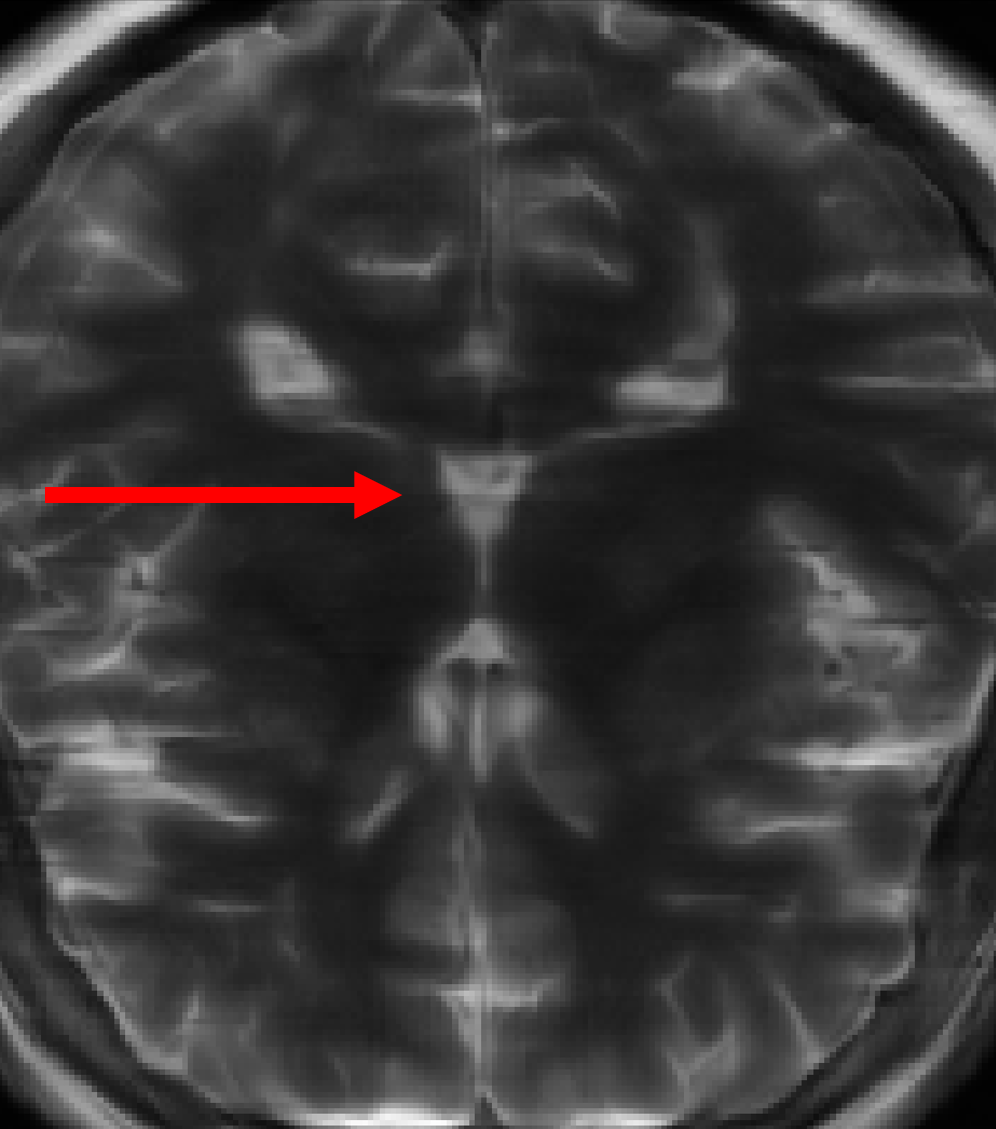} &
\includegraphics[width=0.18\textwidth]{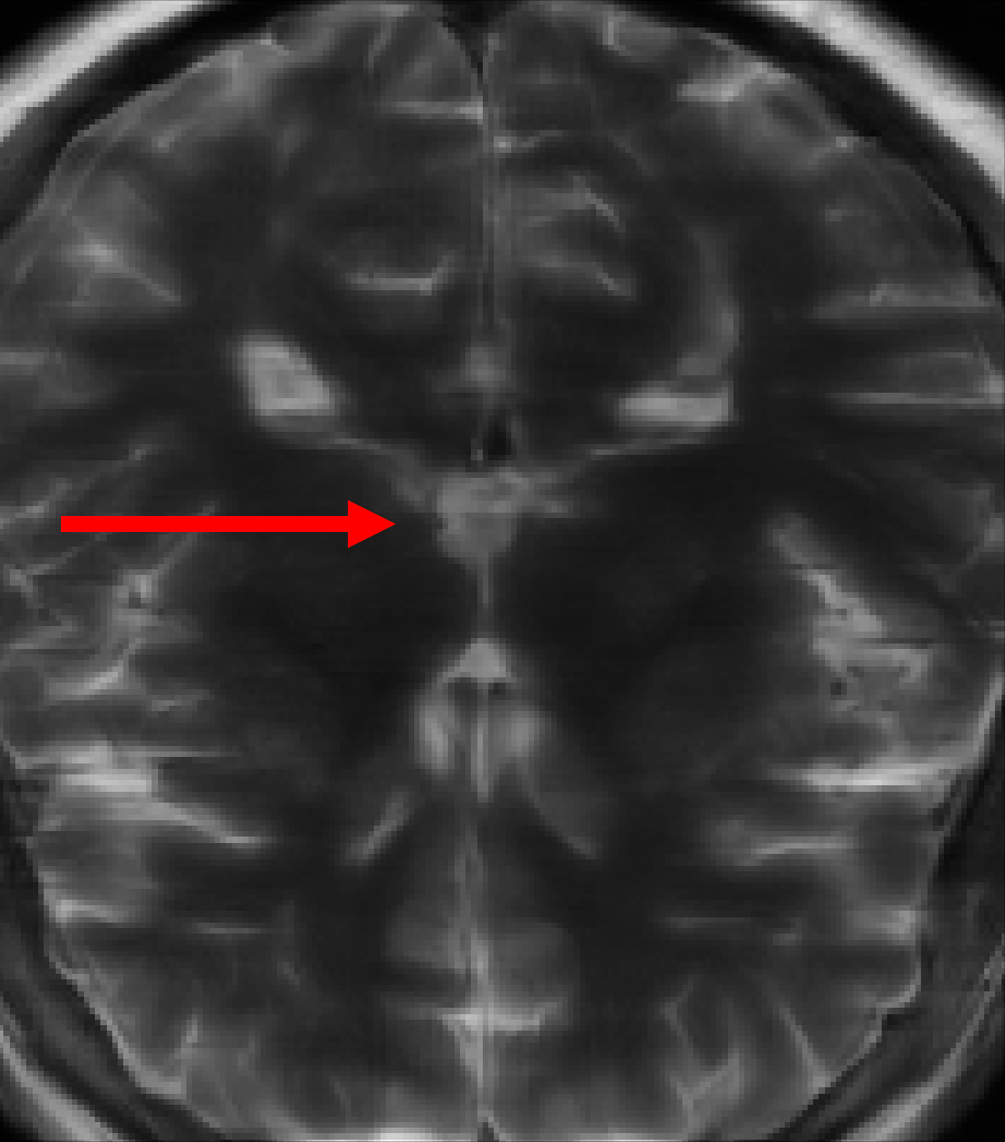} \\
\textbf{E} &
\includegraphics[width=0.18\textwidth]{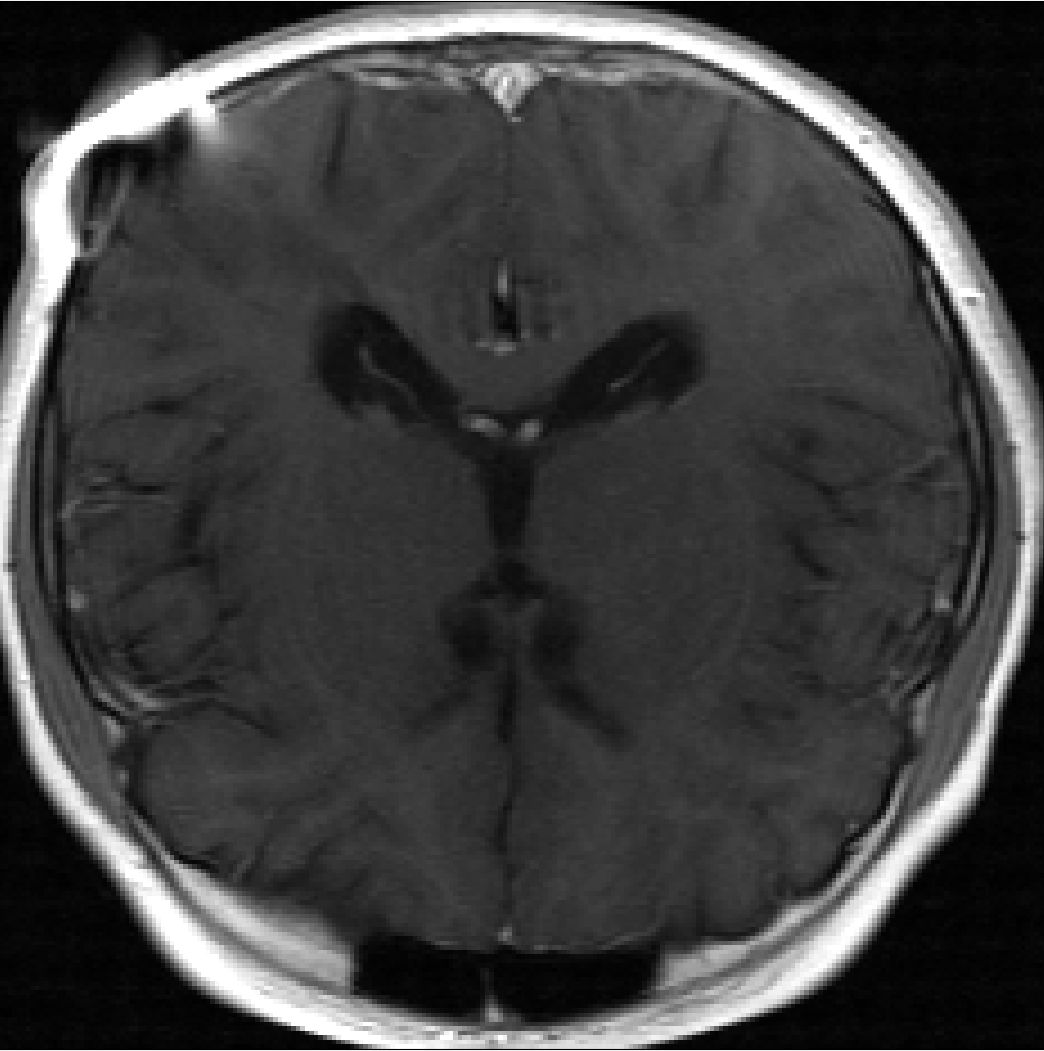} &
\includegraphics[width=0.18\textwidth]{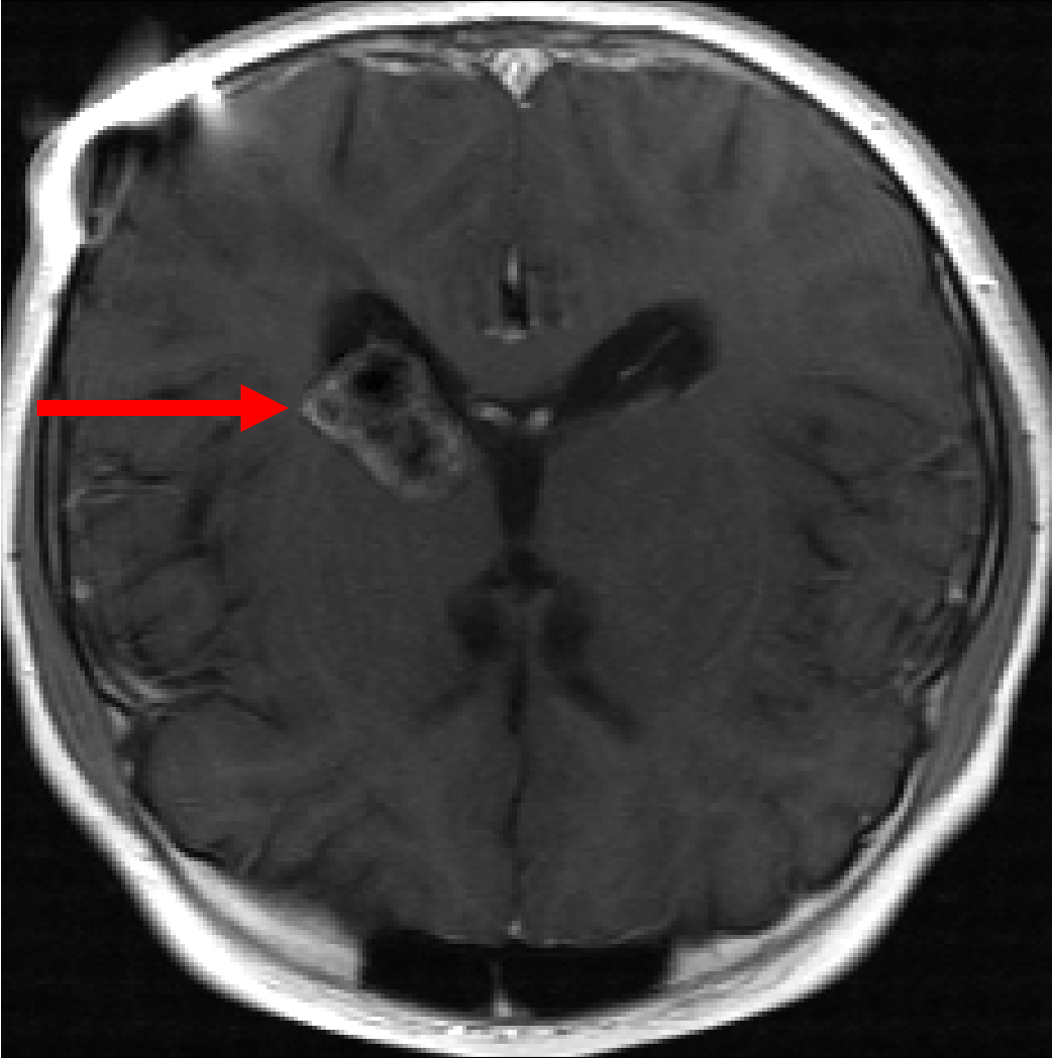} &
\includegraphics[width=0.18\textwidth]{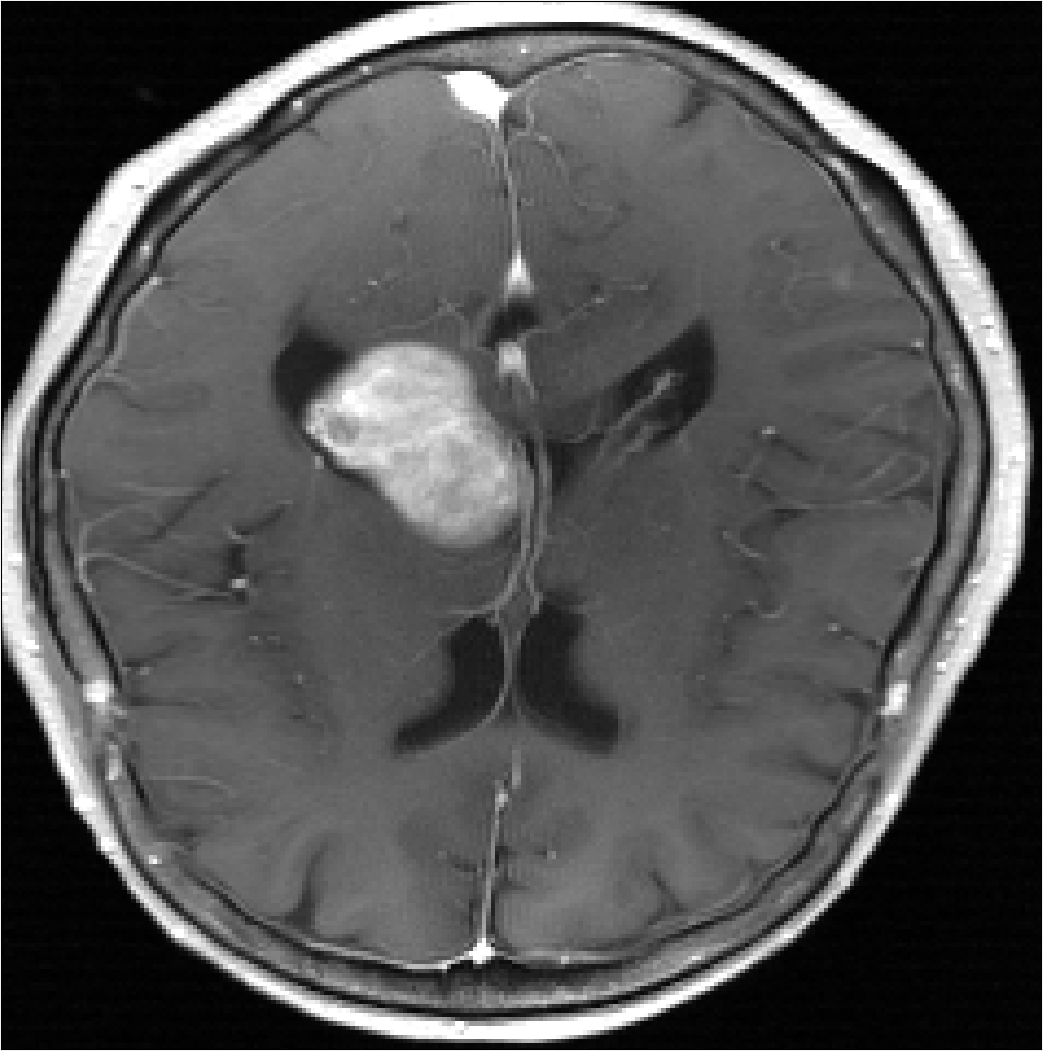} &
\includegraphics[width=0.18\textwidth]{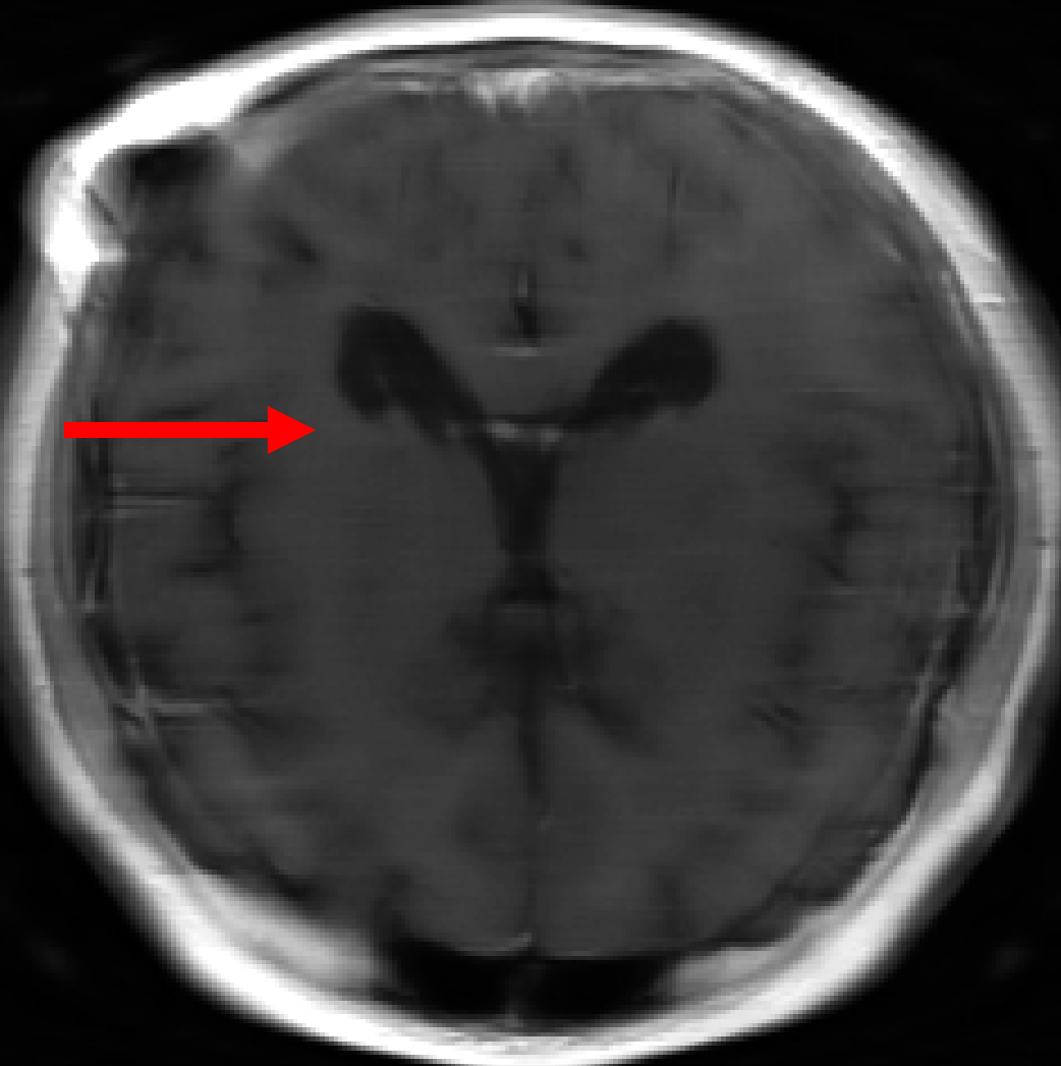} &
\includegraphics[width=0.18\textwidth]{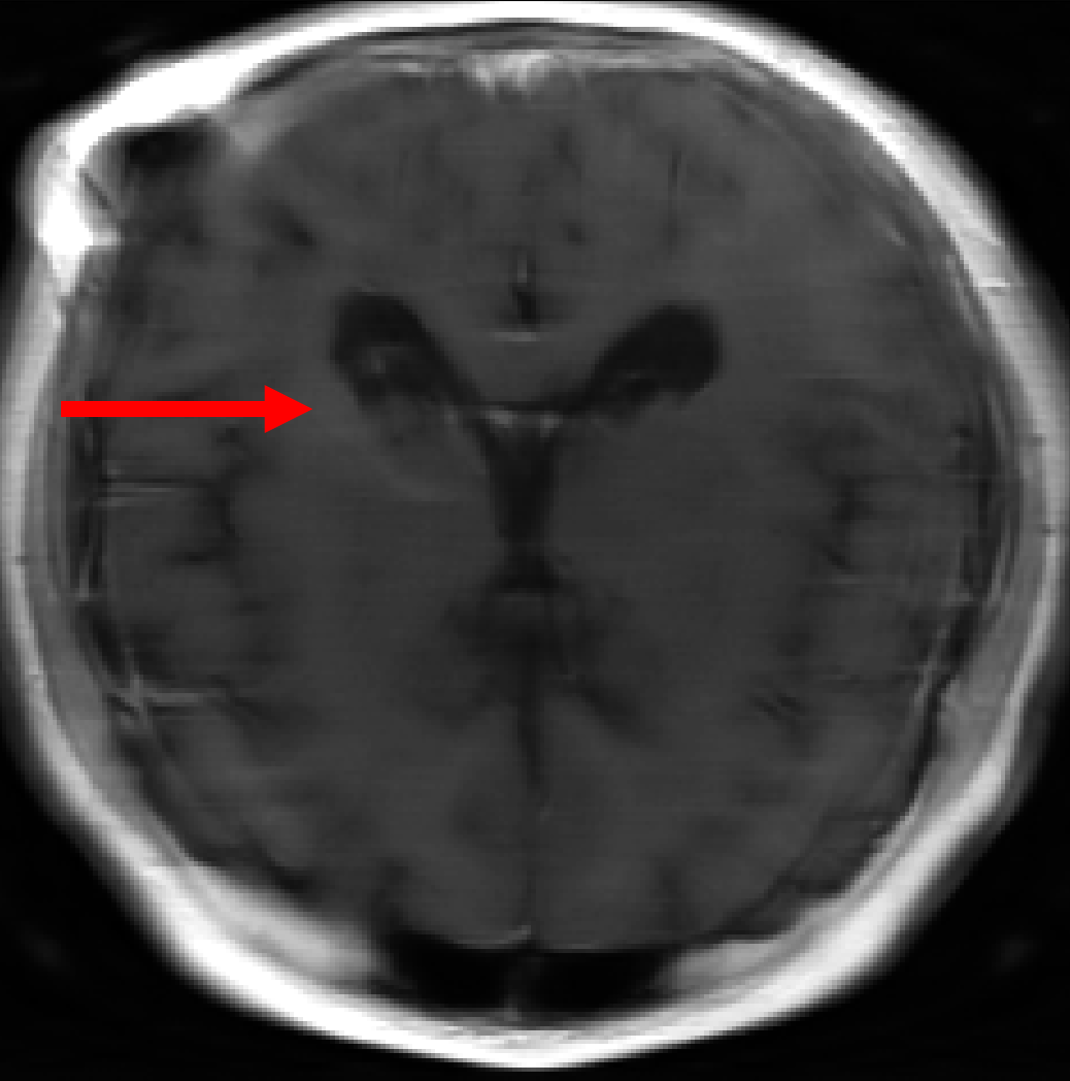} \\
\end{tabular}
\caption{Qualitative hallucinations examples in MRI acceleration. The first three column show the the base image ($x$), the synthetic image  $x+x_{det}$ and the one the detail is taken from the image $x'$. The two last show the reconstruction of the U-net decoder on the measurements of the base image $\phi(fx)$ and on the measurements of the synthetic images  $\phi(f(x+x_{det}))$. In the row A, we added blood vessels, and in row B we removed some. Row C is an example where the central ventricle has been replaced for the quantitative analysis. In row D we added an anomaly labeled as 'Pineal cyst' by \cite{zhao_fastmri_2022}, and in row E we added an anomaly labeled as 'Mass'. A red arrow points at the added or removed detail.}
\label{fig:qual_hall_MRI}
\end{figure*}

\noindent Using the almost invisible detail pasting method described in $\S$ \ref{sec:methods}, we constructed realistic images as shown in Figure \ref{fig:qual_hall_MRI}. For each image, we used two data set images, a base image $x$ and a source image of the detail to paste $x'$, to create a realistic image with a pasted detail, $x + x_{det} \in \cM_1$, consistent with the base input ($fx \approx f(x + x_{det})$). This construction implies that the synthetic image $x + x_{det}$ could plausibly be the ground truth for the given input $y = fx$. As shown in Figure \ref{fig:qual_hall_MRI}, the decoder $\phi$ treats the measurement $fx_{det}$ as a noise vector, since the reconstruction satisfies $\phi(f(x + x_{det})) \approx \phi(fx)$. Furthermore, the patterns observed in the reconstruction $\phi(f(x + x_{det}))$ generally resemble those of the image $x$. Consequently, if the synthetic image $x + x_{det}$ were the ground truth, the detail $x_{det}$ would be lost in the reconstruction. Notably, as illustrated in Figure \ref{fig:qual_hall_MRI}, a detail can represent the addition, removal, or replacement of another detail.
\noindent In the examples presented in rows A, B, C, and D, the synthetic image $x + x_{det}$ appears realistic. The pasted detail closely resembles its original version. Importantly, $x_{det}$ is not visible in any of the U-net reconstructions. Consequently, if the synthetic image $x+x_{det}$ were the ground truth, the decoder $\phi$ would hallucinate the absence of the detail from the image.

\noindent In row E, the mass originally present in the image $x'$ exhibits a non-negligible component in the measurement space. Its null space projection $x_{det}$ differs significantly from its version in $x'$. As a result, it is not possible for a decoder $\phi$ to hallucinate this mass from the input $y = fx \approx f(x + x_{det})$ while maintaining consistency. However, it remains ambiguous whether the hallucinated image $x + x_{det}$ is realistic. If it is realistic, the decoder's omission of this potentially significant anomaly from $y = f(x + x_{det})$ would be notable.
\subsubsection{Verification of the Definition of Hallucinations}\label{par:verif_def_MRI}
\noindent In a first example, we assume that the reference image $x$ is the scan with the anomaly in the second column of Figure \ref{fig:qual_hall_MRI}. The hallucinated image $x + x_{det}$ corresponds to the first reconstruction from the input in row C in Figure \ref{fig:qual_hall_MRI}. In this case, the decoder misses the anomaly.

\noindent In a second example, we assume that the reference image $x$ is the original scan, and that the synthetic scan with the cyst represents the hallucinated image $x + x_{det}$. For Definition \ref{def:detailtrans} to hold in this case, we assume the existence of a noise vector $e_0$ such that the decoder satisfies the following $\phi(fx + e_0) = x + x_{det}$. This second example does not correspond to a hallucination. So if Definition \ref{def:detailtrans} is coherent with the informal description of hallucination, the definition should not be satisfied with relevant values for the hallucination size. 
\begin{table}[h]
\centering
\caption{Quantities of Definition \ref{def:detailtrans} and Theorem \ref{thm:iff_shift} for the example of row C in Figure \ref{fig:qual_hall_MRI}. The noise level corresponds with a SNR value between $40$ and $50$. Only reconstructions that are closer to the image with the detail $x+x_{det}$ than to the image $x$ for the norm $\|.\|$ are kept in the set of noise samples causing hallucinations $\V$. Note that the other images cannot account in the subset of noise for any \textit{hallucination size} $\eta$ in Definition \ref{def:detailtrans}. All the reconstructions are consistent. The distance calculated by restricting to the Region of Interest associated to the anomaly and is normalized by the number of pixels of this region.}
\label{tab:def_hall_quant_MRI}
\begin{tabular}{|c|c|c|c|}
\hline
$\eta_{min}$ & $\eta_{max}$ & $\|x_{det}\|$ & $|\V|$  \\
\hline
$1.33 \times 10^{-5}$ & $1.25 \times 10^{-3}$ & $1.25 \times 10^{-3}$ & $11$ \\
\hline
$9.04 \times 10^{-4}$ & $9.02\times 10^{-4}$ & $9.02\times 10^{-4}$ & $11$ \\
\hline
\end{tabular}
\end{table}
\noindent We use the U-net decoder to predict from $10$ different noisy subsampled images $\set{fx +e_m, \, 1 \leq m \leq 10}$. $\V, \eta_{min},\eta_{max} $ are defined in $\S$ \ref{sec:exp_verif_setup} and reported in Table \ref{tab:def_hall_quant_MRI}. In the first situation, the decoder transfers the detail $x_{det}$ on $x$ for every $\eta \in [\eta_{min},\eta_{max}] $ Theorem \ref{thm:iff_shift} tells that the error for the considered subset of noise is lower than the \textit{minimum hallucination size} $\eta_{min}$ as $2\eta_{min} < \|x_{det} \|$. The reported values tell that it is even negligible with respect to $\eta_{max}$. In the second situation, the \textit{maximum hallucination size} $\eta_{max}$ is limited by $\|x_{det} \|$ which is smaller than the \textit{minimum hallucination size} $\eta_{min}$. So there is no value satisfying Definition \ref{def:detailtrans}.
\subsubsection{Quantitative Link Between the Error and the Diameter of the Feasible Sets}\label{par:QuantHallMRI}
In this paragraph, we analyse quantitatively the link between the diameter of the feasible set $\diam(F_y)$ and the error as described in $\S$ \ref{sec:exp_quant_setup}. We analyse the sharpness of the hallucination lower bound in Theorem \ref{thm:suff_hall} on the FastMRI data for the U-Net. 

\noindent As noted previously, automatic search in the feasible set $F_y$ cannot be directly performed using the FastMRI data set. To obtain quantitative results, we automatically generate 2000 data point pairs within a common feasible set. We observe that, in brain scans, the central ventricle is typically located at a constant position, except in rare cases of enlarged ventricles (38 patients among thousands, according to annotations in \cite{zhao_fastmri_2022}). This allows us to segment the ventricle using a fixed-position rectangle. The segmented ventricles are then used as selected details in the drawing algorithm described in the previous paragraph. This process yields two images associated with similar measurements close to $y \in \cM_2$. The first is the original ground truth image $x$, and the second is the modified scan $x + x_{det}$, where the ventricle is replaced. Both images are assumed to belong to the model set $\cM_1$, as illustrated in Figure \ref{fig:qual_hall_MRI}, row C. Here, we consider the approximate feasible sets by $F_y^N = \{x, x + x_{det}\}$, where the approximation $F_y^N$ clearly underestimates $F_y$. We consider the semi-norm, $\|.\| = \|.\|_{1, 1, \{R\}}$, with $R$ being the bounding box of the segmented ventricle.

\begin{figure}[h!]
    \centering
    \includegraphics[width=0.8\textwidth]{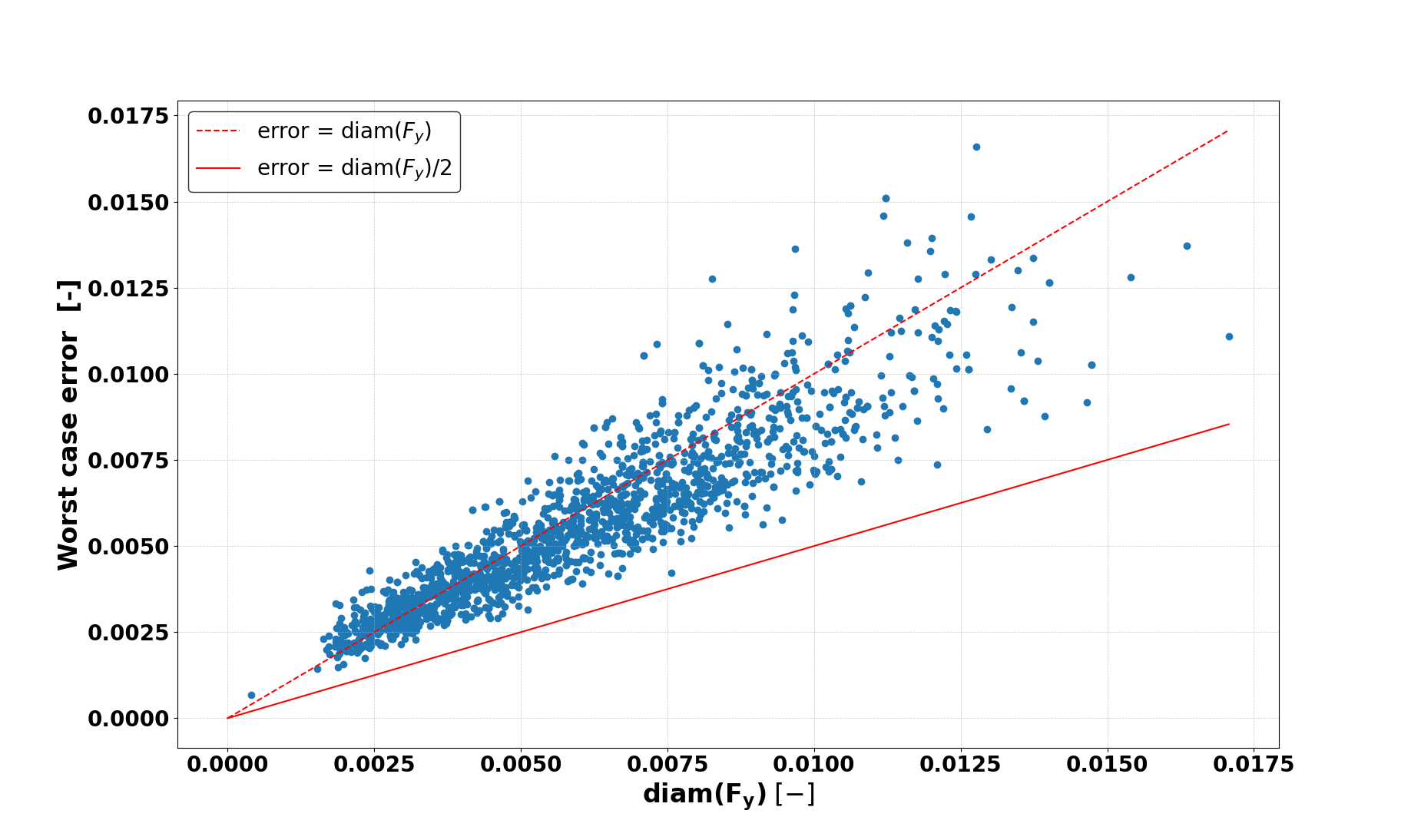}
    \caption{Sharpness of the lower bound in Theorem \ref{thm:suff_hall}. Both the errors and the diameters are measured in terms of $\ell^1$ norm normalised by the number of pixels in a patch. }
    \label{fig:MRI_LB_sharp}
\end{figure}

\noindent In Figure \ref{fig:MRI_LB_sharp}, we exclude MRI scans for which the two measurements (in $\cM_2$) are too dissimilar (SNR below 30).
We also exclude scans for which either reconstruction is inconsistent according to Definition \ref{def:consistent} of consistent decoders (SNR below 26).
Finally, we exclude scans for which the reconstructions differ excessively relative to the size of the detail. After filtering, the plots are generated using 1686 data points out of the original 2000. We present only the graphs based on the reconstructions of the decoder $\phi(fx)$.
Those using the decoder reconstructions $\phi(f(x + x_{det}))$ yield similar results and lead to the same conclusions.
Figure \ref{fig:MRI_LB_sharp} demonstrates that while the lower bound (LB) is rarely achieved, the \textit{worst-case error} remains of the same order of magnitude. Achieving the LB would imply that the decoder predicts the mean between $x$ and $x + x_{det}$, thereby abstaining from being highly accurate for either data point. Instead, the decoder typically predicts an element that corresponds to neither the image $x$ nor the image with the detail $x + x_{det}$.
Additionally, the figure indicates that the diameter of the approximate feasible set $\diam(F_y^N)$ cannot be considered an upper bound (UB) of the worst-case error. This discrepancy arises from the incomplete characterisation of the data set $\cM_1$.
\subsection{Super Resolution of Sentinel-2 Data}\label{sec:S2SR}
Sentinel-2 (S2) data comprises 13 spectral bands: four bands (Red, Green, Blue, and Near Infrared) with a 10 m Ground Sampling Distance (GSD), six bands with 20 m GSD, and three bands with 60 m GSD. Each S2 image results from processing Earth-surface details using the satellite's imaging sensors.
Single-image super-resolution aims to reconstruct a high-resolution image with enhanced visual quality and finer structural details from a low-resolution input. However, even with prior information, the mapping from low to high resolution remains inherently ill-posed, as multiple plausible high-resolution images may correspond to the same low-resolution observation.
In this work, we focus on super-resolving the 10 m GSD Sentinel 2 bands to 2.5 m GSD, achieving a $4 \times$ resolution factor. Unless specified otherwise, we use the decoder introduced in \cite{TrustworthwSR2025}. To evaluate their approach, the authors generated a data set of paired 2.5 m- and 10 m-GSD images from the RGB-NIR bands, following the methodology described in \cite{OpenSRTest2025}.
The proposed decoder $\phi$ for this super-resolution task is stochastic and belongs to the family of diffusion models. Given an input image $y \in \mathbb{R}^{4 \times H \times W}$ (the RGB-NIR bands at 10 m GSD), the decoder outputs a set of reconstructions: $\phi(y) = \{z_1, \dots, z_L\} \subset \X = \mathbb{R}^{4 \times 4H \times 4W}$. For this decoder, $W = H = 128$. A linear model approximates the downsampling process of Sentinel 2 data
\[
\begin{aligned}
    f : \mathbb{R}^{4 \times 4H \times 4W} \to \mathbb{R}^{4 \times H \times W},
\end{aligned}
\]
where $f$ applies downsampling with the model $DS$ to each band:
\[
\begin{aligned}
    DS : \mathbb{R}^{4H \times 4W} \to \mathbb{R}^{H \times W}.
\end{aligned}
\]
\cite{OpenSRTest2025} implement $DS$ using bilinear interpolation for a scaling factor of $1/4$ with anti-aliasing via the PyTorch library. We use a data set from \cite{OpenSRTest2025}, consisting of 179 pairs of 2.5 m GSD images and their downsampled versions obtained using $f$.
\subsubsection{Qualitative Perspective of Hallucinations}\label{par:QualHall_S2SR}
\begin{figure*}[h!]
\centering
\setlength{\tabcolsep}{2pt}
\renewcommand{\arraystretch}{1}
\begin{tabular}{c*{4}{c}}
  &\parbox{3cm}{\centering $x$\\ (Original image)} & \parbox{3cm}{\centering $x+P_{\cN(f)}x_d$\\ (Synthetic image)} &  \parbox{3cm}{\centering $fx$\\ (Low resolution\\ original image)} &  \parbox{3cm}{\centering $f(x+P_{\cN(f)}x_d)$\\ (Low resolution \\ synthetic image)} \\
\textbf{A} &
\includegraphics[width=0.23\textwidth]{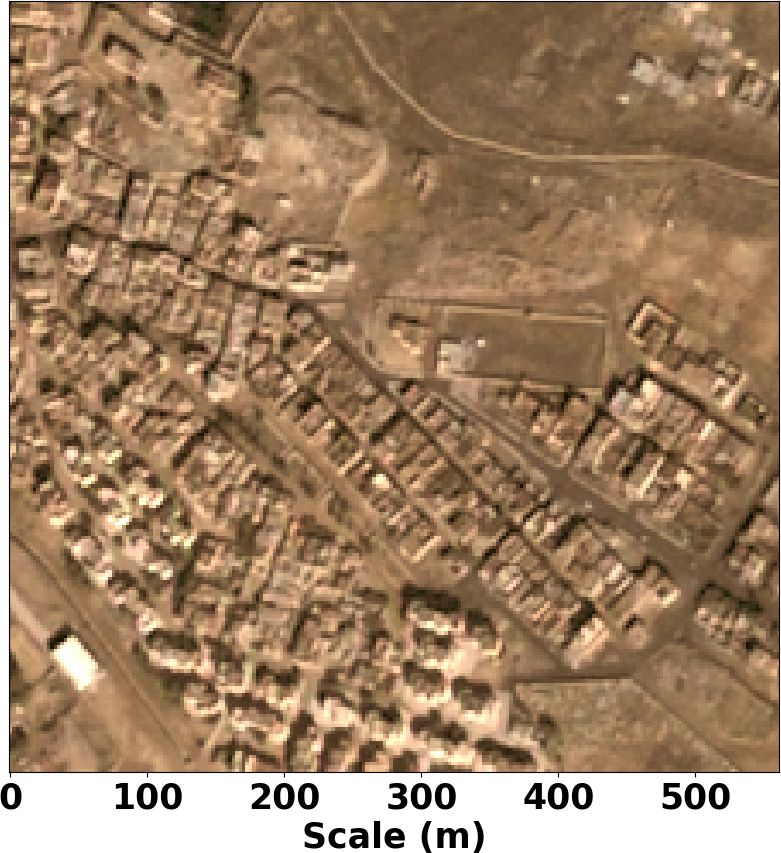} &
\includegraphics[width=0.23\textwidth]{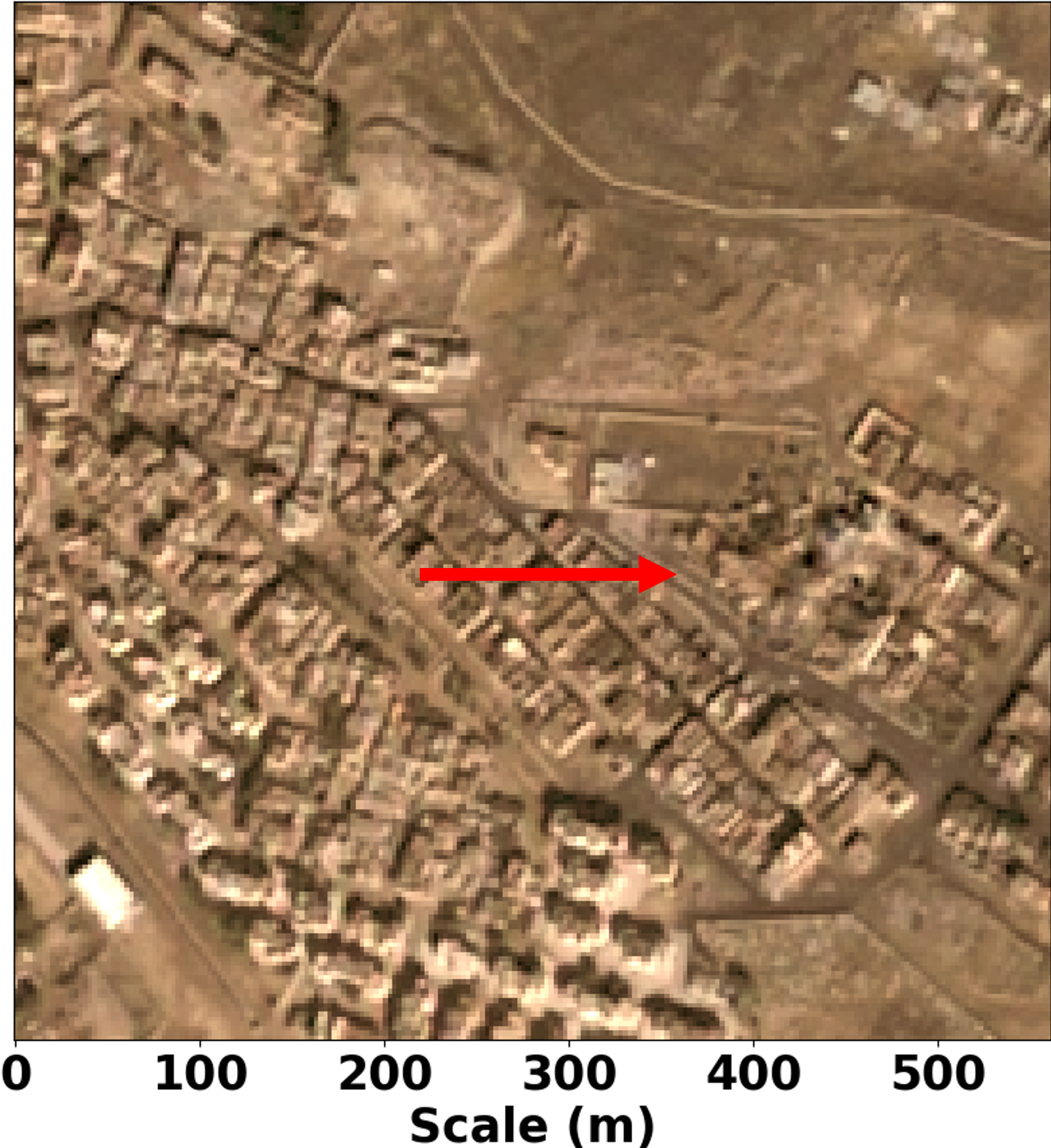} &
\includegraphics[width=0.23\textwidth]{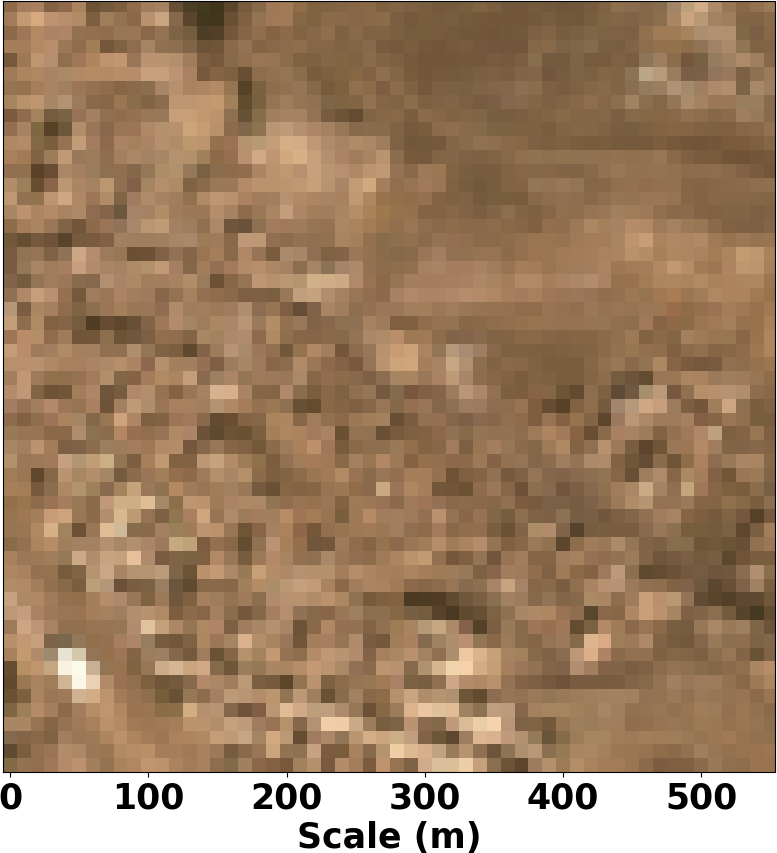} &
\includegraphics[width=0.23\textwidth]{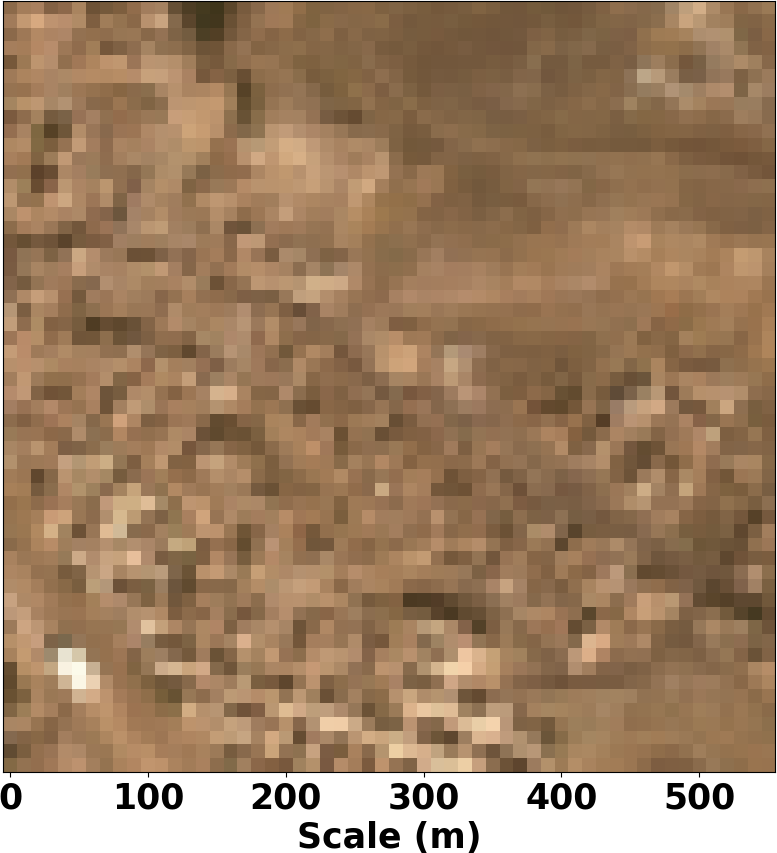}  \\
\textbf{B} &
\includegraphics[width=0.23\textwidth]{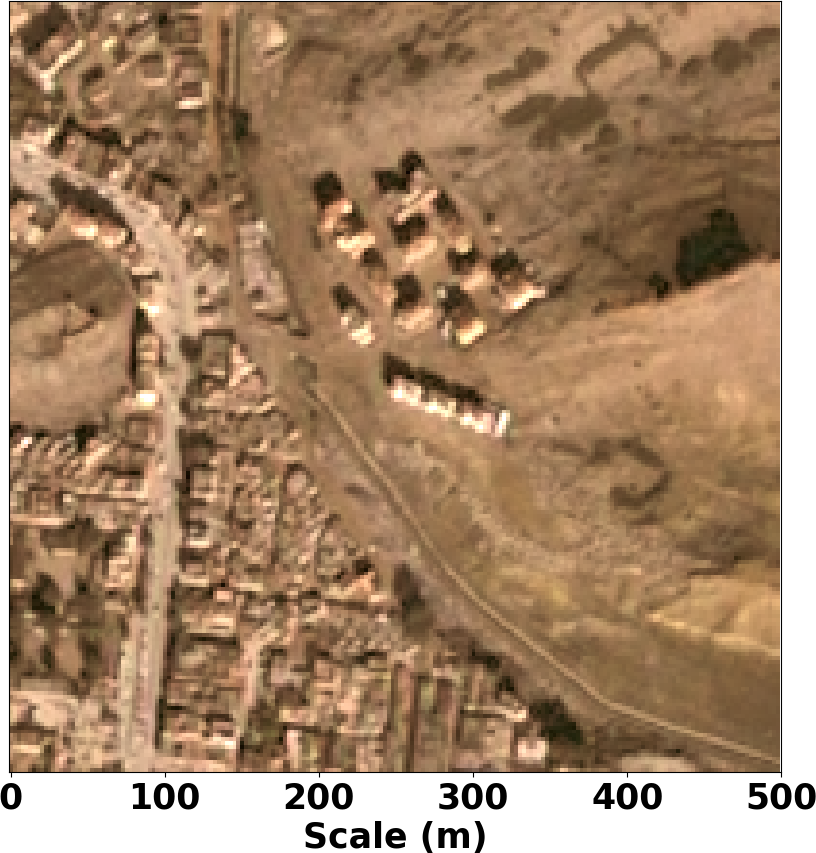} &
\includegraphics[width=0.23\textwidth]{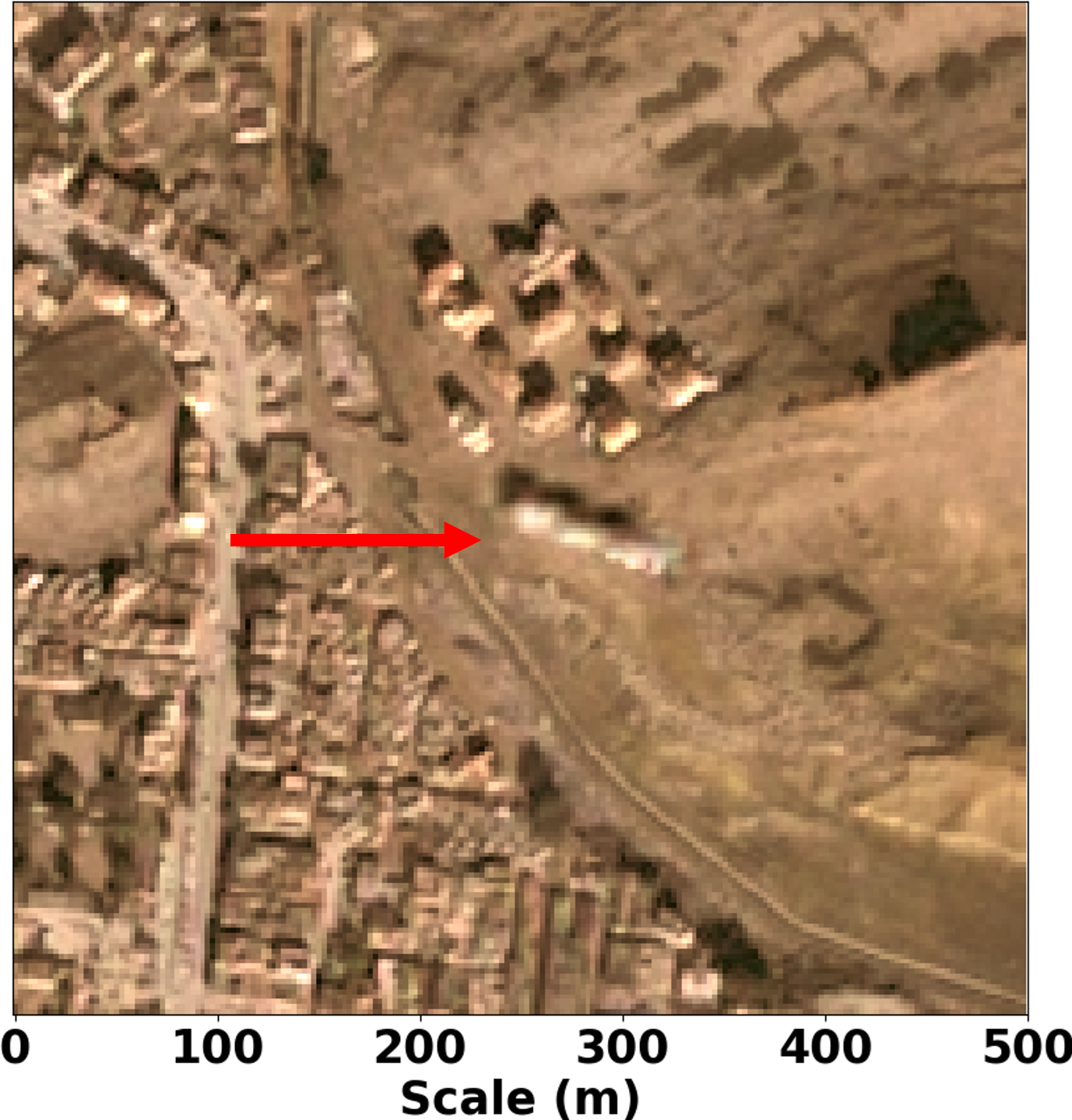} &
\includegraphics[width=0.23\textwidth]{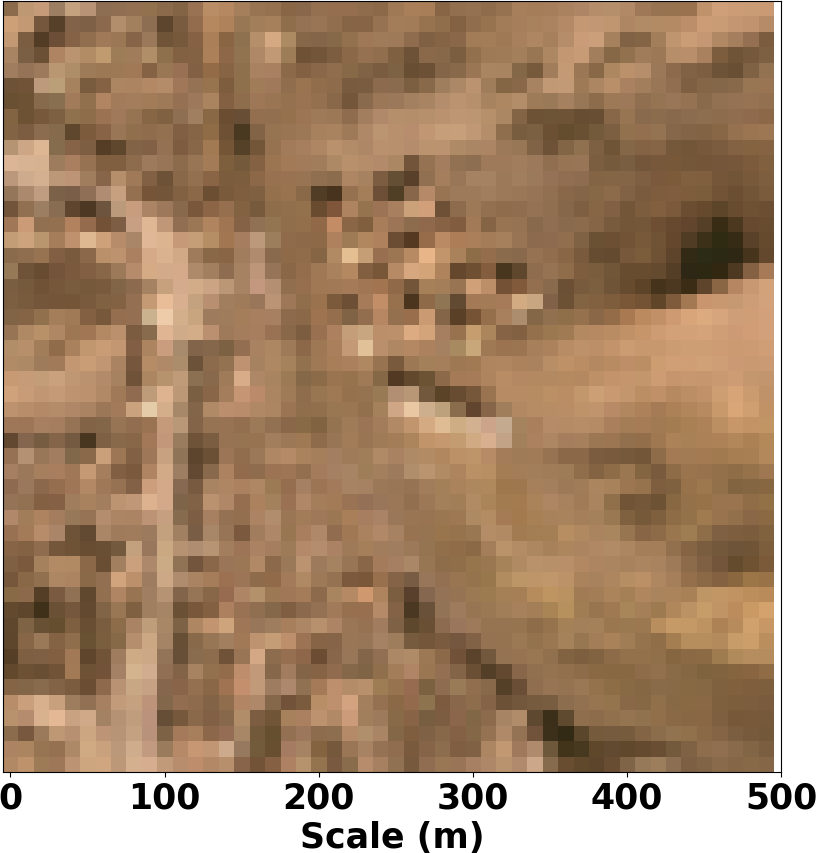} &
\includegraphics[width=0.23\textwidth]{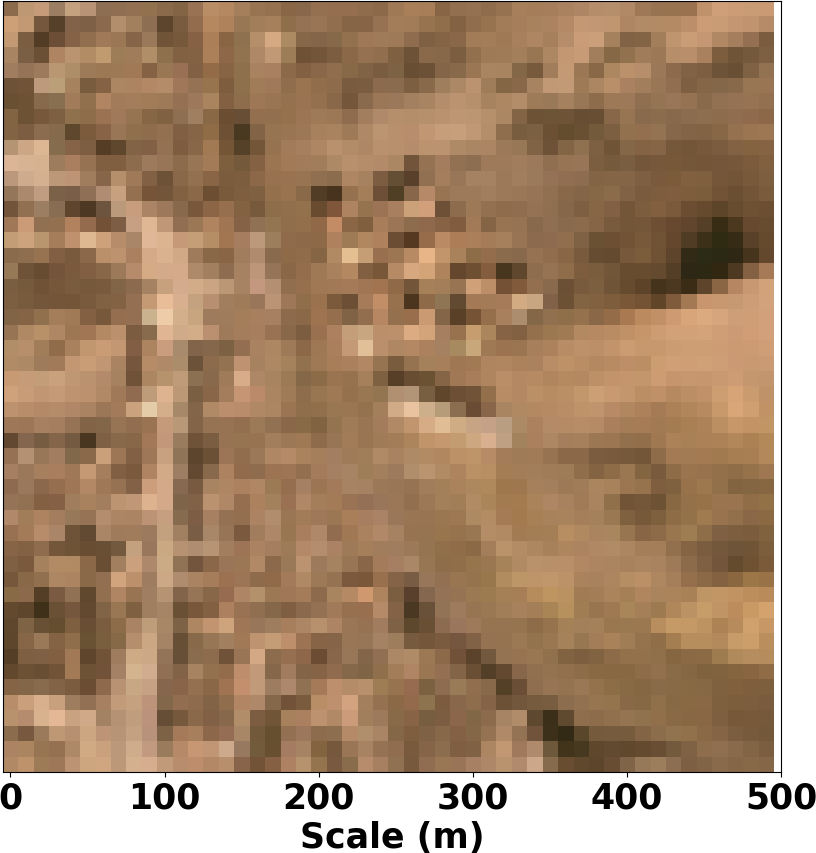}\\
\end{tabular}
\caption{Examples of significant detail drawing following the method described in Figure \ref{fig:kerdraw}. A red arrow points at the added or removed detail when it is not directly recognizable.}
\label{fig:hall_ex_draw_S2SR}
\end{figure*}

First, we identify potentially incorrectly transferred details $x_{d}$ using the decoder-dependent method described in $\S$ \ref{sec:methods}. For a given high-resolution image $x$ in the data set, we either add or remove $x_{det} = P_{\cN(f)}x_d$, where $x_d$ represents a significant detail extracted from the data set (for example, a building, trees, or a path) and $P_{\cN(f)}$ is the approximate null-space projection for the downsampling operator (see Appendix). Two examples of this process are presented in Figure \ref{fig:hall_ex_draw_S2SR}.
\noindent In the first row of Figure \ref{fig:hall_ex_draw_S2SR}, $x$ has buildings in a good state whereas in $x_{det}$ they are not. The projection $P_{\cN(f)}x_d$ is close to $x_d$ and $x+P_{\cN(f)}x_d$ is realistic. Hence, no decoder can determine, from the corresponding LR input alone, which state the building is in.

\noindent In the second row,  $x_d$ represents a removed set of houses, but  $P_{\cN(f)}x_d$ only blurs them. Here,  $x+P_{\cN(f)}x_d$ can hardly be considered as realistic because the houses have a non-negligible component in the measurement space. Therefore, any consistent decoder will predict a similar structure there and can be trusted.
\subsubsection{Verification of Definition \ref{def:detailtrans}}\label{par:verif_def_S2SR}
\begin{figure*}[h!]
\centering
\setlength{\tabcolsep}{2pt}
\renewcommand{\arraystretch}{1.5}
\begin{tabular}{m{2.7cm} c c c}
   & \parbox{2.5cm}{\centering Synthetic image\\ $x$} 
   & \parbox{2.5cm}{\centering LR synthetic image} 
   & \parbox{2.5cm}{\centering SR image \\ from synthetic \\ $x+x_{det} = \phi(fx)_0$} \\ \noalign{\smallskip}
\parbox{2.7cm}{\small{$\eta_{min} = 102.12$} \\ \small{$\eta_{max} = 210.17 $}\\ \small{$\|x_{det}\|=205.10$} }&
$\vcenter{\hbox{\includegraphics[width=0.23\textwidth]{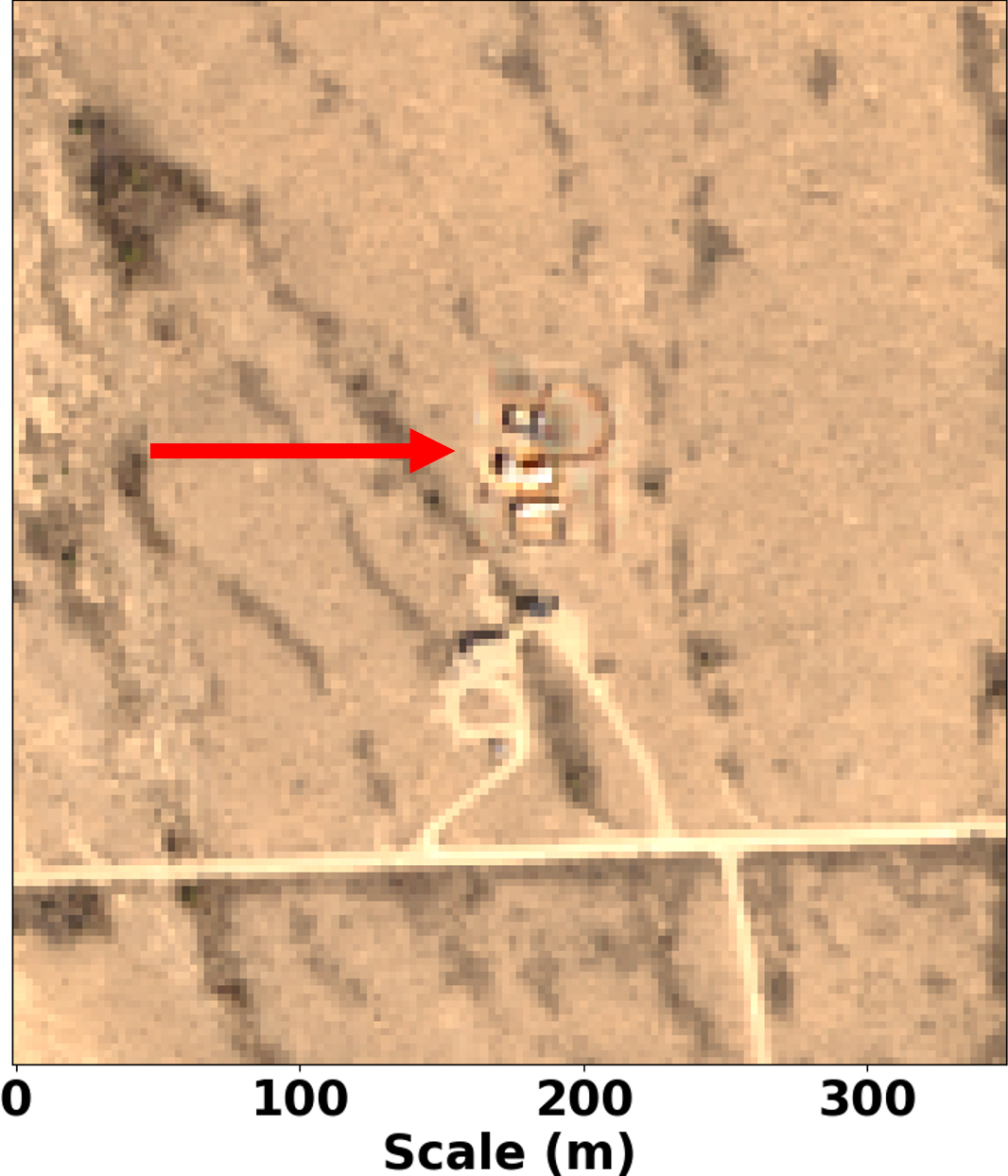}}}$ &
$\vcenter{\hbox{\includegraphics[width=0.23\textwidth]{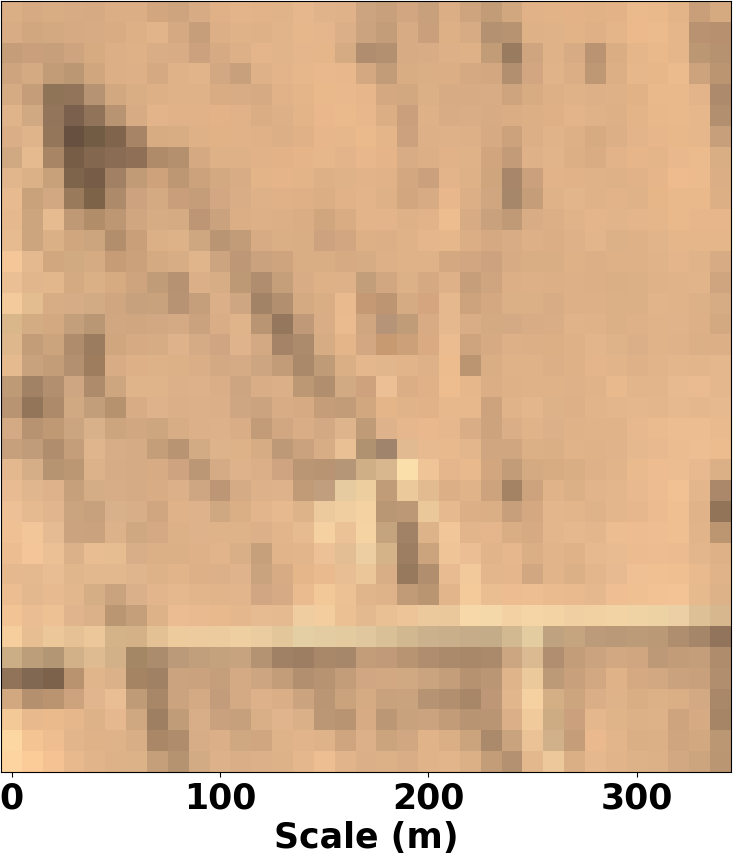}}}$ &
$\vcenter{\hbox{\includegraphics[width=0.23\textwidth]{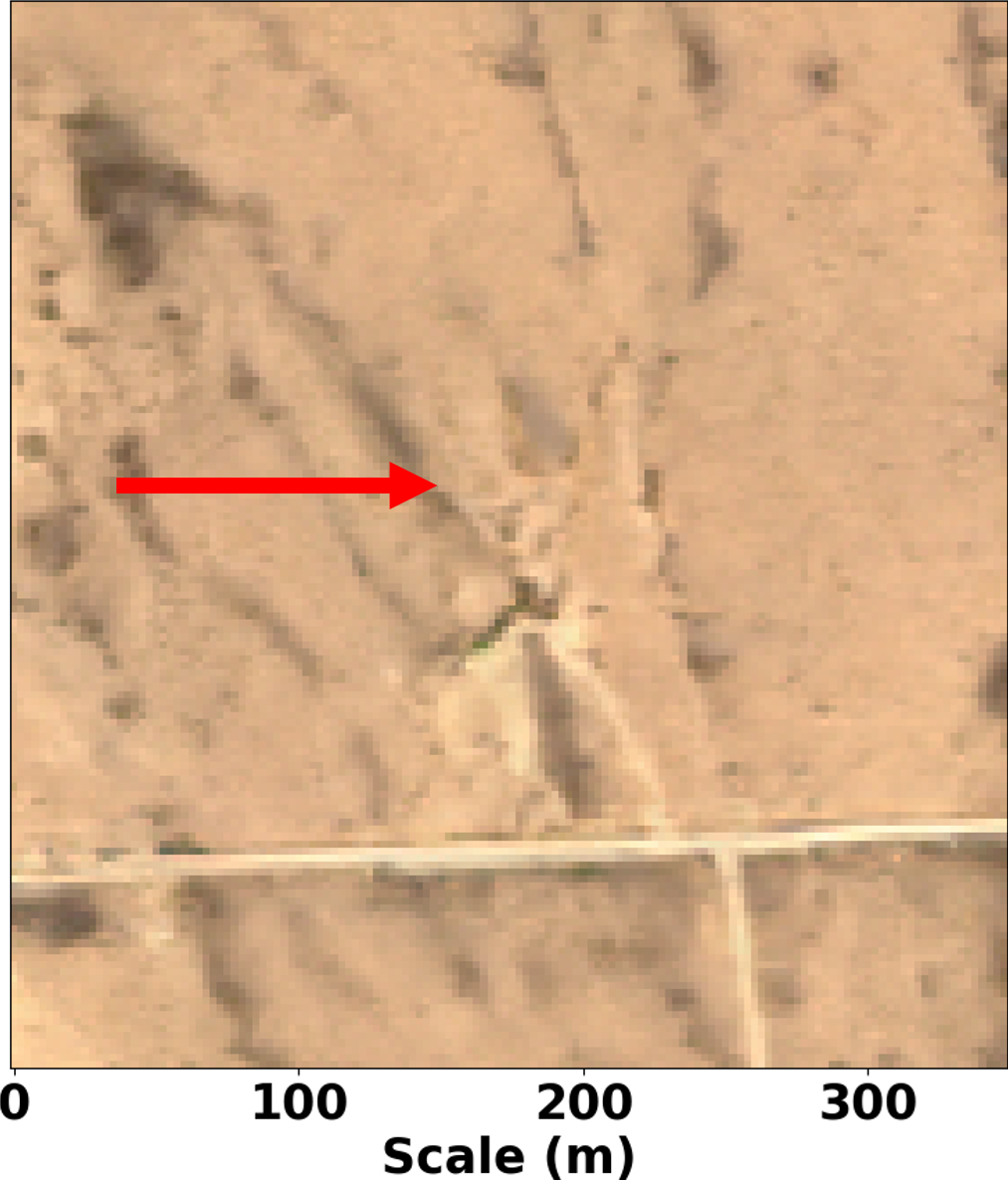}}}$ \\ \noalign{\medskip}
\parbox{2.7cm}{\small{$\eta_{min} = 37.03 $} \\ \small{$\eta_{max} = 89.90 $}\\ \small{$\|x_{det}\| = 86.02$}} &
$\vcenter{\hbox{\includegraphics[width=0.23\textwidth]{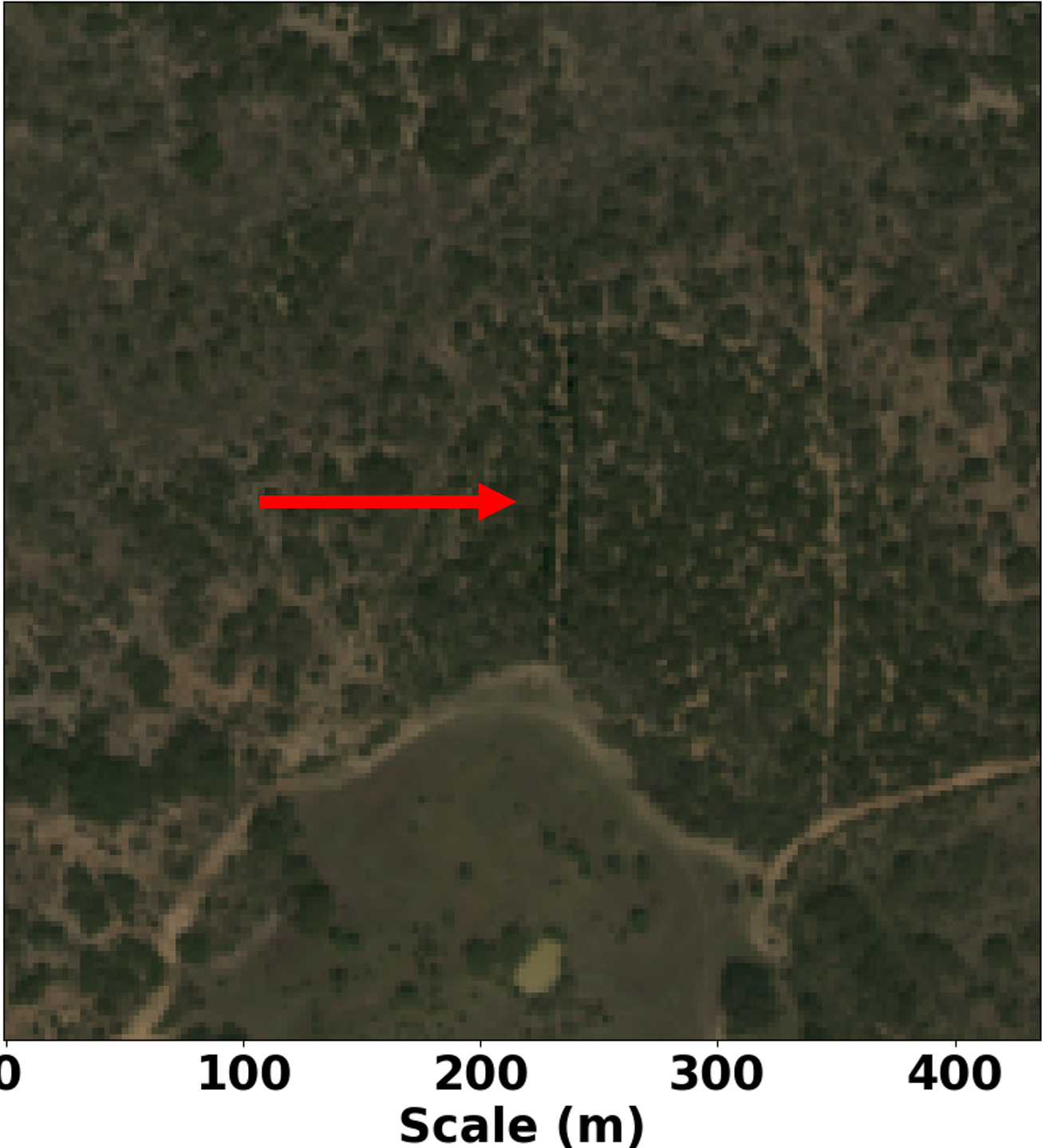}}}$ &
$\vcenter{\hbox{\includegraphics[width=0.23\textwidth]{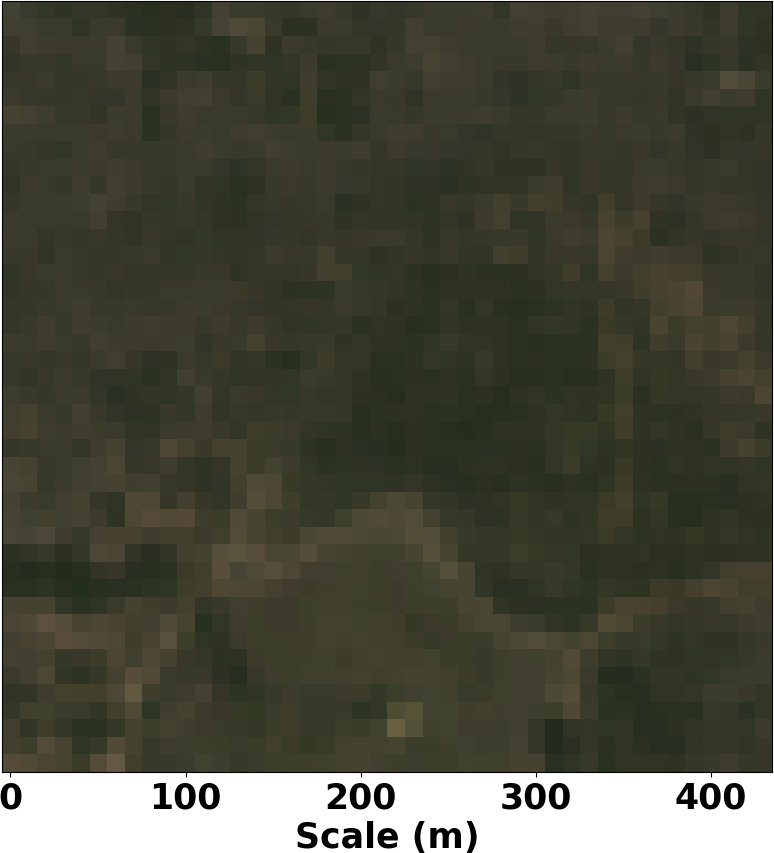}}}$ &
$\vcenter{\hbox{\includegraphics[width=0.23\textwidth]{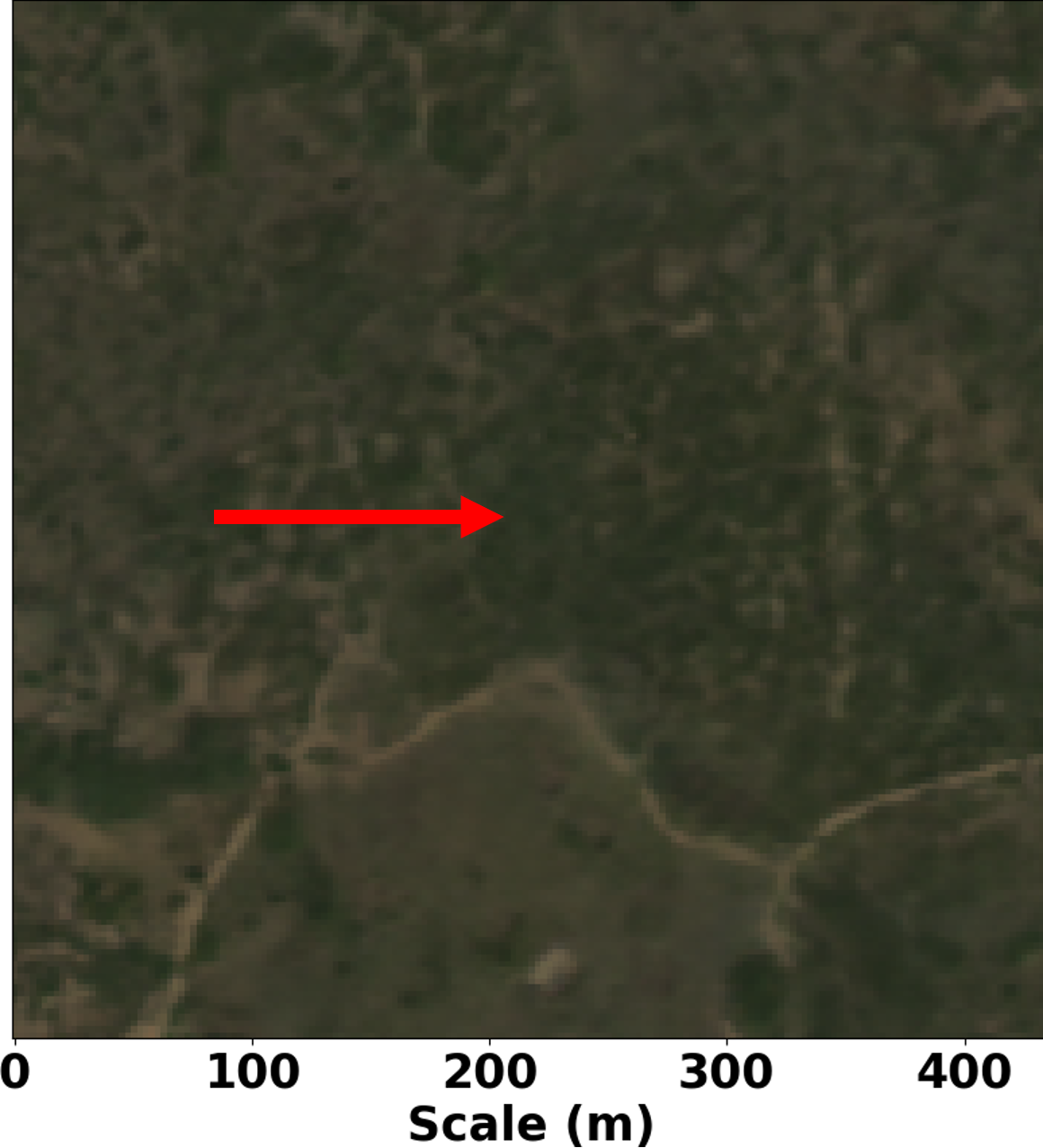}}}$ \\ \noalign{\medskip}
\parbox{2.7cm}{\small{$\eta_{min} = 61.66$} \\ \small{$\eta_{max} = 103.35$}\\ \small{$\|x_{det}\| = 98.21$}} &
$\vcenter{\hbox{\includegraphics[width=0.23\textwidth]{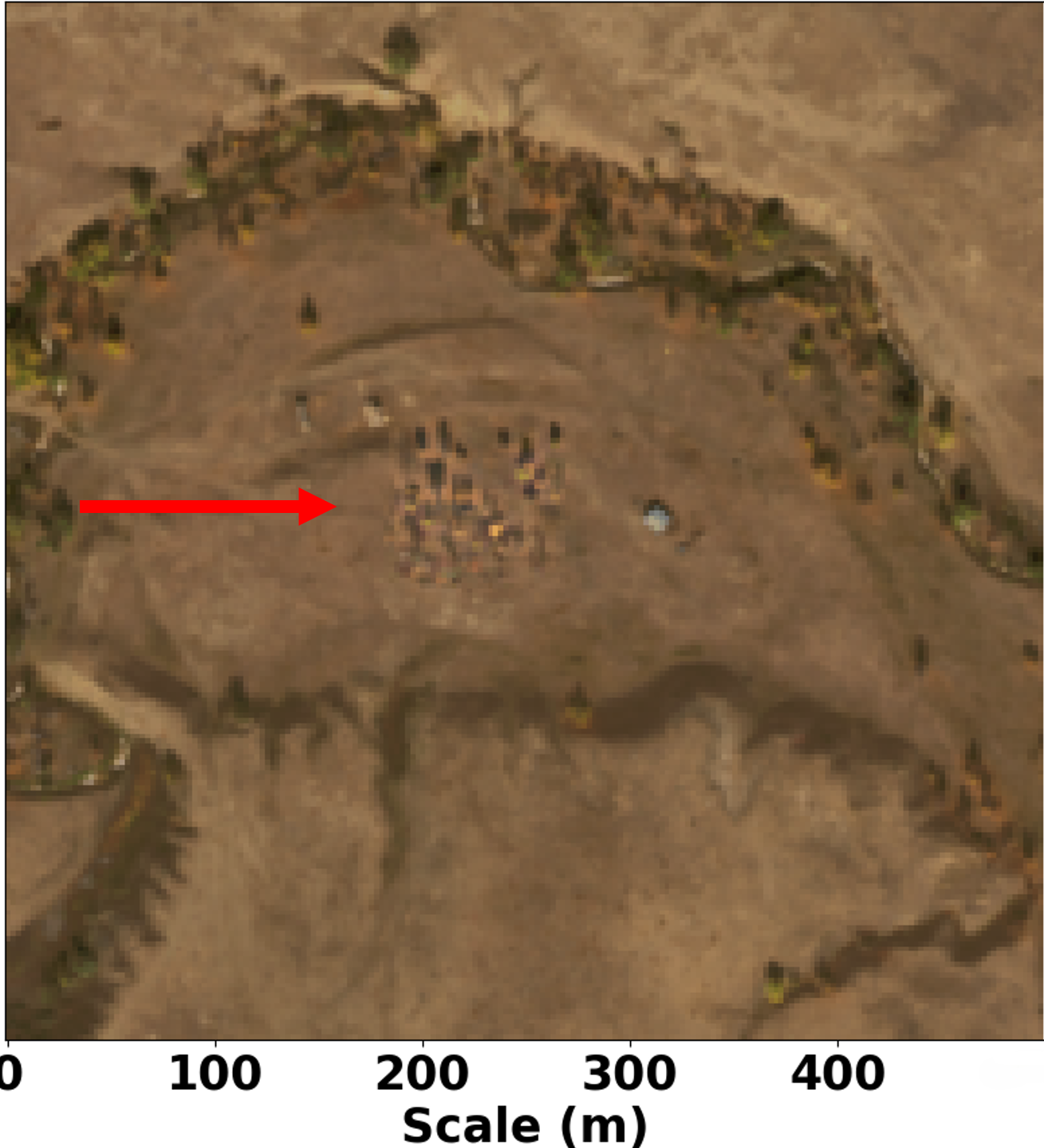}}}$ &
$\vcenter{\hbox{\includegraphics[width=0.23\textwidth]{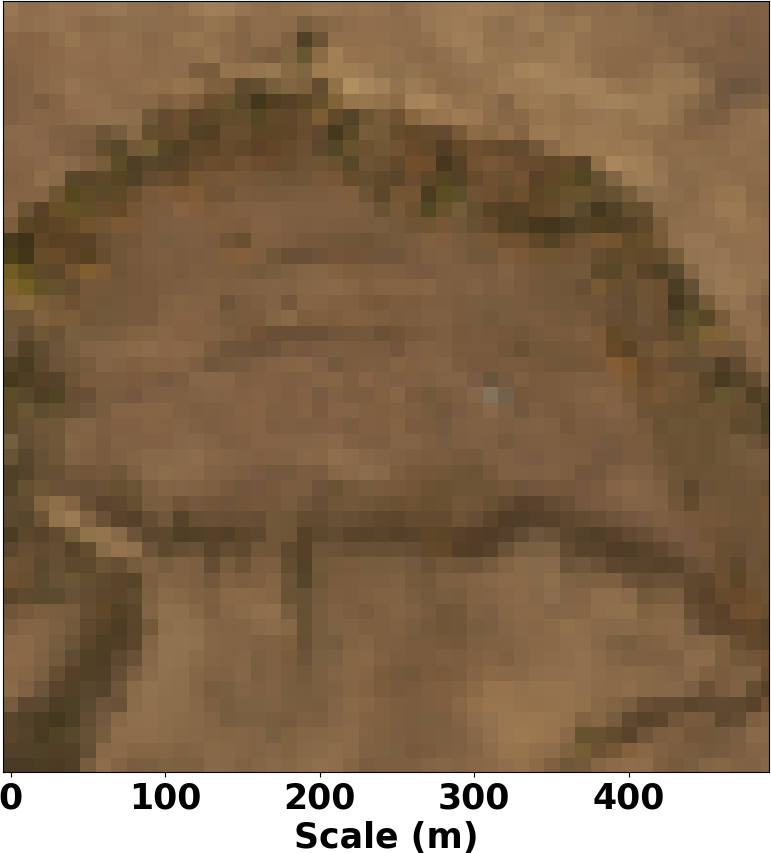}}}$ &
$\vcenter{\hbox{\includegraphics[width=0.23\textwidth]{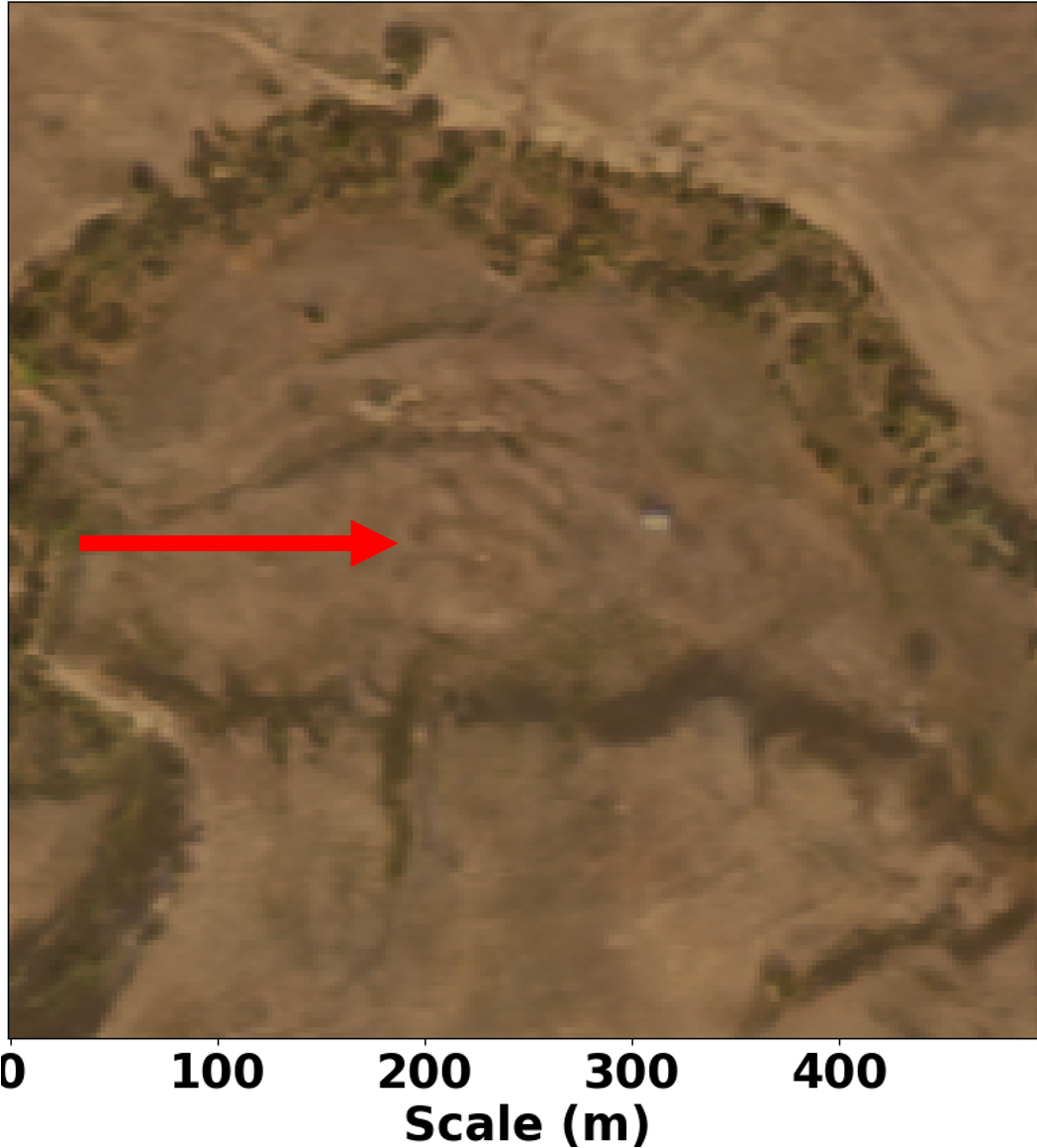}}}$ \\
\end{tabular}
\caption{Quantities of Definition \ref{def:detailtrans} and Theorem \ref{thm:iff_shift} for super-resolution of Sentinel-2 data. All the reconstructions are consistent. The distance is calculated for the region of interest associated with the added detail and is normalised by the number of pixels in this region. For the first row, the quantities were computed with only $7$ selected reconstructions out of $11$ in $\phi(y)$, to satisfy the condition $\eta_{min} \leq \|x_{det}\|/2$. A red arrow points to the added or removed detail when it is not immediately recognisable. Low-resolution (LR) images are shown in the middle column and super-resolved (SR) images in the right column. }
\label{fig:hall_def_S2SR}
\end{figure*}

In this paragraph, we use the detail-pasting method to give examples of detail-transfer.
The quantities in Definition \ref{def:detailtrans} are computed following the setup in $\S$ \ref{sec:exp_verif_setup}.For this experiment, we assume that the synthetic image represents the ground truth $x$.
The original image in the data set represents the image with the incorrect detail transfer. This corresponds to the first reconstruction from the synthetic image, denoted as $\phi(fx)0$. We regard this reconstruction as $x+x_{det}$. We consider a subset of noise restricted to $\mathcal{V} = \{0 \}$ and $\phi(fx)$ is made of the $11$ first generated reconstructions from the diffusion model of \cite{TrustworthwSR2025} : $\phi(fx) =  \set{\phi(fx)_l,\quad 0 \leq l \leq 10}$. Following the definitions in $\S$ \ref{sec:exp_verif_setup}, we have : 
\[
\begin{aligned}
    \eta_{min} &= d(\phi(fx), x+x_{det}) = \sup_{1 \leq l \leq 11} \| \phi(fx)_l- (x+x_{det}) \| ,\\
    \eta_{max} &= d(\phi(fx), x) =\sup_{1 \leq l \leq 11} \| \phi(fx)_l- x \|.
\end{aligned}
\]
Note in Figure \ref{fig:hall_def_S2SR} that the measurement $fx_{det}$ has a negligible magnitude. This is confirmed by the fact that the reconstructions from $fx$ are similar to those from $f(x+x_{det})$. Therefore, we can suppose that the decoder is accurate on $x+x_{det}$ for a larger subset of noise.
Note also that \textit{the minimum hallucination size} $\eta_{min}$ may be relatively large. This occurs because the reconstructions in $\phi(y)$ may contain various semantic details, all of which differ from the original image $x$. However, it is possible to artificially restrict $\phi(fx)$ to only the reconstructions similar to $x+x_{det}$. This can be done down to a single element, $\set{\phi(fx)_0}$, by discarding the other elements of the set $\phi(fx)$. In that extreme case, $\eta_{min} = 0$ and $\eta_{max} = \| x_{det}\|$.
\subsubsection{Application of the Decoder-Agnostic Method for Automated Detail Transfer Anticipation}\label{par:model_agn_S2SR}
\noindent The previous method allows one to manually identify details that any decoder solving the described super-resolution problem inevitably transfers. In this paragraph, we explore the possibility of automatically finding the minimal local error when super-resolving an image. To that end, we use the search in $F_y$ in the decoder-agnostic part of the method described in $\S$ \ref{sec:methods}.

\noindent Due to the high dimension of the images ($4 \times 512 \times 512$ in high resolution), it is highly unlikely that the data set contains non-trivial feasible sets within $179$ pair of images. To mitigate this problem, we split each high-resolution image into patches of size $4 \times 16 \times 16$, and the corresponding low-resolution patch has size $4 \times 4 \times 4$. This way, we obtain a data set consisting of pairs of high- and low-resolution patches.  Then, we apply the algorithms presented in $\S$ \ref{par:algs_kersize} to the patch data set. Since the patch dimension is significantly smaller than that of full images, and the number of patches is on the order of $10^5$, the algorithms can output nontrivial feasible sets. Each data point in this alternative data set corresponds to the size of a detail in Sentinel-2 data. Note that we use Remark \ref{rem:tuples_gen} to apply the algorithms of $\S$ \ref{par:algs_kersize} directly on the patched data set. This application is compatible with the downsampling model because of the sparsity shown in Appendix: each pixel in the LR image depends only on its neighbouring pixels within a distance of 2 in the high-resolution image. More details are shown in Appendix. This is equivalent to considering the semi-norm $\|\cdot \|_{1,1, { \mathcal{P} }}$, where $\mathcal{P}$ denotes a patch in an arbitrary position on the high-resolution images. On the low-resolution images, we consider the norm $\|\cdot\|_{\infty,2,\mathcal{P}_{LR}},$ where $\mathcal{P}_{LR}$ denotes the grid of low-resolution patches.
\begin{figure*}[h!]
\centering
\setlength{\tabcolsep}{2pt}
\renewcommand{\arraystretch}{1}
\begin{tabular}{c*{4}{c}}
  &\parbox{3cm}{\centering Low resolution} & \parbox{3cm}{\centering Super-resolved} & \parbox{3cm}{\centering Image of $lim_1$ patches} & \parbox{3cm}{\centering Image of $lim_2$ patches} \\
\textbf{A} &
\includegraphics[width=0.23\textwidth]{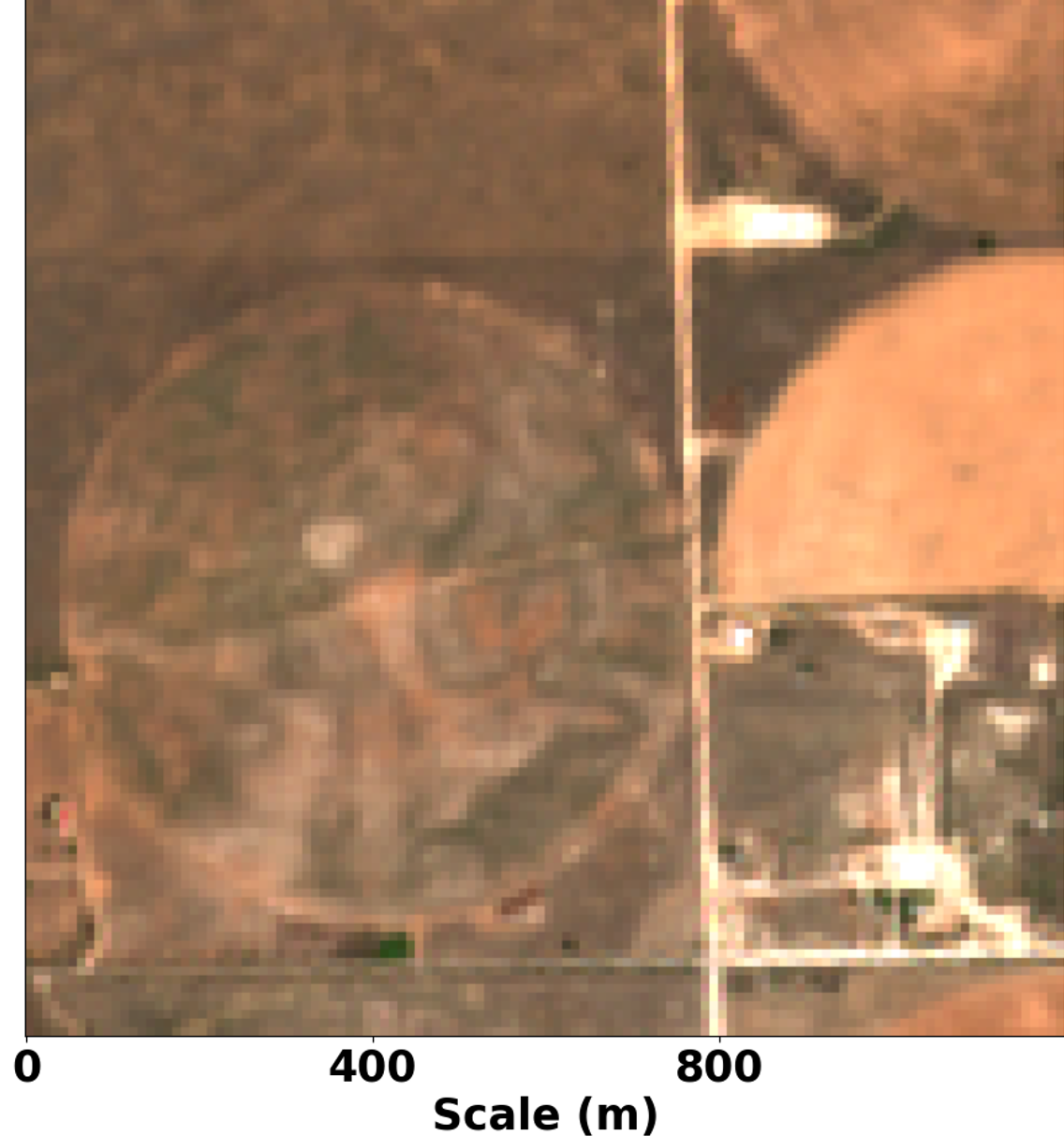} &
\includegraphics[width=0.23\textwidth]{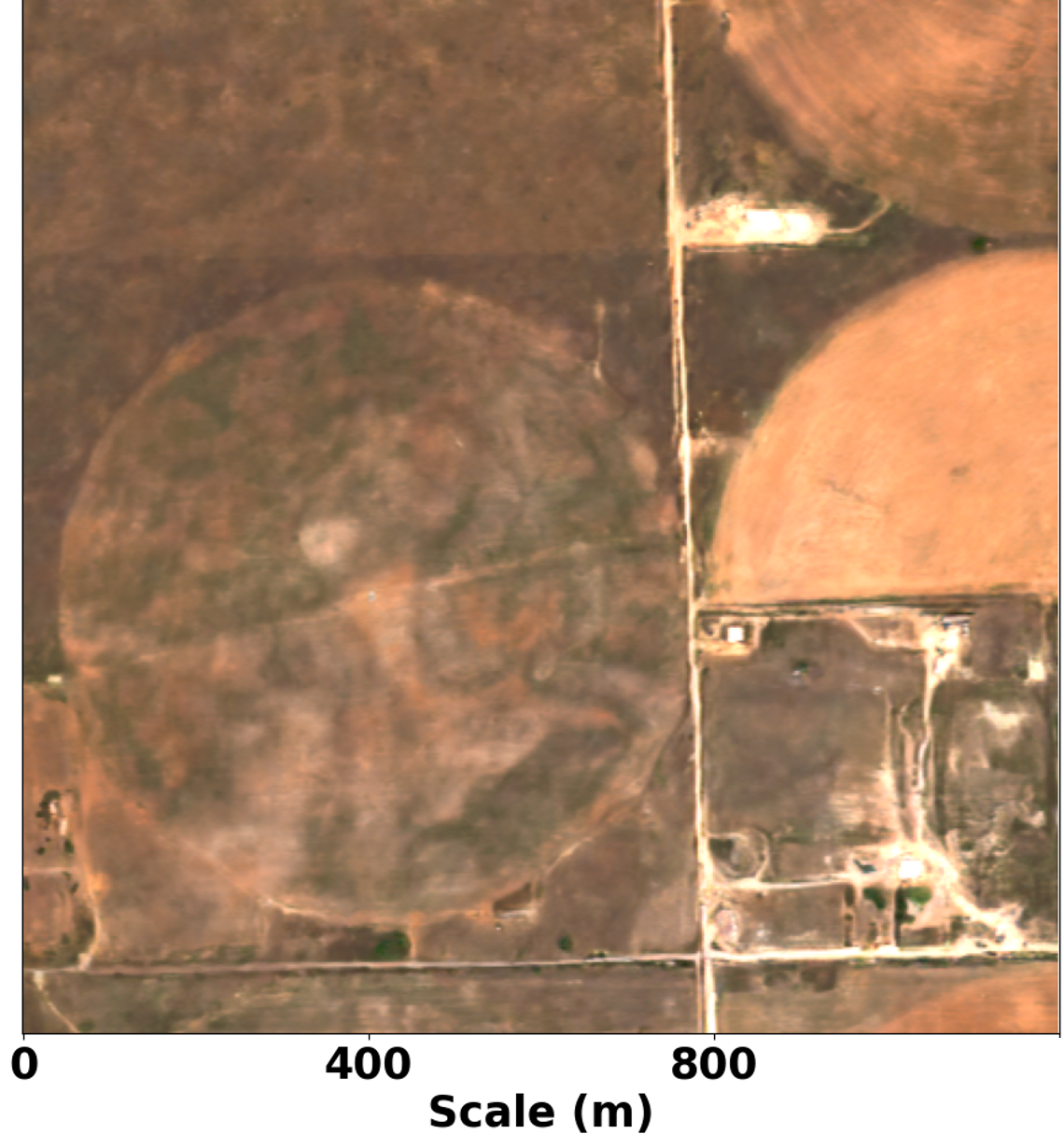} &
\includegraphics[width=0.23\textwidth]{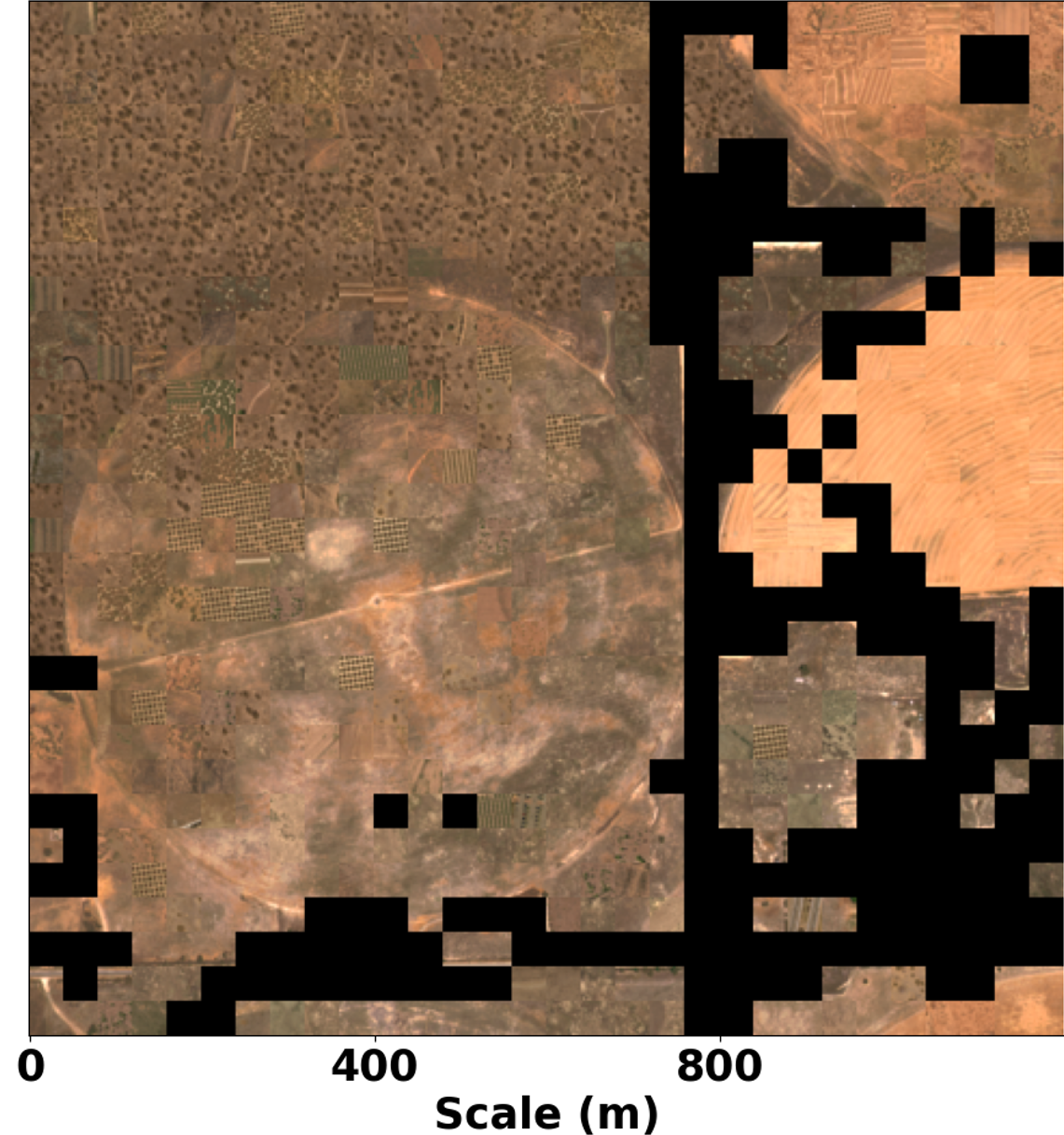} &
\includegraphics[width=0.23\textwidth]{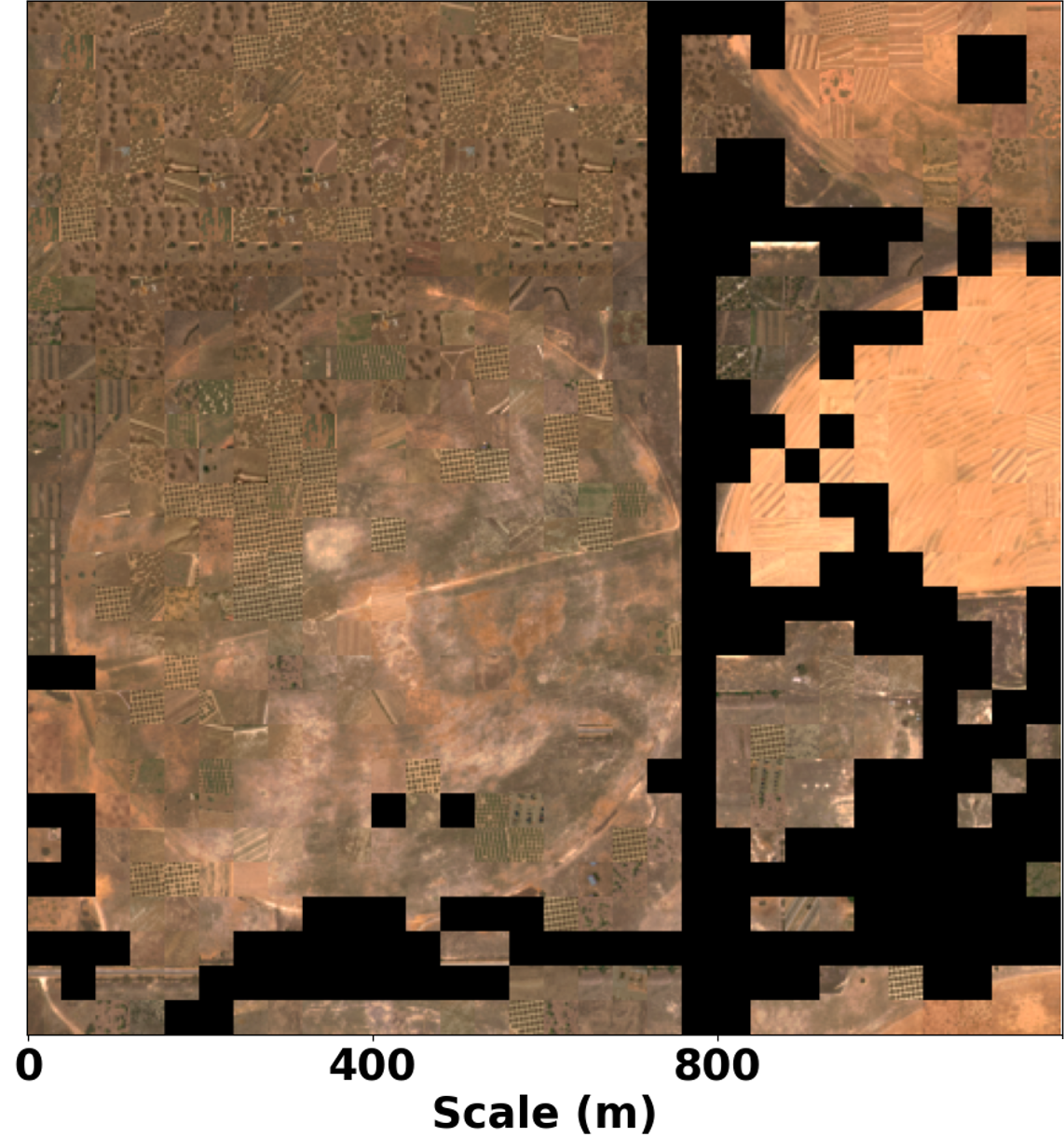}  \\
\textbf{B} &
\includegraphics[width=0.23\textwidth]{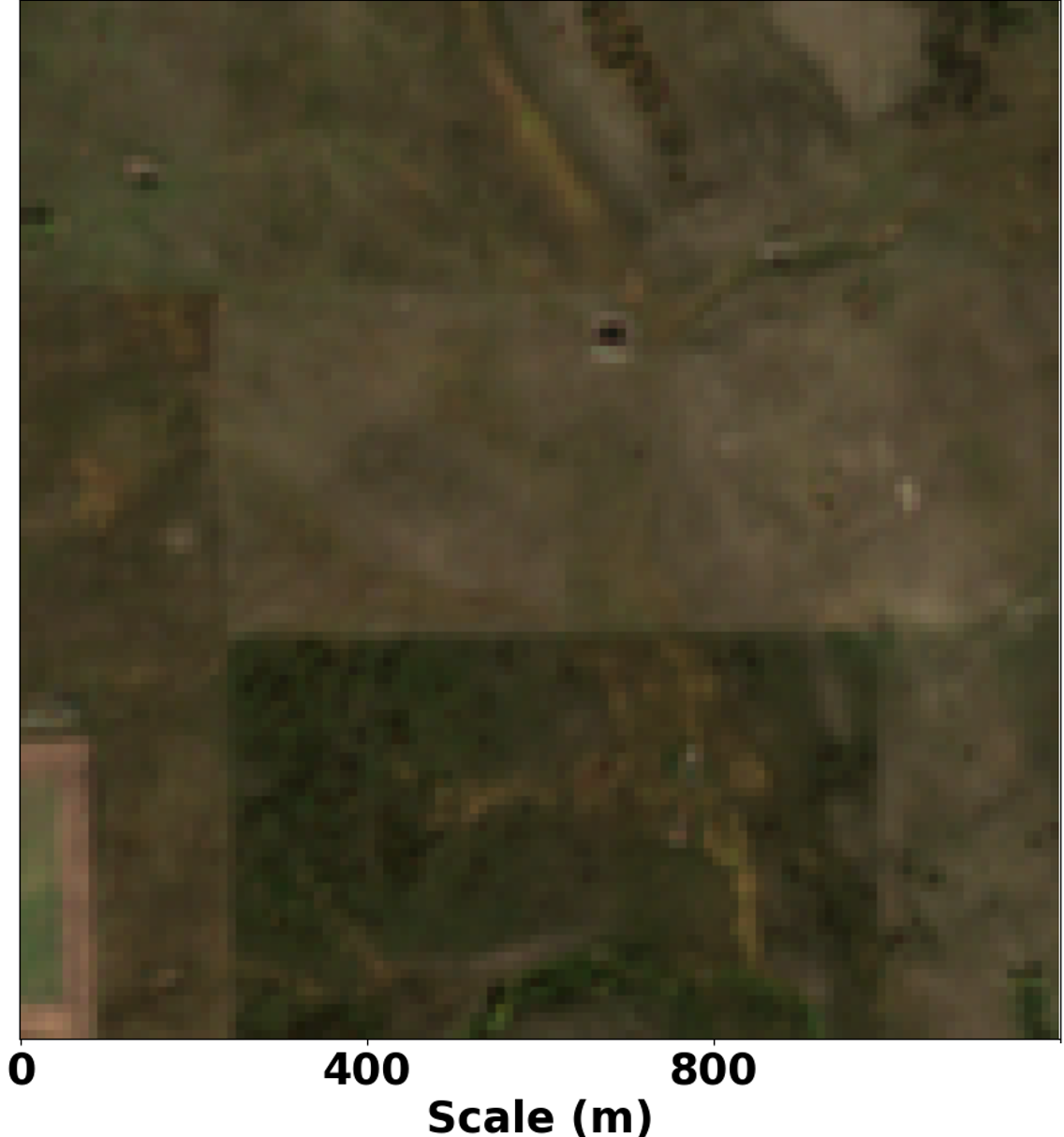} &
\includegraphics[width=0.23\textwidth]{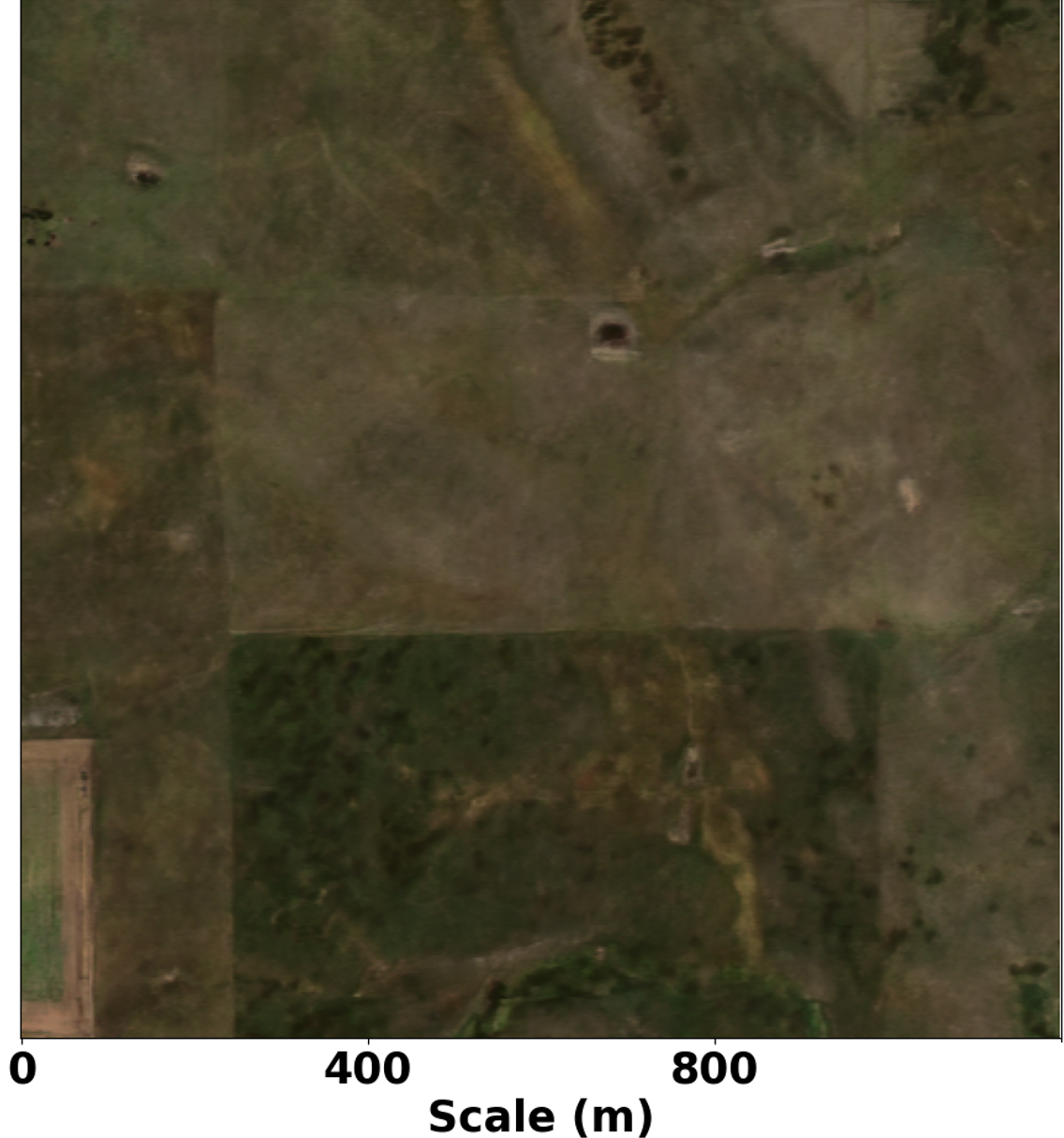} &
\includegraphics[width=0.23\textwidth]{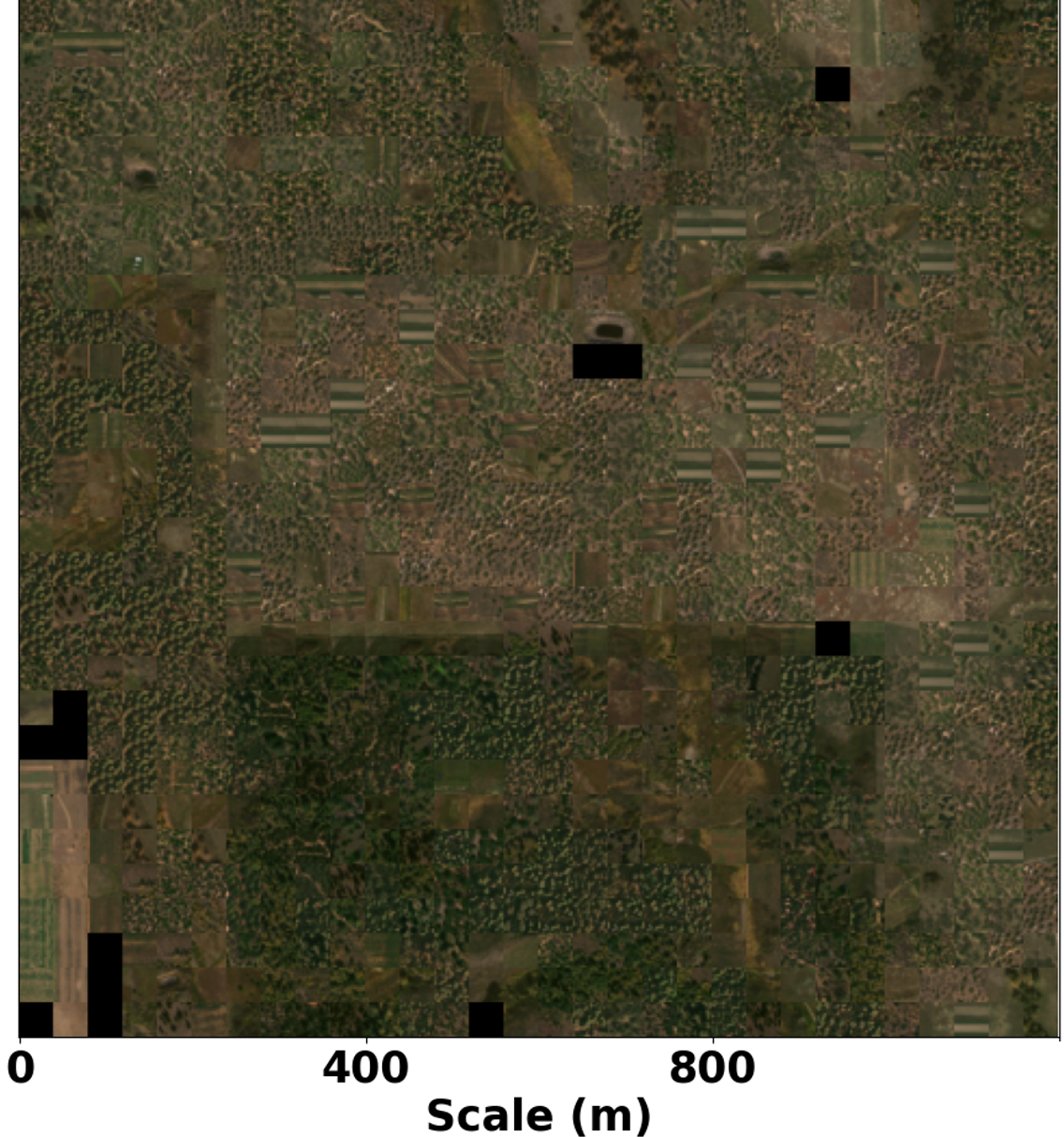} &
\includegraphics[width=0.23\textwidth]{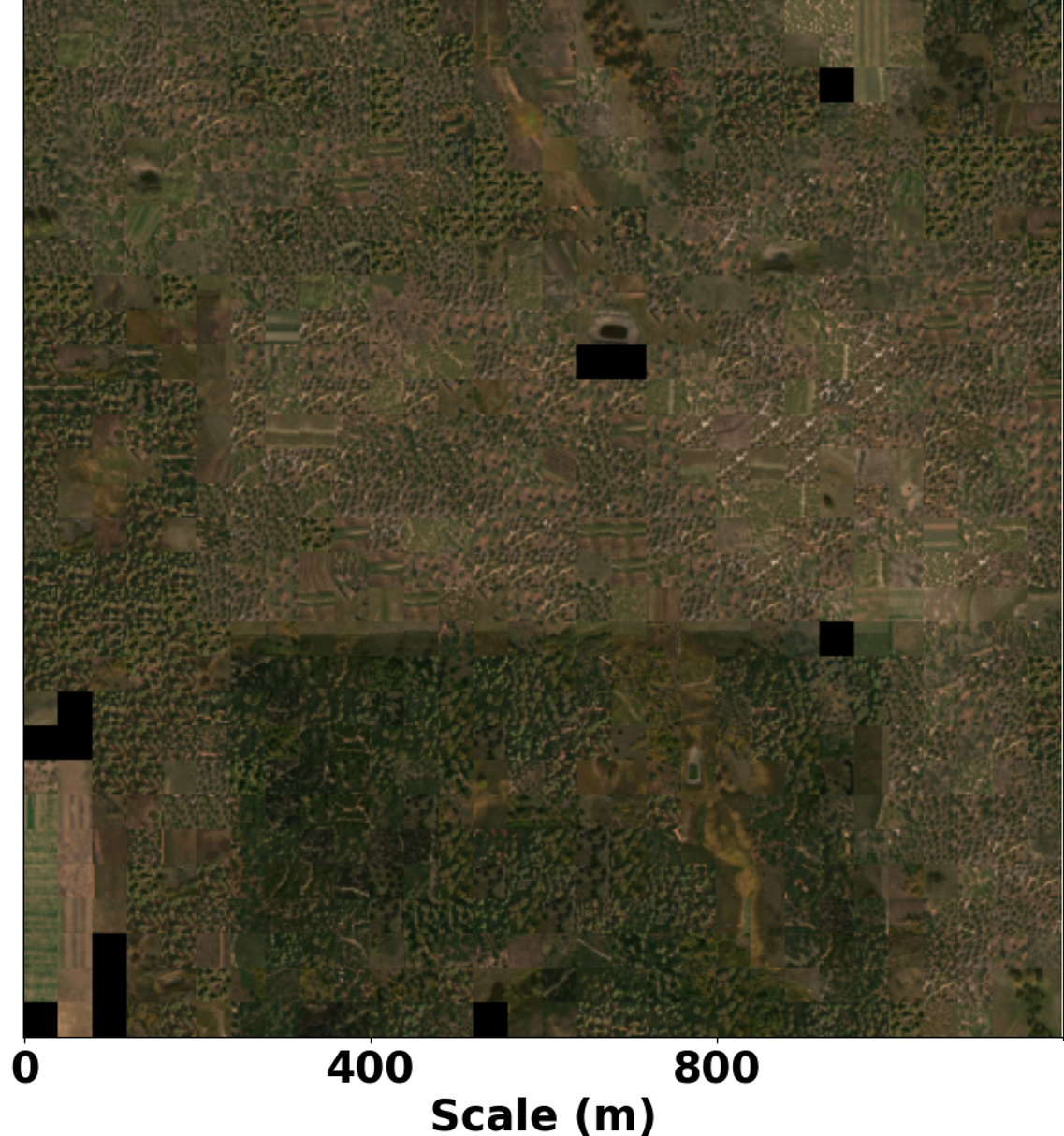}  \\
\end{tabular}
\caption{Patch-wise application of the decoder agnostic method of Figure \ref{fig:method}. Dark areas of the image correspond to patches with a null diameter of the feasible set.}
\label{fig:method_agnostic_S2SR}
\end{figure*}
\noindent Since the desired output for hallucination anticipation comprises potentially hallucinated details, we present two output patches for each low-resolution patch. These output patches are generated by the decoder-agnostic algorithm described in $\S$ \ref{sec:methods}. For visualisation, these results are arranged in the same spatial order as the original input image, as illustrated in Figure \ref{fig:method_agnostic_S2SR}. However, because each patch is processed independently, the resulting images are not expected to form a globally realistic structure. Each patch should therefore be evaluated individually. While extrapolations could be performed to ensure global coherence, such analysis is beyond the scope of this study. For instance, as observed in Figure \ref{fig:method_agnostic_S2SR}, the presence of farmland in the middle of a forest is unlikely, yet remains physically possible if it is consistent with the low-resolution input.
\subsubsection{Quantitative Link Between the Error and the Diameter of the Feasible Sets}\label{par:quant_hall_S2SR}
\begin{figure}[h!]
    \centering
    \includegraphics[width=0.8\textwidth]{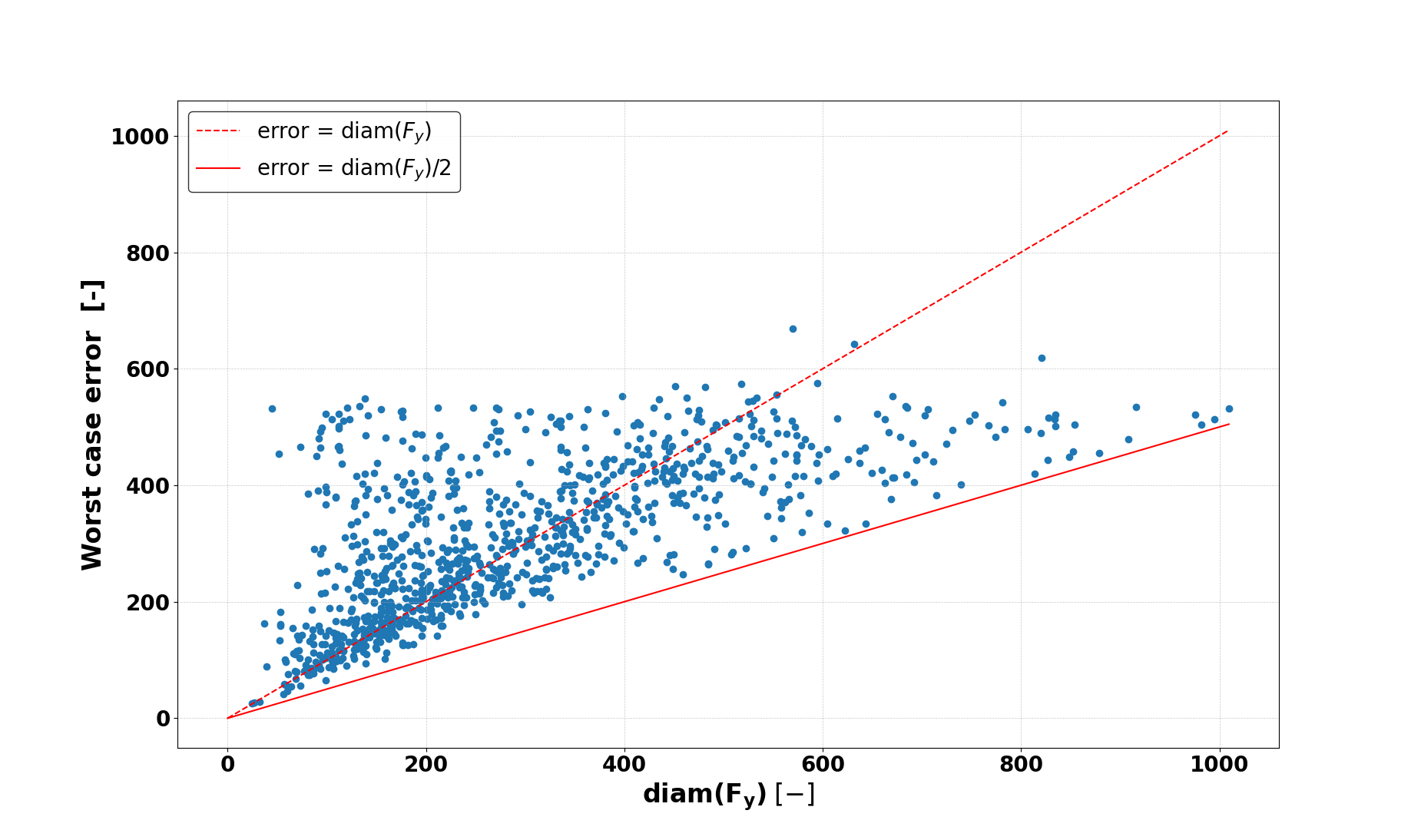}
    \caption{Sharpness of the lower bound in Theorem \ref{thm:suff_hall}. Both the errors and the diameters are normalised by the number of pixels in a patch. $1000$ data points out of more than $160,000$ have been randomly sampled for readability. Each patch has its reflectance values in the range $[0, 10,000]$. The error and the diameter are given in terms of $\ell^1$ norm  normalized by the number of pixels}    \label{fig:S2SR_LB_sharp}
\end{figure}

\noindent The objective of this section aligns with the quantitative analysis presented in $\S$ \ref{par:QuantHallMRI}.
Its principle is described in $\S$ \ref{sec:exp_quant_setup}. As with MRI acceleration, quantitative analysis requires a large number of data point pairs in the data set that lie within a common feasible set. This requirement is satisfied using the patched data set described in the previous section.
\noindent For each low-resolution patch $py$, the approximate feasible set $F_{py}^N$ is determined using the decoder-agnostic method outlined in Figure \ref{fig:method}. The compatibility with the problem on full images is explained in Appendix. In this experiment, we use only the decoder's first reconstruction, treating the decoder $\phi$ as single-valued. The $\ell^1$ norm is employed to compare patches, which is equivalent to using the norm $\|.\|_{p,1,\{P\}}$ on full images, where $P$ denotes a patch at any position and $p \geq 1$.
 Figure \ref{fig:S2SR_LB_sharp} demonstrates that, in our setting, the accuracy lower bound is optimal. This suggests that the decoder predicts values near the centre of the feasible set $F_y^N$, effectively abstaining from producing sharp images. Additionally, Figure \ref{fig:S2SR_LB_sharp} shows that the diameter of the approximate feasible set $\diam(F_y^N)$ can serve as an upper bound for the detail-transfer magnitude in a non-negligible proportion of cases. However, as noted in Remark \ref{rem:sizebound}, it cannot be reliably used as an upper bound for every data point.
\section{Discussion}
\noindent The theoretical results presented in this work provide an analytical foundation for understanding AI hallucinations in inverse problems. We theoretically and empirically link hallucinations to the inherent ill-posedness of the problem through computable quantities such as the worst-case kernel size $\operatorname{Kersize}(F, \cM_1, \E)$ and feasible-set diameters $\diam(F_y)$ for measurements $y \in \cM_2$. Thus, we move beyond empirical observations toward a quantitative characterisation of reconstruction limits for inverse-problem decoders. Since the results are analytical and rely on relatively weak assumptions, they apply to a large variety of inverse problems.

\noindent A notable aspect of our framework is the definition of hallucinations through detail transfer. This definition depends on the choice of a score, norm, or seminorm used to compare reconstructions with reference signals. In the theoretical results, this comparison must satisfy the assumptions of the corresponding statement, in particular the triangle inequality; in the convergence results, the topology induced by a norm is used. In our experiments, region-of-interest metrics ($\S$ \ref{sec:metrics}) separate hallucinations from accurate reconstructions, but other application-specific scores, including wavelet-based measures \cite{mallat2012groupinvariantscattering}, can also be integrated when their mathematical properties match the relevant assumptions. Since hallucination is an ambiguous concept, no metric can capture it perfectly. Determining which hallucination-size ranges are clinically or scientifically problematic remains a task for domain experts, for which our framework provides quantitative vocabulary and theoretical guarantees.

\noindent Several theoretical results, including Theorem \ref{thm:iff_shift}, Proposition \ref{prop:nogo_Fy}, Theorem \ref{thm:suff_hall}, and Proposition \ref{prop:sizenogo1}, require realistic decoder reconstructions, typically $\phi(B) \subset \cM_1$ for some $B \subset \cM_2$. This assumption is relevant because hallucinations are most problematic when they appear realistic despite being inaccurate. Verifying this assumption is closely related to out-of-distribution (OOD) detection. To the best of our knowledge, general guarantees for OOD detection remain open; existing guarantees are restricted to specific architectures, inputs, or problem classes \cite{hendrycks2019scaling,bitterwolf2020certifiably, sayyed2026encore,rifat2024dardadomainawarerealtimedynamic}.

\noindent Several limitations remain. The framework depends on application-specific choices of metrics or seminorms, and some guarantees require assumptions about the realism of decoder outputs. Moreover, the reconstruction-dependent method currently relies on known linear forward models. Future work should address stronger guarantees for OOD detection and the integration of controlled generative models into the proposed framework.

\noindent Theorem \ref{thm:iff_shift} provides necessary and sufficient conditions to characterise hallucinations using computable quantities without relying on arbitrary hyperparameters. A key practical implication is that avoiding hallucinations of a given size requires the training data set and forward model to induce a sufficiently well-posed problem. Our experiments in $\S$ \ref{par:exp_VDSR_MNIST} show that ill-posedness can transfer from training to testing sets, suggesting that future work should minimise the worst-case kernel size and feasible-set diameters through data set curation and refinement of the forward problem.

\noindent Algorithms \ref{alg:diamFy} and \ref{alg:feasapp} approximate feasible-set diameters from below, providing a lower bound on the degree of ill-posedness from a representative data set. As shown in the Sentinel-2 super-resolution experiments, even a small fraction of a standard training set can yield useful insight when reformatted appropriately, for example through patch-based splitting in $\S$ \ref{par:model_agn_S2SR}. Sample-based decoders, such as diffusion models and Markov chain Monte Carlo methods \cite{kaipio2005statistical, stuart2010inverse}, can automatically generate reconstructions for uncertainty or hallucination quantification \cite{TrustworthwSR2025, Bhadra2022MiningTM}, but often lack the hands-on control needed for expert inspection of specific details. Our reconstruction-dependent method addresses this gap for linear forward models by allowing practitioners to assess individual details manually. As distributional control in sampling-based methods improves, methods with stronger guarantees could be integrated into our framework.
\section{Conclusion}
This study establishes a theoretical and algorithmic framework for detecting and anticipating AI hallucinations in inverse problems. By showing that hallucinations are not merely model failures but can arise from the ill-posedness of the underlying problem, we provide a quantifiable path toward more transparent and reliable AI systems.
Our decoder-agnostic and reconstruction-dependent methods enable hallucination assessment even when ground truth is unavailable, bridging theoretical guarantees and real-world applications. Generative models could automate and enhance the methods described in $\S$ \ref{sec:methods}, provided their reconstructions remain within a controlled distribution, such as $\cM_1$ or a subset of it, or that out-of-distribution reconstructions can be detected. This depends on future progress in OOD detection and generative control.
Overall, our results provide a principled way to quantify when hallucinations are unavoidable, assess the faithfulness of reconstructed details, and support expert-in-the-loop validation of AI-based decoders for inverse problems.

\bibliographystyle{abbrv}
\bibliography{references}

\end{document}